\documentclass[journal]{IEEEtran}

\ifCLASSOPTIONcompsoc
  \usepackage[nocompress]{cite}
\else
  \usepackage{cite}
\fi
\ifCLASSINFOpdf
\else
\fi

\pdfoutput=1

\usepackage{xcolor}
\usepackage{footnote}
\usepackage{multirow}
\usepackage{times}
\usepackage{epsfig}
\usepackage{graphicx}
\usepackage{amsmath}
\usepackage{mathtools}
\usepackage{amssymb}
\usepackage{fixltx2e}
\usepackage{color}
\usepackage{algpseudocode}
\usepackage[switch]{lineno} 

\usepackage{makecell}
\usepackage{footnote}
\usepackage{multicol}
\usepackage{subcaption}
\usepackage[linesnumbered,ruled,vlined]{algorithm2e}
\usepackage[]{threeparttable}

\newcommand\red[1]{{\color{black}#1}}

\graphicspath{{./figures/}{./figures/draw/}}

\usepackage[pagebackref=True,breaklinks=true,letterpaper=true,bookmarks=false]{hyperref}

\begin{document}

\title{\hspace{-5pt}Unsupervised Deep Cross-modality Spectral Hashing\hspace{-5pt}}


\author{
Tuan Hoang, Thanh-Toan Do, Tam V. Nguyen, Ngai-Man Cheung
\IEEEcompsocitemizethanks{
\IEEEcompsocthanksitem Tuan Hoang and Ngai-Man Cheung are with the Singapore University of Technology and Design (SUTD), Singapore.
Email:\newline nguyenanhtuan\_hoang@mymail.sutd.edu.sg, ngaiman\_cheung@sutd.edu.sg

Thanh-Toan Do is with the University of Liverpool, United Kingdom. ~Email: thanh-toan.do@liverpool.ac.uk
\IEEEcompsocthanksitem Tam V. Nguyen is with the University of Dayton, United States.
E-mail: \protect\\{tamnguyen}@udayton.edu
}
}



\IEEEtitleabstractindextext{%

}

\maketitle

\IEEEdisplaynontitleabstractindextext

\IEEEpeerreviewmaketitle

\newcommand{\ba}{\mathbf{a}}
\newcommand{\bb}{\mathbf{b}}
\newcommand{\bc}{\mathbf{c}}
\newcommand{\bk}{\mathbf{k}}
\newcommand{\bo}{\mathbf{o}}
\newcommand{\bs}{\mathbf{s}}
\newcommand{\bt}{\mathbf{t}}
\newcommand{\bu}{\mathbf{u}}
\newcommand{\bv}{\mathbf{v}}
\newcommand{\bw}{\mathbf{w}}
\newcommand{\bx}{\mathbf{x}}
\newcommand{\bz}{\mathbf{z}}

\newcommand{\bA}{\mathbf{A}}
\newcommand{\bB}{\mathbf{B}}
\newcommand{\bC}{\mathbf{C}}
\newcommand{\bD}{\mathbf{D}}
\newcommand{\bE}{\mathbf{E}}
\newcommand{\bG}{\mathbf{G}}
\newcommand{\bI}{\mathbf{I}}
\newcommand{\bL}{\mathbf{L}}
\newcommand{\bO}{\mathbf{O}}
\newcommand{\bR}{\mathbf{R}}
\newcommand{\bS}{\mathbf{S}}
\newcommand{\bT}{\mathbf{T}}
\newcommand{\bU}{\mathbf{U}}
\newcommand{\bV}{\mathbf{V}}
\newcommand{\bW}{\mathbf{W}}
\newcommand{\bWt}{\mathbf{W}^t}
\newcommand{\bWm}{\mathbf{W}^m}
\newcommand{\bWtt}{(\mathbf{W}^t)^\top}
\newcommand{\bWmt}{(\mathbf{W}^m)^\top}
\newcommand{\bX}{\mathbf{X}}
\newcommand{\bXt}{\mathbf{X}^t}
\newcommand{\bXm}{\mathbf{X}^m}
\newcommand{\bXtt}{(\mathbf{X}^t)^\top}
\newcommand{\bXmt}{(\mathbf{X}^m)^\top}
\newcommand{\bY}{\mathbf{Y}}
\newcommand{\bYs}{\mathbf{Y}^\star}
\newcommand{\bYm}{\mathbf{Y}^m}
\newcommand{\bYt}{\mathbf{Y}^t}
\newcommand{\bYst}{(\mathbf{Y}^\star)^\top}
\newcommand{\bYtt}{(\mathbf{Y}^t)^\top}
\newcommand{\bYmt}{(\mathbf{Y}^m)^\top}
\newcommand{\bZ}{\mathbf{Z}}

\newcommand{\bOnes}{\mathbf{1}}
\newcommand{\bZeros}{\mathbf{0}}

\newcommand{\bTheta}{\mathbf{\Theta}}

\newcommand{\transpose}{\hspace{-0.15em}^\top\hspace{-0.15em}}
\newcommand{\eq}{\hspace{-0.15em}=\hspace{-0.15em}}

\newcommand{\bbR}{\mathbb{R}}

\newcommand{\ndis}{\mathcal{N}}
\newcommand{\xset}{\mathcal{X}}
\newcommand{\oset}{\mathcal{O}}
\newcommand{\vset}{\mathcal{V}}
\newcommand{\tset}{\mathcal{T}}
\newcommand{\uset}{\mathcal{U}}
\newcommand{\dis}{\mathcal{D}}
\newcommand{\func}{\mathcal{F}}

\newcommand{\tr}{\text{Tr}}
\newcommand{\mAP}{\textbf{\textit{mAP}}}
\newcommand{\x}{\times}

\begin{abstract}


This paper presents a novel framework, namely \textit{Deep Cross-modality Spectral Hashing (DCSH)}, to tackle the unsupervised learning problem of binary hash codes for efficient cross-modal retrieval. The framework is a two-step hashing approach which decouples the optimization into (1) binary optimization and (2) hashing function learning. In the first step, we propose a novel spectral embedding-based algorithm to simultaneously learn single-modality and binary cross-modality representations. While the former is capable of well preserving the local structure of each modality, the latter reveals the hidden patterns from all modalities.
In the second step, to learn mapping functions from informative data inputs (images and word embeddings) to binary codes obtained from the first step, we leverage the powerful CNN for images and propose a CNN-based deep architecture to learn text modality.
Quantitative evaluations on three standard benchmark datasets demonstrate that the proposed DCSH method consistently outperforms other state-of-the-art methods.

\end{abstract}

\section{Introduction}
The last few years have witnessed 
an exponential surge in the amount of information available online in heterogeneous modalities, e.g., images, tags, text documents, videos, and subtitles. 
Thus, it is desirable to have a single efficient system that can facilitate large-scale multi-media searches. In general, this system should support both single and cross-modality searches, i.e., the system returns a set of semantically relevant results of all modalities given a query in any modality.
In addition, to be used in large scale applications, the system should have efficient storage and fast searching. To handle the above challenges, several cross-modality hashing approaches have been proposed, in both supervised \cite{CRH,HTH,SMH,QCH,SePH,DCH,DLFH,SMFH,8425016,8331146,THN,DCMH,DDCMH,CMDVH,PRDH,DVSH} and unsupervised \cite{CVH,PDH,IMH,CMFH,LSSH,ACQ,FSH,CRE,DMHOR,MSAE,CorrAE,UGACH,SCQ} learning. Furthermore, as the unsupervised hashing does not require any label information, it is suitable for large-scale retrieval problems 
with limited/scarce label information.
Thus, in this work, we focus on the unsupervised setting of the cross-modality hashing problem for retrieval tasks.

Recent unsupervised cross-modality hashing methods \cite{ACQ,FSH} jointly learn binary codes and linear hash functions. However, mapping data from very high-dimensional non-linear spaces to a common Hamming distance space using linear models is likely to underfit training data. 
Learning highly expressive non-linear models (e.g., CNN) with binary constraints on the model outputs in the unsupervised context is not a straightforward solution. It is non-trivial to obtain low-bit hash codes that preserve the structure of the high-dimensional inputs without any supervised information (i.e., semantically-similar inputs may be mapped to very dissimilar binary codes). 
To overcome this challenge, in this paper, we rely on the two-step hashing approach \cite{SelfTaughtHF,2stephash} for learning binary codes and hashing functions.
By decoupling the binary optimization and hashing function learning, the two-step approach helps to 
simplify both the problem and the optimization, which leads to better binary codes.

Particularly, in the first step, i.e., learning binary codes, it is essential to preserve the intra and inter-modality similarities jointly in a common Hamming space. To preserve the intra-modality similarity, previous unsupervised cross-modality hashing methods \cite{CVH,IMH} utilize the well-known spectral embedding-based approach to discover the neighborhood structure of data \cite{SpectralClusterTutorial}. 
Nevertheless, these methods produce different embedding spaces for different modalities, which is a non-optimal solution for the cross-modality retrieval problem \cite{ACMR,CMFH}. To overcome this drawback, we propose a spectral embedding-based approach to learn a {joint binary representation} that is well represented for all modalities simultaneously. We additionally include the independence and balance constraints on binary codes, i.e., different bits in the binary codes are independent from each other, and each bit has a $50\%$ chance of being $1$ or $-1$ \cite{SpH,5539994}. The independence property is to ensure hash codes do not capture redundant information, while the balance property is to ensure hash codes contain the maximum amount of information \cite{5539994}.

Furthermore, we note that the quality of data-to-data similarity matrix plays an important role in the success of the spectral clustering. 
The Anchor Graph \cite{AnchorGraph} is widely adopted in recent works \cite{DGH,SADH,FSH} to construct the similarity matrices. 
This method, however, only considers the data-to-anchor connection, while the anchor-to-anchor connections, which potentially contain some useful information, are not put into consideration.
To achieve better graph representations, we propose to further improve the Anchor Graph by introducing the anchor-to-anchor mapping.

\begin{figure*}[th]
\centering
\includegraphics[width=0.9\textwidth]{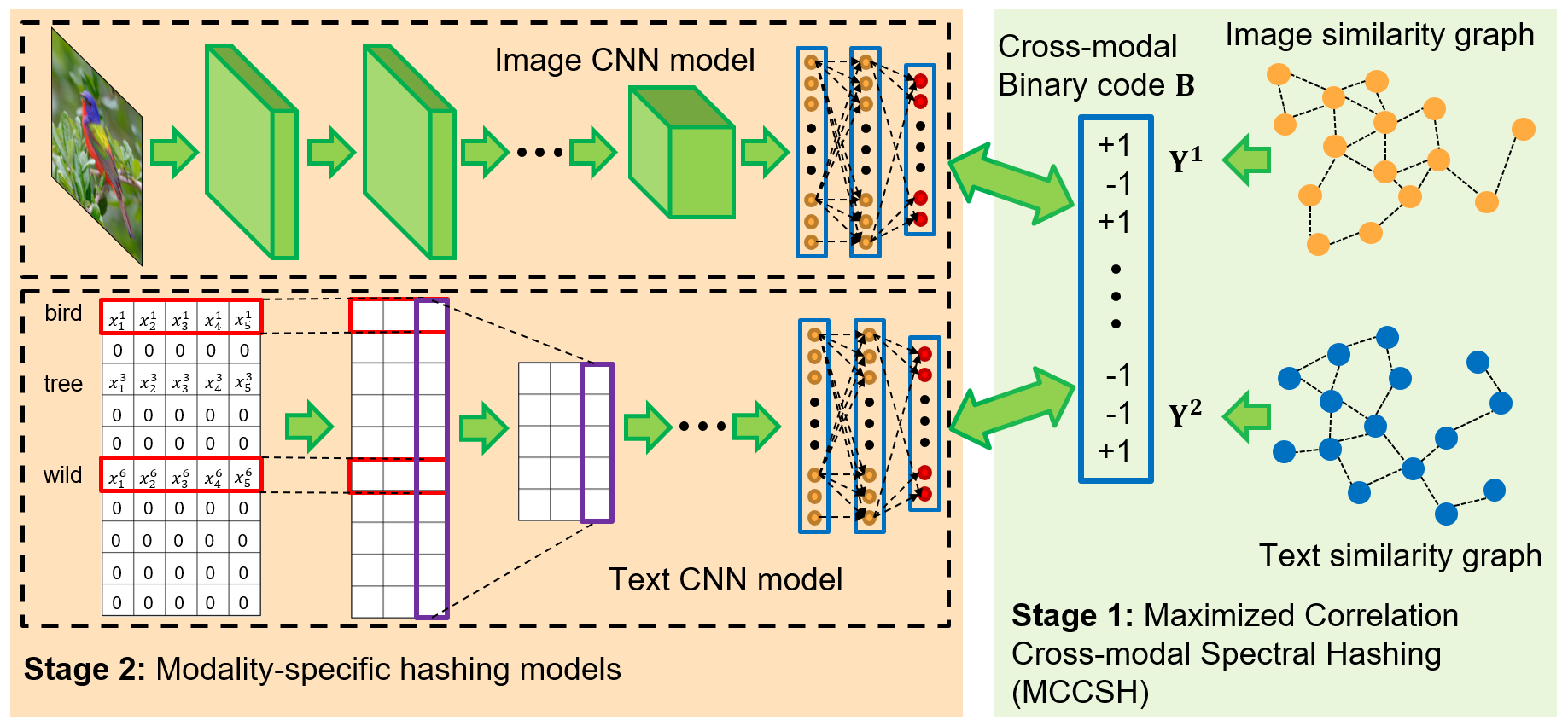}
\caption{The overview of our proposed method: Deep Cross-modality Spectral Hashing (DCSH). Our method includes the following two stages. In the first stage, we learn the cross-modality binary representation $\bB$ that can represent the underlying local structure of all modalities well. In the second stage, given the cross-modality binary codes, we train deep models to map informative inputs (e.g., images, word embeddings) to their corresponding binary codes.}
\label{fig:overview}
\end{figure*}

The second step of the two-step hashing approach is to learn hash functions to map inputs to the learned binary codes in the first step.  
In this step, since there is no challenging requirement as in the first step (i.e., preserving the intra and inter-modality similarities jointly in a common Hamming space, independent and balanced hash codes), any deep-model can be chosen to fit the learned binary codes easily. Specifically, for image modality, we leverage the powerful CNN to learn hash functions from input images as this would help to reduce the bias from the classification task, on which the CNN model are pre-trained, i.e., some information of input images is discarded if it is irrelevant to classification labels.
{For text modality, instead of using the limited-information bag-of-words (BOW) textual features as previous cross-modality hashing works \cite{PDH,IMH,CMFH,LSSH,ACQ,FSH,CRE}, we propose a CNN-based deep architecture to learn the binary representation from a set of word embeddings (named as document embeddings). In fact, the textual BOW is commonly constructed from most frequent words which appear in the entire training set. However, the BOW textual features do not put the meaning of words into consideration. For example, many words having similar meaning are considered as completely different words (e.g., \textit{tree} and \textit{plant}, \textit{street} and \textit{road}, \textit{women} and \textit{female}). Hence, using word embeddings can help to mitigate this issue as word meanings are put into consideration (e.g., words with similar meaning tend to have similar embeddings) \cite{glove,word2vec}.}

In summary, in this paper, by adopting the spectral embedding with the two-step hashing, we propose a novel framework, dubbed {\textit{Deep Cross-modality Spectral Hashing (DCSH)}}. Our main contributions are:

\begin{itemize}
    \item We introduce the {anchor-to-anchor mapping} which provides datapoints broader views of local structures, i.e., capturing the manifold structures of datasets better. The proposed anchor-to-anchor mapping is novel and robust. {In section \ref{sec:ablation_anchor2anchor}, we empirically show that it is beneficial in building more informative anchor graphs for learning binary codes.}
    \item In the first stage of the two-step hashing, we propose a novel algorithm which is based on spectral embedding to simultaneously learn single modality representations and a cross-modality binary representation. The former presents the underlying structure of each individual modality, while the latter captures the common hidden patterns among all modalities.
    {
    \item In the second stage of the two-step hashing, for the text modality, we propose to leverage the word embeddings \cite{glove,word2vec} to obtain more informative input for the text modality (named as document embeddings), instead of using the limited-information bag-of-words (BOW) textual features as previous cross-modality hashing works \cite{PDH,IMH,CMFH,LSSH,ACQ,FSH,CRE}. We then propose a CNN-based deep architecture to learn the binary representation from the informative document embeddings. For the image modality, learning hash functions directly from images also help to reduce the bias from the classification task, on which the CNN model are pretrained.
    }
    \item We compare our proposed method against various state-of-the-art unsupervised cross-modality hashing methods on three standard cross-modal benchmark datasets, i.e., MIR-Flickr25K, IAPR-TC12, and NUS-WIDE. The quantitative results clearly justify our contributions and demonstrate that our proposed method outperforms the compared methods on various evaluation metrics. 
\end{itemize}

\noindent The overview of the proposed method is presented in Fig. \ref{fig:overview}.
\section{Related work}
\label{ssec:related_works}
In this section, we first give a brief overview on hashing methods. We then focus on works that relate to our method, i.e., supervised and unsupervised cross-model hashing.

\textbf{Single modal hashing:}
Existing single modal hashing methods can be categorized as data-independent and data-dependent schemes \cite{DBLP:journals/corr/WangLKC15,7915742,Grauman_review}. 
Data-independent hashing methods~\cite{lsh_vldb09,KLSH_iccv09,KLSH_nips09,DBLP:journals/pami/KulisJG09} rely on random projections for constructing hash functions. Although representative data-independent hashing methods such as Locality-Sensitive Hashing (LSH)~\cite{lsh_vldb09} and its kernelized versions~\cite{KLSH_iccv09,KLSH_nips09} have theoretical guarantees which show that similar data have a higher probability to be mapped into similar binary codes. They require long codes to achieve high precision. 
Different from data-independent approaches, data-dependent hashing methods use available training data for learning hash functions in an unsupervised or supervised manner and they usually achieve better retrieval results than data-independent methods. The unsupervised hashing methods~\cite{SpH,ITQ,DBLP:conf/cvpr/HeWS13,BA,CVPR12:SphericalHashing,DeepHash_TIP17,deepbit2016,8801918,8709820,SCQ} try to preserve the neighbor similarity of samples in the Hamming space without label information. The representative unsupervised hashing methods are Iterative Quantization (ITQ)~\cite{ITQ}, Spherical Hashing (SPH)~\cite{CVPR12:SphericalHashing}, and K-means Hashing (KMH)~\cite{DBLP:conf/cvpr/HeWS13}. 
The supervised hashing methods~\cite{Kulis_learningto,CVPR12:Hashing,CVPR2014Lin,Shen_2015_CVPR,DBLP:conf/icml/NorouziF11} try to preserve the label similarity of samples using labeled training data. The representative supervised hashing methods are Kernel-Based Supervised Hashing (KSH)~\cite{CVPR12:Hashing}, Semi-supervised Hashing (SSH)~\cite{DBLP:journals/pami/WangKC12}, and Supervised Discrete Hashing (SDH)~\cite{Shen_2015_CVPR}. 
The readers can refer to~\cite{DBLP:journals/corr/WangLKC15} and \cite{7915742} for extensive reviews on single model hashing research. 

Recent works show that joint learning image representations and binary hash codes in an end-to-end deep learning-based supervised hashing framework~\cite{DSRH,DRSCH,simulfeature,8451192,DSH,DHN,DPSH,CauchyHashing} have demonstrated a considerable boost in retrieval accuracy. By joint optimization, the produced hash codes are more sufficient to preserve the semantic similarity between images. In those works, the network architectures often consist of a feature extraction sub-network and a subsequent hashing layer to produce hash codes. 

Most aforementioned hashing methods are designed for a single modality input. However, recent years have witnessed an exponential surge in the amount of information available online in heterogeneous modalities, e.g., images, tags, text documents, videos, and subtitles. Thus, it is desirable to have a single efficient system that can facilitate large-scale multi-media searches. 

\textbf{Supervised Cross-modal hashing:}
Supervised hashing methods can explore the semantic information to enhance the data correlation from different modalities (i.e., reduce modality gap) and reduce the semantic gap. Many supervised cross-modal hashing methods with shallow architectures have been proposed, for instance Co-Regularized Hashing (CRH) \cite{CRH}, Heterogeneous Translated Hashing (HTH) \cite{HTH}, Supervised Multi-Modal Hashing (SMH) \cite{SMH}, Quantized Correlation Hashing (QCH) \cite{QCH}, Semantics-Preserving Hashing (SePH) \cite{SePH}, Discrete Cross-modal Hashing (DCH) \cite{DCH}, and Supervised Matrix Factorization Hashing (SMFH) \cite{SMFH}. 
All of these methods are based on hand-crafted features, which cannot effectively capture heterogeneous correlation between different modalities and may therefore result in unsatisfactory performance. Unsurprisingly, recent deep learning-based works \cite{THN,DCMH,DDCMH,CMDVH,SSAH,PRDH,DVSH,8331146} can capture heterogeneous cross-modal correlations more effectively.
Deep cross-modal hashing (DCMH) \cite{DCMH} simultaneously conducts feature learning and hash code learning in a unified framework. 
Pairwise relationship-guided deep hashing (PRDH) \cite{PRDH}, in addition, takes intra-modal and inter-modal constraints into consideration. 
Deep visual-semantic hashing (DVSH) \cite{DVSH} uses CNNs, long short-term memory (LSTM), and a deep visual semantic fusion network (unifying CNN and LSTM) for learning isomorphic hash codes in a joint embedding space. 
However, the text modality in DVSH is only limited to sequence texts (e.g., sentences). 
In Cross-Modal Deep Variational Hashing \cite{CMDVH}, given learned representative cross-modal binary codes from a fusion network, the authors proposed to learn generative modality-specific networks for encoding out-of-sample inputs.
{In Cross-modal Hamming Hashing \cite{CMHH}, the author proposed Exponential Focal Loss which puts higher losses on pairs of similar samples with Hamming distance much larger than 2 (in comparison with the sigmoid function with the inner product of binary codes). 
Mandal \textit{et al.} \cite{8425016} proposed Generalized Semantic Preserving Hashing (GSPH) which can work for unpaired inputs (i.e., given a sample in one modality, there is no paired sample in other modality.).}

Although supervised hashing typically achieves very high performance, it requires a labor-intensive process to obtain large-scale labels, especially for multi-modalities, in many real-world applications.
In contrast, unsupervised hashing does not require any label information. Hence, it is suitable for large-scale image search in which the label information is usually unavailable.

\textbf{Unsupervised Cross-modal hashing:}
Cross-view hashing (CVH) \cite{CVH} and Inter-Media Hashing (IMH) \cite{IMH} adopt Spectral Hashing \cite{SpH} for the cross-modality hashing problem. These two methods, however, produce different sets of binary codes for different modalities, which may result in limited performance.
In Predictable Dual-View Hashing (PDH) \cite{PDH}, the authors introduced the {predictability} to explain the idea of learning linear hyper-planes each of which divides a particular space into two subspaces represented by $-1$ or $1$. The hyper-planes, in addition, are learned in a self-taught manner, i.e., to learn a certain hash bit of a sample by looking at the corresponding bit of its nearest neighbors. 
Collective Matrix Factorization Hashing (CMFH) \cite{CMFH} aims to find consistent hash codes from different views by collective matrix factorization. 
Also by using matrix factorization, Latent Semantic Sparse Hashing (LSSH) \cite{LSSH} was proposed to learn hash codes in two steps: first, latent features from images and texts are jointly learned with sparse coding, and then hash codes are achieved by using matrix factorization.
Inspired by CCA-ITQ \cite{ITQ}, Go \textit{et al.} \cite{ACQ} proposed Alternating Co-Quantization (ACQ) to alternately minimize the binary quantization error for each of modalities. 
In \cite{FSH}, the authors applied Nearest Neighbor Similarity \cite{NSS} to construct Fusion Anchor Graph (FSH) from text and image modals for learning binary codes. 

{In addition to the aforementioned shallow methods, several works \cite{DMHOR,MSAE,CorrAE} utilized (stacked) auto-encoders for learning binary codes. These methods try to minimize the distance between hidden spaces of modalities to the preserve inter-modal semantic. They also used the reconstruction criterion for each modality to maintain the intra-modal consistency. However, the reconstruction criterion is not a direct way for preserving similarity \cite{BA,UH-BDNN}, and may lead to non-optimal binary codes. 
Recently, Zhang \textit{et al.} \cite{UGACH} designed a graph-based unsupervised correlation method to capture the underlying manifold structure across different modalities, and a generative adversarial network to learn the manifold structure. Taking a different approach, we adopt a spectral embedding-based approach which can effectively capture the local structure of the input datasets.
Wu \textit{et al.} \cite{UDCMH} proposed  Unsupervised Deep Cross Modal Hashing (UDCMH), which alternatives between feature learning,
binary latent representation learning and hash function learning.  
Su \textit{et al.} \cite{DJSRH} proposed a joint-semantics affinity matrix, which integrates the neighborhood information of two modals, for mini-batch samples to train deep network in an end-to-end manner. This method is named as Deep Joint Semantics Reconstructing Hashing (DJSRH).
On the contrary, we decouple the learning of binary codes and hashing function. This results in simpler optimization and better binary codes.
Additionally, inspired by Anchor Graph \cite{AnchorGraph}, we also propose a novel method to achieve more informative similarity graphs for the multi-modal context. Moreover, both UDCMH \cite{UDCMH} and DJSRH \cite{DJSRH} are only proposed for two modalities, and their abilities to adapt to the general case of $M$ modalities are unclear, while our method is proposed for $M$ modalities.
In Cycle-Consistent Deep Generative Hashing (CYC-DGH) \cite{CYC-DGH}, the authors leverage the cycle-consistent loss \cite{CycleGAN2017} to learn hashing functions that map inputs of different modalities into a common hashing space without requiring paired inputs. However, this setting is not the main focus of our work, and we leave it for future study.
}

\section{Preliminary works}
\textbf{Spectral Clustering} 
is one of the most popular clustering algorithms due to its simple implementation and effectiveness in exploring hidden structures of the input data \cite{SpectralClusterTutorial}. The well-known formulation of spectral clustering of an undirected weighted graph $\bG = \{\bX; \bA \}$, with a vertex set $\bX\in\bbR^{N\times D}$ ($N$ data points of $D$-dimension) and an edge affinity matrix $\bA\in\bbR^{N\times N}$, is defined as follows:
\begin{equation}
\label{eq:spectral_clustering}
    \min_\bY~\tr\left(\bY\transpose\bL\bY\right),\quad\text{s.t. }\bY\transpose\bY = \bI_L,
\end{equation}
where $\bY\in\bbR^{N\times L}$ is an $L$-dimension spectral embedding ($L\ll D$) of the input data $\bX$, $\bL=\bD-\bA$ is the Laplacian graph matrix, and $\bD \in \bbR^{N\times N}$ is a diagonal matrix with $\bD_{ii} = \sum_{j=1}^N \bA_{ij}$. The solutions of \eqref{eq:spectral_clustering} are simply the $L$ eigenvectors corresponding to the $L$ smallest positive eigenvalues of $\bL$. 
{The clusters are obtained by conduct $k$-means on the $L$-dimension spectral embedding.}

\textbf{Anchor Graph} 
\cite{DGH} uses a small set of $P$ points (called \textit{anchors}) $\bU=\{\bu_i\}_{i=1}^P\in\bbR^{P\times D}$ to approximate the neighborhood structure underlying the input dataset $\bX =\{\bx_i\}_{i=1}^N\in \bbR^{N\times D}$ ($N \gg P$). $\bU$ is commonly achieved by using lightweight clustering methods such as $k$-means, DeepVQ \cite{DeepVQ}.
Given a kernel function $\mathcal{K}(\bx_1, \bx_2)$, e.g., Gaussian Kernel $\mathcal{K}(\bx_1, \bx_2) = \exp\left(-{\|\bx_1- \bx_2\|^2}/{\sigma}\right)$ with the bandwidth parameter $\sigma$, the nonlinear data-to-anchor mapping ($\bbR^D\mapsto\bbR^P$) is defined as follows:
\begin{equation}
z(\bx_i)=\left[\delta_1\mathcal{K}(\bx_i, \bu_1), \dots, \delta_P\mathcal{K}(\bx_i, \bu_P)\right]^\top\hspace{-0.2em}/N_{l1},     
\end{equation}
where $\delta_j\in\{0, 1\}$ and $\delta_j=1$ if and only if anchor $\bu_j$ is one of the $k\ll P$ closest anchors of $\bx_i$ in $\bU$,
 and $N_{l1}=\sum_{j=1}^P\delta_j\mathcal{K}(\bx_i, \bu_j)$.
Then, the data-to-data affinity matrix is achieved as $\bA=\bZ\mathbf{\Lambda}^{-1}\bZ^\top$, where $\bZ=\left[z(\bx_1), \dots, z(\bx_n)\right]^\top$ is the data-to-anchor affinity matrix and $\mathbf{\Lambda}=\text{diag}(\bZ\transpose\bOnes)$. 
Each element of $\bA$ can be probabilistically interpreted as the transition probability from one data point to another. 
\section{Proposed Method}
\label{sec:proposed_method}

\setlength{\textfloatsep}{5pt}
\begin{algorithm}[t]
\caption{Deep Cross-modality Spectral Hashing (DCSH)}
\label{algo:main}
\SetKwData{Left}{left}\SetKwData{This}{this}\SetKwData{Up}{up}
\SetKwFunction{Union}{Union}\SetKwFunction{FindCompress}{FindCompress}
\SetKwInOut{Input}{Input}\SetKwInOut{Output}{Output}
\Input{Training data $\oset=\{\bo_i\}_{i=1}^N$; code length $L$;\\parameters $P, k, k_a, \lambda_1, \lambda_2, \alpha, \gamma_1, \gamma_2$}
\Output{Modality-specific hashing functions \hspace{-3pt}$\{\func^m\}_{m=1}^M$\hspace{-3pt}} 
\BlankLine
\tcc{I - Anchor Graph}
Learn anchor sets for modalities (Sec. \ref{sec:anchors});\\
Compute the \textit{expanded} data-to-anchor affinity matrices $\{\hat{\bZ}^m\}_{m=1}^M$  (Sec. \ref{sec:anchor2anchor});\\
Compute the Laplacian graphs: $\{\bL^m\}_{m=1}^M$ (Sec. \ref{ssec:related_works});\\
\tcc{II - Maximized Correlation Cross-modality Binary Codes}
Obtain cross-modality binary codes $\bB$ using Algo. \ref{algo:cross_modal_bin}; \\

\tcc{III - Hashing functions}
Train modality-specific hashing functions $\{\func^m\}_{m=1}^M$ with the objective function \eqref{eq:deep_model}, given $\bB$ from the above step;\\
\Return{$\{\func^m\}_{m=1}^M$.} 
\end{algorithm}


Given a dataset of $N$ instances including $M$ modalities, denoted as  $\oset=\{\bo_i\}_{i=1}^N$, $\bo_i=(\bx_i^1,\dots, \bx_i^M)$, where $\bx_i^m\in \bbR^{D^m}$ is the $D^m$ dimensional feature of $m$-th modality. 
We also denote the feature matrix for $m$-th modality as $\bX^m = \{\bx_i^m\}_{i=1}^N \in \bbR^{N\times D^m}$. We now delve into the details of our proposed method. The overview of the proposed method is presented in Fig. \ref{fig:overview} and Algorithm \ref{algo:main}.

\subsection{Anchor Graph}
\label{sec:anchor_graph}
Inspired by the anchor graph method \cite{AnchorGraph,DGH} for single modality data, we firstly propose to construct the anchor graphs for the multi-modal context. Furthermore, we propose to introduce the anchor-to-anchor mapping, 
which leverages the information among anchor points when constructing the anchor graph. 

\subsubsection{Anchors}
\label{sec:anchors}
By learning the anchor for each modality separately, the anchor sets lose the one-to-one relationship among modalities, i.e., there is no connection from an anchor point in a modality, e.g., image, to any anchor in another modality, e.g., text, as in the original datasets. Losing this property could lead to difficulty in modality fusion.


To handle this problem, we propose to learn the anchor sets for all modalities together as follows. \textit{(i)} Firstly, we normalize the data in $m$-th modality to have the unit-variance by dividing the data to the factor of $\sqrt{\sum_{i=1}^{D^m}\lambda_i^m}$, where $\lambda_i^m$ is the $i$-th eigenvalue of the covariance matrix of the $m$-th modality.
\textit{(ii)} Afterwards, we concatenate the datasets of all modalities together, i.e., $\bX\eq[\bX^1,\dots,\bX^M]\hspace{-0.2em}\in\hspace{-0.2em}\bbR^{N\times D}$, where $D\hspace{-0.05em}\eq\hspace{-0.05em}\sum_{m=1}^M D^m$. \textit{(iii)} We then apply $k$-means over $\bX$, which results in $P$ cross-modal centroids as $\bU\hspace{-0.1em}\eq\hspace{-0.1em}[\bU_1^\top, \cdots, \bU_P^\top]\transpose\hspace{-0.2em}\in\hspace{-0.2em}\bbR^{P\times D}$.
Note that the unit-variance normalization step (first step) is  important, as this helps to avoid $k$-means being biased toward modalities which have higher variances. We will analyze this aspect further in the experiment section (Section \ref{sec:ablation_norm}).
\textit{(iv)} Finally, we achieve the centroids for each modality by splitting each cross-modal centroid as $\bU_p\eq[\bu^1_p,\cdots, \bu^M_p]$ where $\bu^m_p\in\bbR^{D^m}$ is $p$-th centroid of $m$-th modality. 

Given the anchor sets for different modalities $\{\bU^m\}_{m=1}^M$, where $\bU^m\eq\{ \bu^m_1, \cdots,  \bu^m_P\}$, we can construct the data-to-anchor affinity matrix $\{\bZ^m\}_{m=1}^M$ as discussed in  Section \ref{ssec:related_works} - Anchor Graph.

\subsubsection{Anchor-to-Anchor mapping}
\label{sec:anchor2anchor}
In Anchor Graph \cite{AnchorGraph}, we can observe that the connections among anchors are not being considered. Nevertheless, these information can be  beneficial in helping data points to have broader views of data structures, especially for the cases of large clusters. In other words, the anchor-to-anchor mapping enhances the connectivity between data points of the same clusters. We note that by increasing the number of considered nearest anchors ($k$ in Section \ref{ssec:related_works}-Anchor graph) when computing data-to-anchor matrix, data points can also have broader views. However, the resulting graphs could be noisy. We will investigate this aspect in our experiments (Section \ref{sec:ablation_anchor2anchor}).

Hence, we propose to build the \textit{anchor-to-anchor} affinity matrix $\bS^m$ for the $m$-th modality.
Similar to the data-to-anchor affinity matrix $\bZ^m$, the nonlinear anchor-to-anchor mapping ($\bbR^P\mapsto \bbR^P$) for $m$-th modality is defined as:
\begin{equation}
s^m(\bu_i^m) = \left[\delta_1\mathcal{K}(\bu_i^m, \bu_1^m), \dots, \delta_P\mathcal{K}(\bu_i^m, \bu_P^m)\right]^\top\hspace{-0.2em}/N_{l1},\hspace{-0.5em}
\end{equation}
where $\delta_j\in\{0,1\}$ and $\delta_j=1$ if and only if anchor $\bu_j$ is the one of the $k_a\ll P$ \textit{reciprocal} (mutual) nearest anchors of $\bu_i$, and $N_{l1} = \sum_{j=1}^P\delta_j\mathcal{K}(\bu_i^m, \bu_j^m)$. Note that the \textit{reciprocal} nearest criterion is used to determine neighbor anchors to enhance the likelihood they belong to a same data manifold \cite{mutual_kmeans}. Consequently, we obtain the anchor-to-anchor affinity matrix for the $m$-th modality: $\bS^m=[s^m(\bu_1^m), \dots, s^m(\bu_P^m)]^\top$.
We now can achieve the {\textit{expanded} data-to-anchor affinity matrix by combining the data-to-anchor matrix $\bZ^m$ and the proposed anchor-to-anchor matrix $\bS^m$ as follows}:
\begin{equation}
    \hat{\bZ}^m = \bZ^m\bS^m,
\end{equation}
and subsequently compute the data-to-data affinity matrix $\bA^m$ and then the Laplacian graph $\bL^m$ as presented in Section~\ref{ssec:related_works}-Anchor graph. We note that the data-to-data affinity matrix $\bA^m$ computed from $\hat{\bZ}^m$ can still be interpreted as the probability transition matrix.
Given $\{\bL^m\}_{m=1}^M$, we now present the proposed two-step approach for learning the hash functions.

\begin{algorithm}[t]
\caption{Maximized Correlation Cross-modal Spectral Hashing (MCCSH)}\label{algo:cross_modal_bin}
\SetKwData{Left}{left}\SetKwData{This}{this}\SetKwData{Up}{up}
\SetKwInOut{Input}{Input}\SetKwInOut{Output}{Output}
\Input{$\{\bL^m\}_{m=1}^M$}
\Output{$\bB$}
\BlankLine
Initialize $\{\bY^m\}_{m=1}^M$ as in Sec. \ref{sec:init_Ym} (Algo. \ref{algo:init_Ym});\\
\Repeat{ \textit{converges}}{
\hspace{-8pt} Update $\bB$ with MPEC-EPM in Sec. \ref{sec:update_B}(Algo. \ref{algo:bin});\hspace{-15pt}\\
\hspace{-8pt} Update $\{\bY^m\}_{m=1}^M$ as in Sec. \ref{sec:update_Ym} (Algo. \ref{algo:al});\\
}
\Return{$\bB$.}
\end{algorithm}


\subsection{\textbf{Stage 1:} Maximized Correlation Cross-modal Binary Representation}
In this section, we  propose  a  novel spectral embedding-based method to learn a joint binary representation that is well represented for all modalities simultaneously.

Given the set of Laplacian graphs of different modalities $\{\bL^m\}_{m=1}^M$, 
we aim to simultaneously learn the compact $L$-bit joint binary representations $\bB=[\bb_1^\top, \cdots, \bb_N^\top]^\top\in\{-1,+1\}^{N\times L}$ for $N$ inputs, 
and set of representations for each modality $\{\bY^m\}_{m=1}^M$, in which $\bY^m\in\bbR^{N\times L}$. 
For convenience, we call $\bY^m$ as a \textit{single modality} spectral representation and $\bB$ as the \textit{cross-modality} spectral binary representation.

To naturally fuse heterogeneous modalities, we would like to simultaneously \textit{(i)} minimize the spectral clustering error of each modality, i.e., preserve intra-modality similarity,
and \textit{(ii)} maximize the correlation between the cross-modality binary spectral representation $\bB$ and each single modality spectral representation $\bY^m$.
This resulting $\bB$ is well represented for all the modalities.
We now formally propose our objective function as follows:
\begin{subequations}
\label{eq:bin_opt}
\begin{align}
    \min_{\{\bY^m\}_{m=1}^M, \bB}~~ &\sum\nolimits_{m=1}^M\mathcal{J}(\bB, \bY^m), \tag{\ref{eq:bin_opt}}\\
    \text{s.t.}\quad&\bB\in\{-1,+1\}^{N\times L}, \label{eq:bin_constraint}\\
                    &\bB^\top\bB=N\bI_L, ~~\bB^\top\bOnes_{N\times 1}=\bZeros_{L\times 1},\label{eq:independent_balance_constraints}\\
                    & (\bY^m)^\top \bY^m=N\bI_L,\quad \forall m, \label{eq:orthogonal_constraint}
\end{align}
where 
\begin{equation}
\label{eq:bin_opt_single_obj}
    \mathcal{J}(\bB, \bY^m)=\tr\left((\bY^m)^\top \bL^m\bY^m\right)-\alpha\tr\left(\bB^\top \bY^m\right).
\end{equation}
\end{subequations}
The parameters $\alpha$ controls the trade-off of the spectral clustering objective and the spectral representation correlation between the $m$-th modality spectral representation $\bY^m$ and the cross-modality spectral representation $\bB$. 
We constrain $(\bY^m)^\top \bY^m$ to be $N\bI_L$, instead of $\bI_L$, to make $\bY^m$ comparable with $\bB$. Besides the binary constraint \eqref{eq:bin_constraint}, we also enforce $\bB$ to be independent and balanced \eqref{eq:independent_balance_constraints}
\cite{DeepHash_TIP17,UH-BDNN,SpH,5539994}.

The problem \eqref{eq:bin_opt} is well-known to be NP-hard due to the binary constraint \eqref{eq:bin_constraint} and the orthogonal constraints \eqref{eq:independent_balance_constraints},\eqref{eq:orthogonal_constraint}. 
To handle this challenging problem \eqref{eq:bin_opt}, we propose the novel algorithm \ref{algo:cross_modal_bin},  to handle these two constraints alternatively, i.e., we iteratively optimize $\{\bY^m\}_{m=1}^M$ and $\bB$. Before going to details of solving each sub-problem, we first discuss the initialization, which is not only necessary to make the optimization procedure robust, but also leads to a better local minimum.

\subsubsection{Initialize $\{\bY^m\}_{m=1}^M$}
\label{sec:init_Ym}
We propose to initialize $\{\bY^m\}_{m=1}^M$ by considering different modalities individually. Each sub-problem becomes the well-known spectral clustering problem:
\begin{equation}
\label{eq:init_Ym}
    \min_{\bY^m}\tr\left((\bY^m)^\top\bL^m\bY^m\right),\quad\text{s.t.}~(\bY^m)^\top \bY^m=N\bI_L.
\end{equation}

The closed-form solution of \eqref{eq:init_Ym} can be simply obtained as $\hat{\bY}^m = \sqrt{N} \bE_L^m$, 
where $\bE_L^m$ is the $L$ eigenvectors corresponding to the $L$ smallest positive eigenvalues of $\bL^m$. Furthermore, we note that the closed-form solution is not the unique optimum for \eqref{eq:init_Ym}. In fact, by rotating $\hat{\bY}^m$ via an arbitrary orthogonal matrix $\bR^m$, the objective value of \eqref{eq:init_Ym} is unchanged. 
Therefore, we further propose to find an orthogonal rotation matrix for each modality such that the correlations between all pairs of rotated spectral representations are maximized. Equivalently, we have the following problem:
\begin{equation}
\label{eq:find_Rm}
\begin{split}
    \max_{\{\bR^m\}_{m=1}^M}~&\sum\nolimits_{\substack{m,t=1,\\m\neq t}}^M\tr\left((\bR^t)^\top(\hat{\bY}^t)^\top\hat{\bY}^m\bR^m\right),\\
    \text{s.t. }~&(\bR^m)^\top\bR^m =\bI_L, \forall m.
\end{split}
\end{equation}
This problem can be solved by alternatively updating each $\bR^m$. By fixing $\{\bR^t, \forall t\neq m\}$, the problem \eqref{eq:find_Rm} becomes:
\begin{equation}
\label{eq:find_single_Rm}
    \max_{\bR^m}~\tr\left((\hat{\bY}^\star)^\top\hat{\bY}^m\bR^m\right), \quad\text{s.t. }(\bR^m)^\top\bR^m =\bI_L,
\end{equation}
where $\hat{\bY}^\star = \sum_{t=1,t\neq m}^M\hat{\bY}^t\bR^t$. The problem \eqref{eq:find_single_Rm} is actually the classic Orthogonal Procrustes problem \cite{orthogonalProcrustes} and has a closed-form solution. 
In our case, the two point sets are the two spectral representation matrices, i.e., $\hat{\bY}^m$ and $\hat{\bY}^\star$. $\bR^m$ can be obtained as follows: first, compute the SVD of the $L\times L$ matrix $(\hat{\bY}^\star)^\top\hat{\bY}^m$ as $S\Omega S^\top$ and, then let $\bR^m = SS^\top$. We note that alternatively updating $\{\bR^m\}_{m=1}^M$ is guaranteed to converge: \textit{(i)} each sub-problem has the closed-form solution which ensures a non-decreasing objective value, \textit{(ii)} the objective value of \eqref{eq:find_Rm} is upper-bounded by $(_MC_2)NL$, where $(_MC_2)$ denotes $M$ combinations of 2.

Finally, we can achieve $\bY^m = \sqrt{N} \bE_L^m \bR^m$. The resulting $\bY^m$ is not only represented for the local-structure of each modality data well, but it also has the knowledge about the local-structure of data of all other modalities. 
More importantly, the highly pairwise-correlated $\{\bY^m\}_{m=1}^M$ are very beneficial for the learning $\bB$ step as common patterns among all single-modality representations are aligned. We present the algorithm for the $\{\bY^m\}_{m=1}^M$ initialization step in Algorithm \ref{algo:init_Ym}. 
{It is also worth noting that computing eigen-decomposition for $\bL^m$ to obtain $\bE_L^m$, requiring a computational complexity of $\mathcal{O}(N^3)$, could be a computational bottleneck. We note that a number of approximations to reduce the computational cost ($\mathcal{O}(LN^2)$) have been proposed in the literature \cite{Nystrom_spectal_clustering, power_method_spectal_clustering, minibatch_spectral_clustering}. However, errors could be accumulated as more eigen-vectors are required.
}

\begin{algorithm}[t]
\caption{$\{\bY^m\}_{m=1}^M$ Initialization}\label{algo:init_Ym}
\SetKwData{Left}{left}\SetKwData{This}{this}\SetKwData{Up}{up}
\SetKwInOut{Input}{Input}\SetKwInOut{Output}{Output}
\Input{$\{\bL^m\}_{m=1}^M$}
\Output{$\{\bY^m\}_{m=1}^M$}
\BlankLine
$\hat{\bY}^m=\sqrt{N}\bE^m_L, \forall m$; where $\bE_L^m$ are the $L$ eigenvectors corresponding to $L$ smallest positive eigenvalues of $\bL^m$;\\
$\bR^m=\bI_L, \forall m$;\\
\Repeat{\textit{converges}}{

\For{$m=1$ \KwTo $M$}{
$\hat{\bY}^\star = \sum_{t=1,t\neq m}^M\hat{\bY}^t\bR^t$;\\
Compute SVD of $(\hat{\bY}^\star)^\top\hat{\bY}^m$ as $S\Omega S^\top$;\\
$\bR^m = SS^\top$;\\
}

}
$\bY^m = \hat{\bY}^m\bR^m, \forall m$;\\
\Return{$\{\bY^m\}_{m=1}^M$.}
\end{algorithm}

\subsubsection{Fix $\{\bY^m\}_{m=1}^M$, update $\bB$}
\label{sec:update_B}

Given fixed $\{\bY^m\}_{m=1}^M$, the problem \eqref{eq:bin_opt} is still challenging as both the binary constraint \eqref{eq:bin_constraint} and the orthogonal constraint \eqref{eq:independent_balance_constraints} exist.
Similar to \cite{DGH,SADH}, we relax the independent and balance constraints \eqref{eq:independent_balance_constraints} by converting them into penalty terms. We reformulate the problem \eqref{eq:bin_opt} as follows:
\begin{subequations}
\label{eq:update_B}
\begin{equation}
    \min_\bB\mathcal{L}(\bB), \quad \text{s.t. }\bB\in\{-1,+1\}^{N\times L}, \tag{\ref{eq:update_B}}
\end{equation}
where 
\begin{equation}
\footnotesize
%
\hspace{-3pt}\mathcal{L}(\bB)=-\hspace{-2pt}\sum\limits_{m=1}^M\hspace{-1pt}\tr\left(\bB^\top\hspace{-1.5pt}\bY^m\right)
+\frac{\lambda_1}{4}\hspace{-1pt}\left\|\bB\hspace{-1pt}^\top\hspace{-1.5pt}\bB-N\bI_L\right\|_F^2+\frac{\lambda_2}{2}\hspace{-1.5pt}\left\|\bB^\top\hspace{-1pt}\bOnes_{N\times 1}\right\|_F^2,
\end{equation}
\end{subequations}
and $\lambda_1, \lambda_2$ are penalty parameters for the independent and balance terms respectively. We leverage the recent advanced optimization technique for the binary constraint, specifically the MPEC-EPM \cite{MPEC} (Mathematical Programming with Equilibrium Constraints - Exact Penalty Method).
We note that the proof of Lemma 3 in \cite{MPEC} is valid for both convex and non-convex objective functions. Hence, the MPEC-EPM method still guarantees to converge to a local \textit{binary} optimum for our problem \eqref{eq:update_B}. We present the detailed algorithm for solving problem \eqref{eq:update_B} using MPEC-EPM in Appendix \ref{sec:appendix_1_1}.
(Algorithm \ref{algo:bin}).


\subsubsection{Fix $\bB$, update $\{\bY^m\}_{m=1}^M$} 
\label{sec:update_Ym}

Given the fixed $\bB$, $\{\bY^m\}_{m=1}^M$ are independent and can be solved separately. The problem \eqref{eq:bin_opt} w.r.t. $\bY^m$ can be re-written as follows:
\begin{equation}
\label{eq:update_Ym}
\begin{split}
    \min_{\bY^m}~&\tr\left((\bY^m)^\top\bL^m \bY^m\right)-\alpha\tr(\bB^\top \bY^m), \\
    \text{s.t. }~&(\bY^m)^\top \bY^m=N\bI_L.
\end{split}
\end{equation}
This sub-problem \eqref{eq:update_Ym}, even though, no longer contains the binary constraint, it is still challenging due to the orthogonal constraint. To the best of our knowledge, there is no simple method for achieving a closed-form solution for $\bY^m$. We propose to directly solve the orthogonal constraint optimization problem \eqref{eq:update_Ym} using Augmented Lagrangian (AL) method. By introducing the Lagrange multipliers $\Gamma \in\bbR^{L\times L}$, we target to minimize the following \textit{unconstrained} AL function:
\begin{equation}
\label{eq:LA_loss}
    \mathcal{L}_{AL}(\bY^m,\Gamma,\mu)=\mathcal{J}(\bB, \bY^m)-\tr(\Gamma^\top \Phi)+\frac{\mu}{2}\|\Phi\|_F^2,
\end{equation}
where $\Phi = (\bY^m)^\top \bY^m-\bI_L$ and $\mu$ is a penalty parameter on the constraint. The AL algorithm for solving \eqref{eq:LA_loss} is presented in Algorithm
\ref{algo:al} (Appendix \ref{sec:appendix_1}). When $\mu$ is large, the constraint violation is severely penalized, thereby the minimal values of the AL function \eqref{eq:LA_loss} is forced to be closer to the feasible region of the original constrained function \eqref{eq:update_Ym}. Additionally, it has been theoretically shown in \cite{nocedal2006numerical} that the Lagrange multiplier $\Gamma$ is improved at every step of the algorithm, i.e., getting closer to the optimal multiplier $\Gamma^\star$. Hence, it is not necessary to increase $\mu \to +\infty$ to achieve a local optimum of \eqref{eq:update_Ym}.

\subsection{\textbf{Stage 2:} Modality-specific hashing models}
\label{sec:hash_model}
Given the binary code $\bB$ for the training data, we now learn the modality-specific hashing functions $\{\func^m\}_{m=1}^M$ which are used to produce binary codes for new data points. 
We propose to use the powerful architecture: Convolutional Neural Network (CNN) as the hash functions to directly learn the binary codes from more informative input data, instead of the extracted feature vectors.
Formally, we minimize the following objective function:
\begin{equation}
\label{eq:deep_model}
\begin{split}
    \min_{\{\bTheta^m\}_{m=1}^M}&\frac{1}{2}\sum\nolimits_{m=1}^M\|\func^m(\bX^m)-\bB\|_F^2 \\ 
    -\gamma_1&\sum\nolimits_{\substack{m,t=1,\\m\neq t}}^M \tr\left(\func^m(\bX^m)\transpose\func^t(\bX^t)\right)\\
    +\gamma_2&\sum\nolimits_{m,t=1}^M\left\|\func^m(\bX^m)\transpose\func^t(\bX^t) - N\bI_L\right\|_F^2.
\end{split}
\end{equation}
Here, $\bTheta^m$ is the parameters of the hashing function $\func^m$ for $m$-th modality. 
The first term of \eqref{eq:deep_model} is to minimize the discrepancy between outputs of the deep models and the learned binary codes $\bB$. The second is to maximize the correlations between outputs of all pairs of modalities. Hence, the binary codes from different modalities are learned to be matched with each others. The last term is to re-enforce the output hash codes to be independent. 

\textbf{For image modality}, we can naturally use images as inputs with any common CNN architecture, e.g., AlexNet \cite{alexnet}, VGG \cite{VGG}, and ResNet \cite{resnet}. Specifically, in this work, we use the pretrained VGG16 model \cite{VGG} as the based model. We then replace the last fully-connected layer of the VGG16 model, which is used for the classification task, by 2 layers: $[4096\to1024\to L]$. 

\textbf{For text modality}, firstly, we propose to define a textual input from a set of word embeddings, e.g., \cite{glove,word2vec}, denoted as a \textit{document embedding}, as follows: Given a dictionary $\mathcal{W}$ consisting of $|\mathcal{W}|$ words and their corresponding word embeddings $\{\bw_i\}_{i=1}^{|\mathcal{W}|}$, where
$\bw_i$ is a $D_w$-dimension word embedding of the $i$-th word in $\mathcal{W}$:
\begin{equation}
\label{eq:document_embedding}
    \bx_{r\_i}^t=[\delta_1\bw_1^\top,\dots, \delta_{|\mathcal{W}|}\bw_{|\mathcal{W}|}^\top]^\top \in \bbR^{|\mathcal{W}|\times D_w},
\end{equation}
where $[\delta_1, \dots, \delta_{|\mathcal{W}|}]$ is the BOW vector of the $i$-th document.

\begin{table}[t]
\begin{threeparttable}
\small
\centering
\setlength{\tabcolsep}{9pt}
\caption{The configuration of CNN for text modality.}
\label{tb:cnn_text}
\begin{tabular}{|c|l|l|l|}
\hline
\# & Layer & Kernel size &  Output size \\ \hline 
0 & Input & & $n\times |\mathcal{W}|\times d_w$  \\\hline
1 & 1D-Conv1 & $|\mathcal{W}|\times 512\times 1$ &  $n\times 512\times d_w$ \\
2 & Linear1 & $d_w\times 256$ &  $n\times 512\times 256$\\ \hline
3 & 1D-Conv2 & $512\times 256\times 1$ &  $n\times 256\times 256$ \\
4 & Linear2 & $256\times 128$ &  $n\times 256\times 128$ \\ \hline
5 & 1D-Conv3 & $256\times 128\times 1$ &  $n\times 128\times 128$ \\
6 & Linear3 & $128\times 64$ &  $n\times 128\times 64$ \\ \hline
7 & 1D-Conv4 & $128\times 64\times 1$ &  $n\times 64\times 64$ \\
8 & Linear4 & $64\times 32$ &  $n\times 64\times 32$ \\ \hline
9 & Flatten & & $n\times 2048$ \\ \hline
10 & Linear5 & $2048\times 1024$ & $n\times 1024$ \\
11 & Linear6 & $1024\times L$ & $n\times L$ \\ \hline
\end{tabular}
\begin{tablenotes}
\item\hspace{-1.1em} $\bullet$ $|\mathcal{W}|$ is the size of word dictionary, $d_w$ is the dimension of Glove word embedding \cite{glove}, and $n$ is the mini-batch size. 
\item\hspace{-1em} $\bullet$ For 1D-Conv layer, the kernel size values are respectively the number of input channels, the number of output channels, and the receptive field size; for Linear layer, the kernel size values are respectively the number of input features and the number of output features. 
\end{tablenotes}
\end{threeparttable}
\end{table}

Secondly, regarding the CNN model for text, unlike in images, there is no clear local spatial information in word embeddings. Hence, we propose to alternatively apply two convolutional operations with kernel sizes\footnote{Respectively for the first 2 layers only. Subsequent layers have different kernel sizes.} of $(|\mathcal{W}|\times 1)$ and $(1\times d_w)$ over document embedding data. By doing so, we alternatively consider each dimension of the word embeddings independently and consider all dimensions of the word embeddings jointly. This is similar to the spatial-separable convolution, except we include a non-linear activation, i.e., ReLU, after each convolutional operation to enforce the model to learn more complex mappings. 
The detailed configuration is shown in Table \ref{tb:cnn_text}. In addition, Figure \ref{fig:overview} can provide a better understanding about our proposed CNN architecture for text modality.
%
Please note that the main goal of this section is to introduce a CNN model to learn textual features from informative text inputs, i.e., sets of word embeddings or document embeddings. But how to design an optimal CNN for the same task is not a focus of this paper and will be leaved for future studies.

\smallskip
The models are trained using back-propagration. Finally, given the trained modality-specific hashing models, the binary code for a test sample of the corresponding modality can be obtained as $\text{sign}\left(\mathcal{F}^m(\bx^m)\right)$.

\section{Experiments}
In this section, we conduct a wide range of experiments to validate our proposed method on three standard benchmark datasets for the cross-model retrieval task, i.e., {MIR-Flickr25k} \cite{MIRFlickr25k}, {IAPR-TC12}, and {NUS-WIDE} \cite{nuswide}. 

\subsection{Experiment settings}
\label{sec:exp_settings}
\textbf{Datasets:} The \textbf{MIR-Flickr25K} dataset\footnote{https://press.liacs.nl/mirflickr/} \cite{MIRFlickr25k} is collected from Flickr website, which contains 25,000 images together with 24 provided labels. 
For the text features, we preprocess the textual tags by firstly \textit{(i)} removing stop words, then \textit{(ii)} selecting the tags that appear at least 20 times, and finally \textit{(iii)} removing tags that are not English words\footnote{Words are not in the 400,000 word dictionary of Glove \cite{glove}.}. Finally, we obtain 1,909-dimension BOW textual features.
We randomly select 2,000 image-text pairs as the query set, and the remaining as the training set and the database.

The \textbf{IAPR-TC12} dataset\footnote{https://www.imageclef.org/photodata} contains 20,000 images with corresponding sentence descriptions. These images are collected from a wide variety of domains, such as sports, actions, people, animals, cities, and landscapes. Following the common practice in \cite{DVSH,CMDVH}, we select the subset of the top 22 frequent labels from the 275 concepts obtained from the segmentation task. 
Similar to the MIR-Flickr25K dataset, we preprocess the description sentences by removing stop words, selecting the tags that appear at least 10 times, and removing non English words. As a result, we obtain 1,275-dimension BOW textual features.
We randomly select 2,000 pairs for the query set and the remaining data is used as the training set and the database.

The \textbf{NUS-WIDE} dataset\footnote{https://lms.comp.nus.edu.sg/research/NUS-WIDE.htm} \cite{nuswide} is a multi-label image dataset crawled from Flickr, which contains 296,648 images with associated tags. Each image-tag pair is annotated
with one or more labels from 81 concepts. 
Following the common practice \cite{FSH,DDCMH}, we select image-tag pairs which have at least one label belonging to the top 10 most frequent concepts. 
In this dataset, each tag is represented by an 1,000-dimension preprocessed BOW feature.
We randomly choose 200 image-tag pairs per label as the query set of total 2,000 pairs, and the remaining as the database. We then randomly sample 20,000 pairs from the database to form the training set.

For images, we extract FC7 features from the PyTorch pretrained VGG-16 network \cite{VGG}, and then apply PCA to compress to 1024-dimension.

\textbf{Compared Methods and Evaluation Metrics:}  
We compare our proposed method DCSH against five other state-of-the-art cross-modal hashing methods, i.e., Cross-View Hashing (CVH) \cite{CVH}, Predictable Dualview Hashing (PDH) \cite{PDH}, Collective Matrix Factorization Hashing (CMFH) \cite{CMFH}, Alternating Co-Quantization (ACQ) \cite{ACQ}, and Fusion Similarity Hashing (FSH) \cite{FSH}. The evaluations are presented in four tasks: \textbf{(1)} Img $\to$ Txt,  \textbf{(2)} Txt $\to$ Img, \textbf{(3)} Img $\to$ Img, and \textbf{(4)} Txt $\to$ Txt; i.e., images (Img)/texts (Txt) are used as queries to retrieve image/text database samples accordingly.
In addition, the quantitative performance is evaluated by the standard evaluation metrics \cite{graphHashing,BA,UH-BDNN,SSDH}: \textit{(i)} mean Average Precision (\mAP) and \textit{(ii)} precision of Hamming radius $\le 2$ (\textbf{\textit{Prec@R$\le$2}}) which measures precision on retrieved images/texts with a Hamming distance to query $\le 2$ (note that we report zero precision in the case of no satisfactory result).
The image-text pairs are considered to be similar if they share at least one common label and be dissimilar otherwise.

\textbf{Implementation Details:}
For fair comparison, we use the extracted features, which are used as the input for compared methods, to construct anchor graphs. 
Note that the extracted features are used to build the anchor graphs only, we use images and document embeddings as inputs for the CNN models (Sec. \ref{sec:hash_model}) to produce binary codes.
For the anchor graph, we set $P\eq 500, k\eq 3$ and $k_a\eq 2$. We learn $\sigma$ by the mean squared distances to $k$-nearest neighborhoods. In addition, we empirically set $\lambda_1\eq\lambda_2\eq 1, \gamma_1\eq\gamma_2\eq 100$, and $\alpha\eq 1$ by cross-validation.

\subsection{Ablation studies}
\label{ssec:ablation}

{\subsubsection{The necessity of unit-variance normalization before learning anchor sets}
\label{sec:ablation_norm}
Table \ref{tb:analysis_unit_normalize} shows an analysis on Mean Squared Error (MSE) of reconstruction MIR-Flickr25k dataset by its $P=500$ anchors\footnote{$MSE=\frac{1}{N}\sum_{i=1}^N\|\bx_i-\bu_i\|^2$ 
where $\bu_i$ is the nearest anchor of $\bx_i$.} in the cases of learning anchors with/without the unit-variance normalization step
(Sec. \ref{sec:anchor_graph}). 
The experimental results show that without the normalization process, $k$-means severely biases toward the image modality. In specific, there is almost no-change in MSE of the image modality when learning the anchor set independently or jointly with textual modality, while the MSE of the text modality changes significantly. 
Regarding the case with normalized data, the changes in MSE of image and textual modalities are comparable. 
In other words, unit-variance normalization can help to avoid $k$-means being biased toward any modality in learning anchors. As a result, the anchor sets well approximate the local structures of all modalities, and the anchor graphs can capture the manifold structures of the dataset \cite{AnchorGraph}.}

\begin{table}[t]
\small
\centering
\def\arraystretch{1.1}
\setlength{\tabcolsep}{9pt}
\caption{Analysis of the necessity of unit-variance normalization before learning anchors by measuring the change in MSE for MIR-Flickr25k dataset. \textbf{I}: learn anchors independently. \textbf{J}: learn anchors jointly by concatenating datasets. \textbf{G}: the increasing gap (in percentage) of MSE when learning anchors independently and jointly ($\text{MSE}_G=(\text{MSE}_J-\text{MSE}_I)/\text{MSE}_I$).}
\label{tb:analysis_unit_normalize}
\begin{tabular}{|c|c|c|c|c|}
\hline
{Setting} & \multicolumn{2}{c|}{Normalized} & \multicolumn{2}{c|}{\textbf{NOT} normalized}  \\ \hline
& Img & Txt & Img & Txt \\ \hline 
I & 0.6759 & 0.6416 & 715.8516 & 0.5954 \\ \hline
J & 0.8088 & 0.6974 & 714.3574 & 0.8393 \\  \hline
G & \textbf{19.66\%} & \textbf{8.70\%} & \textbf{-0.13\%} & \textbf{47.3\%} \\ \hline
\end{tabular}
\end{table}

\begin{figure}[t]
\centering
\includegraphics[width=0.25\textwidth]{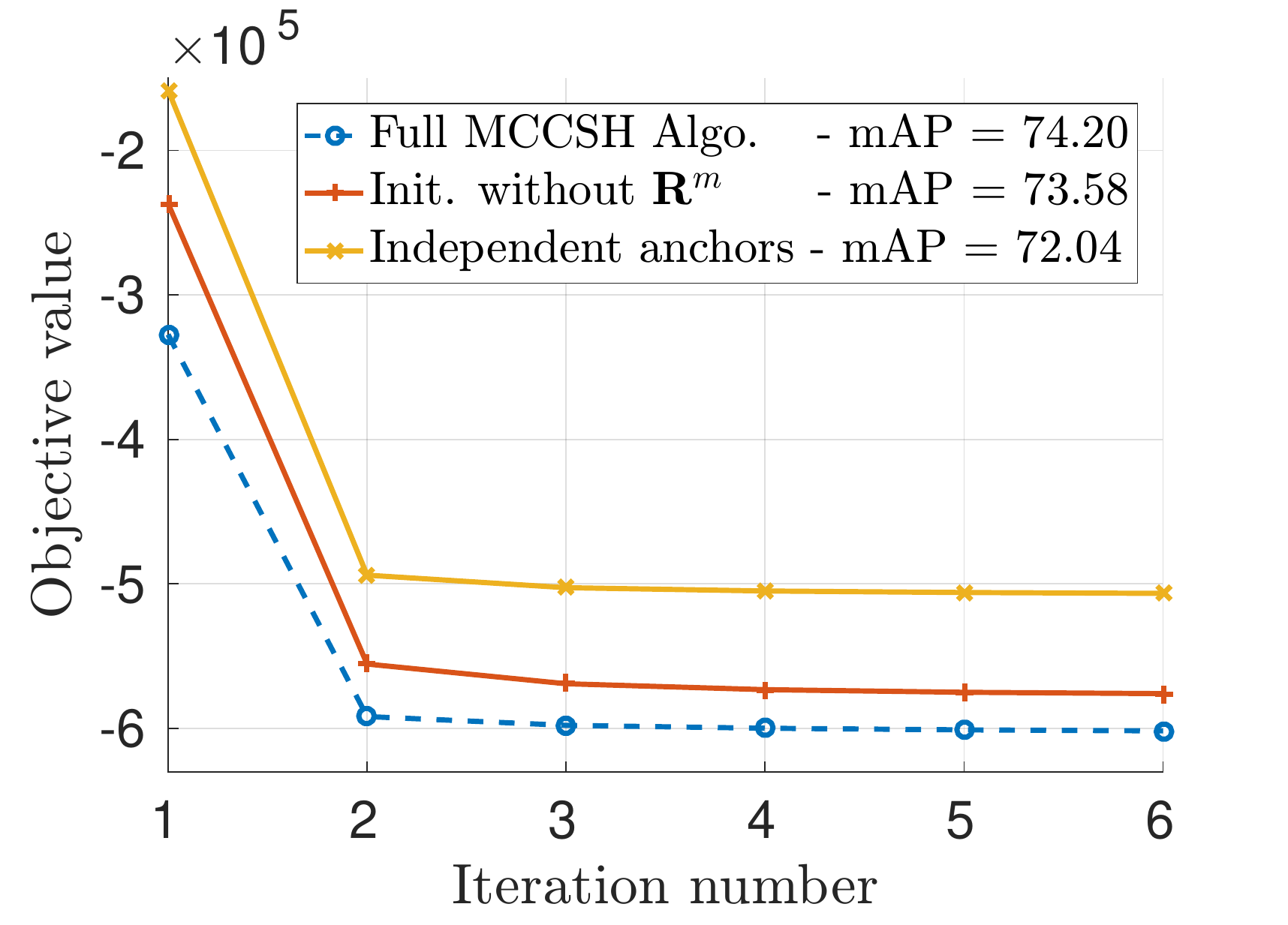}
\caption{MCCSH Algorithm analyses: convergence curves in several settings using MIR-Flickr25k dataset at $L=32$. The average \textit{mAP}s for Img $\to$ Txt and Txt $\to$ Img retrieval tasks are also presented in the legend.
}
\label{fig:analysis}
\end{figure}



\begin{figure*}[h]
\centering
\includegraphics[width=0.25\textwidth]{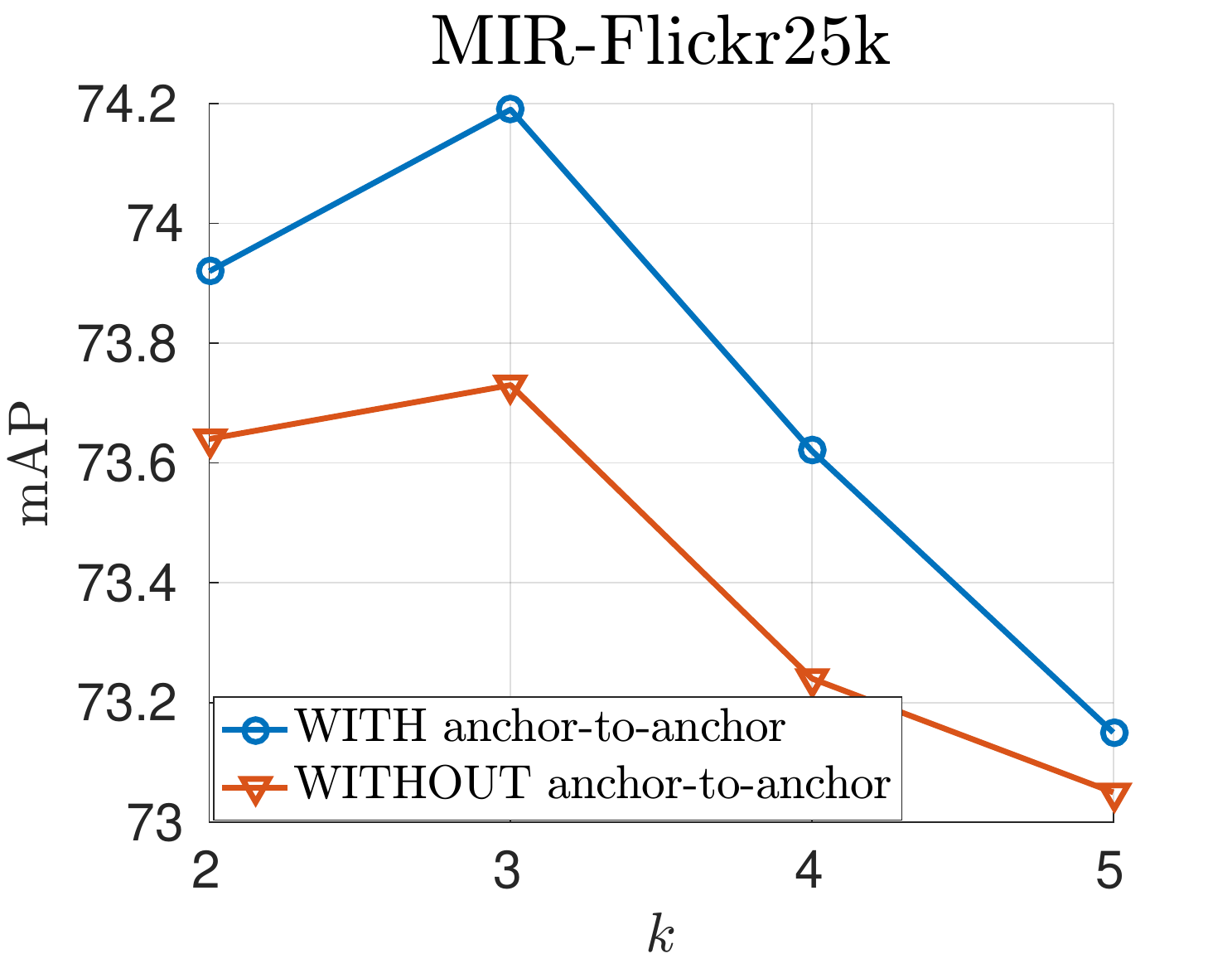}
~
\includegraphics[width=0.25\textwidth]{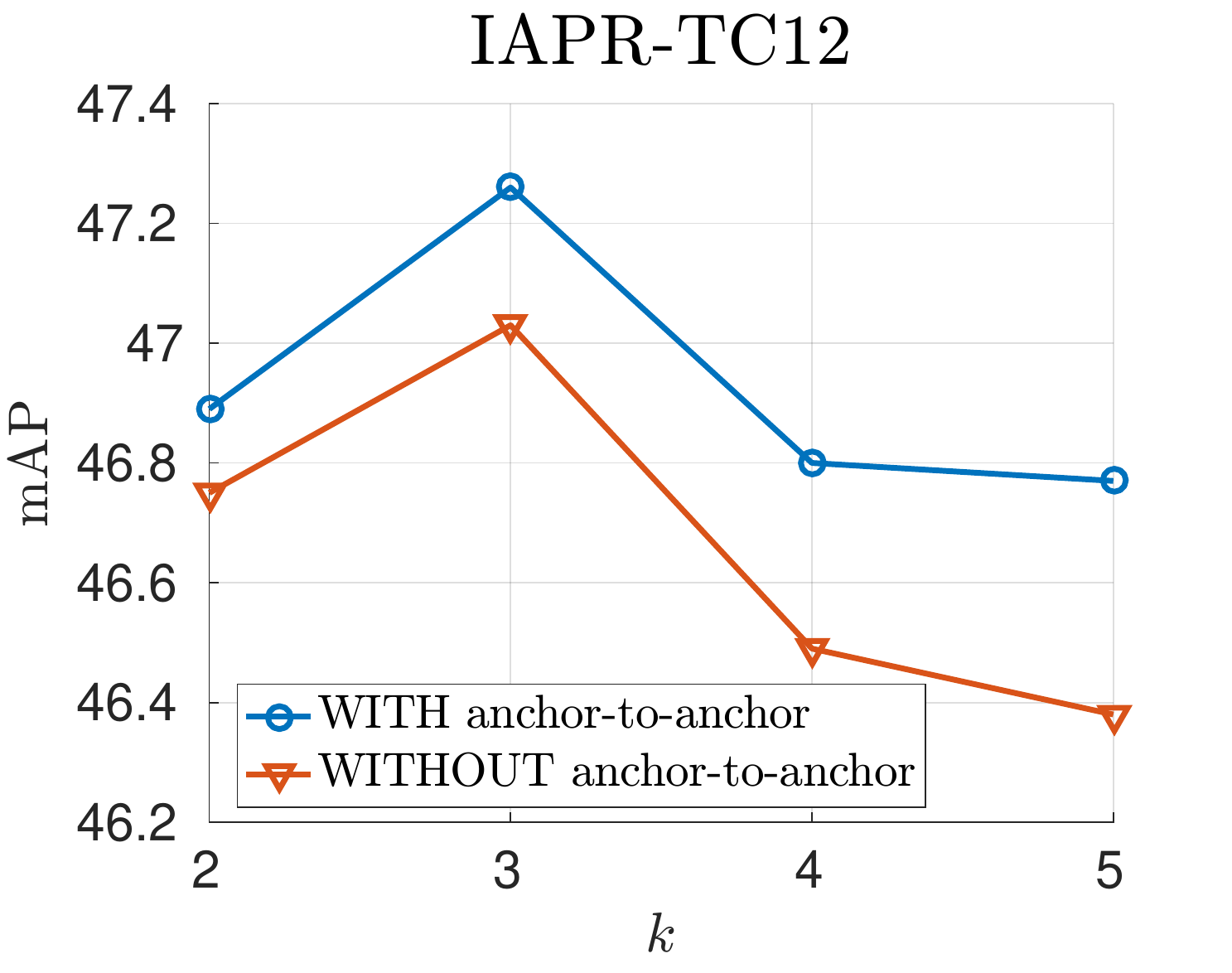}
~
\includegraphics[width=0.25\textwidth]{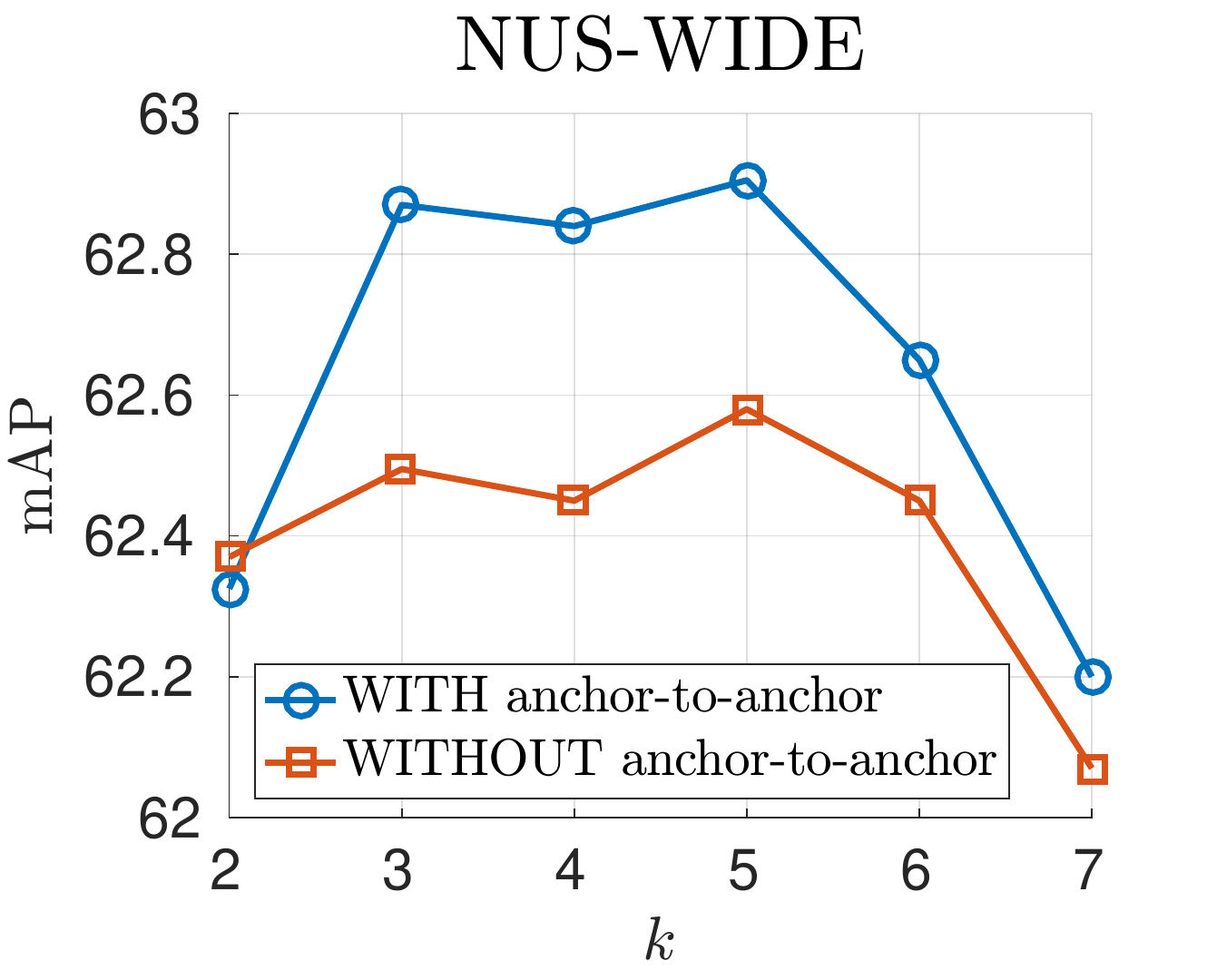}

\caption{The effect of including anchor-to-anchor mapping ($k_a=2$) on MIR-Flickr25k, IAPR-TC12, and NUS-WIDE datasets at $L=32$. The y-axis \textit{mAP} is the \textit{average} of \textit{mAP} for Img $\to$ Txt and Txt $\to$ Img retrieval tasks.}
\label{fig:anchor_to_anchor}
\end{figure*}



\begin{figure*}[h]
\centering
\includegraphics[width=0.25\textwidth]{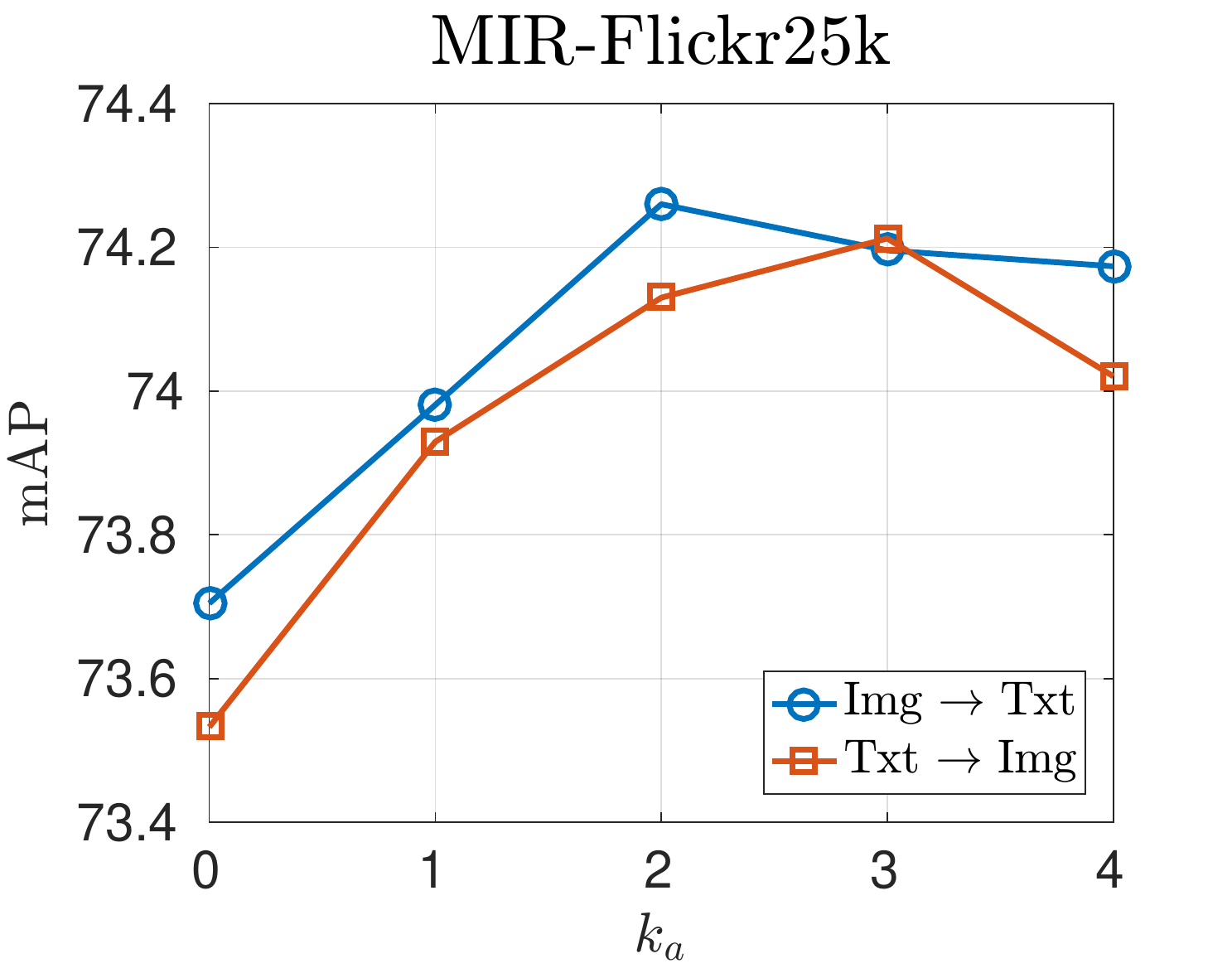}
~
\includegraphics[width=0.25\textwidth]{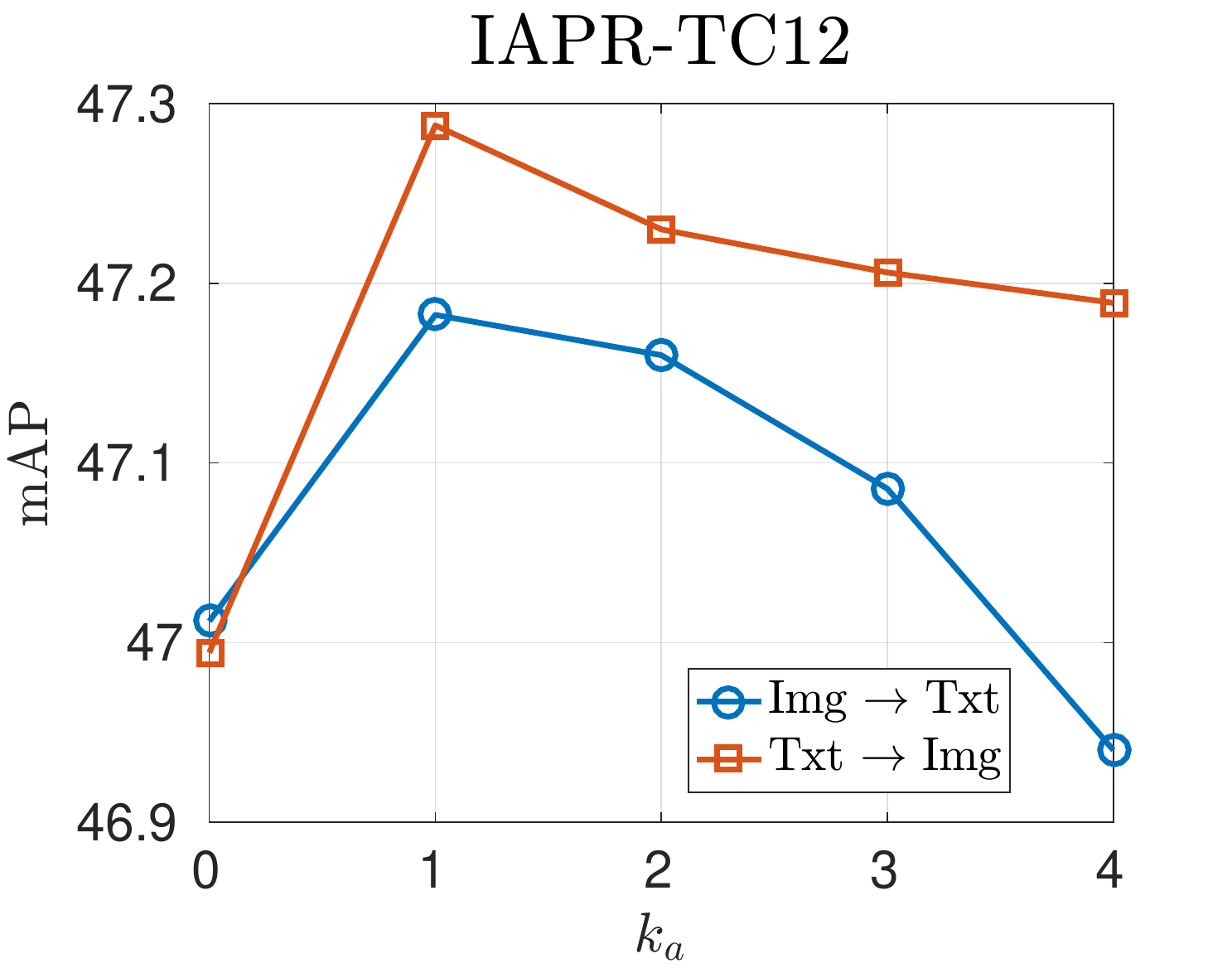}
~
\includegraphics[width=0.25\textwidth]{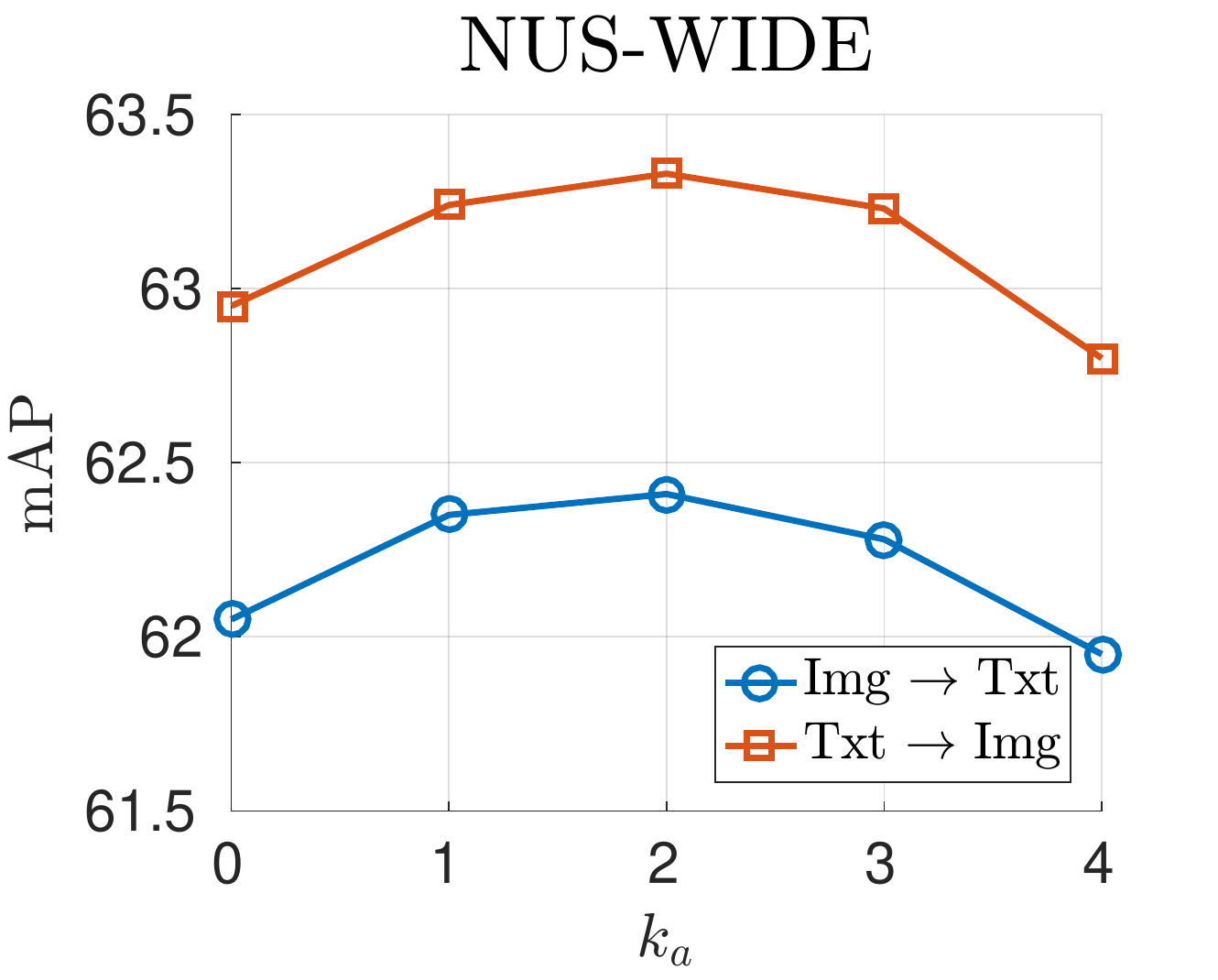}
\caption{The effect of varying $k_a$ anchor-to-anchor mapping(with fixed $k= 3$) on MIR-Flickr25k, IAPR-TC12, and NUS-WIDE datasets at $L=32$. $k_a=0$ means no anchor-to-anchor mapping is applied.
}
\label{fig:ka}
\end{figure*}

\subsubsection{Algorithm analysis}
\label{sec:algo_analysis}
We would like to conduct experiments to empirically analyze our proposed Algorithm \ref{algo:cross_modal_bin}.
Fig. \ref{fig:analysis} shows an example of the convergence of Algorithm \ref{algo:cross_modal_bin} using MIR-Flickr25k dataset
at $L\eq 32$
with different settings: \textit{(i)} learning anchors independently, \textit{(ii)} without learning the orthogonal rotation matrices in the initialization step (Sec. \ref{sec:init_Ym}), and \textit{(iii)} the fully-proposed algorithm. Firstly, the graph shows that the Algorithm \ref{algo:cross_modal_bin} takes a few iterations to converge. Secondly, as we note that lower objective values indicate the fact that the algorithm can potentially achieve better binary cross-modality representations, i.e., well represented for all modalities. The fact is also confirmed by the average \textit{mAP}s of Img $\to$ Txt and Txt $\to$ Img retrieval tasks. These empirical results emphasize the importance of jointly learning the anchors (Sec. \ref{sec:anchors}) and the necessity of initializing $\{\bY^m\}_{m=1}^M$ such that they are highly pairwise-correlated.

\subsubsection{Benefits of anchor-to-anchor mapping} 
\label{sec:ablation_anchor2anchor}
We additionally investigate the effect of including the anchor-to-anchor mapping when building the anchor graph. We first conduct the experiments on MIR-Flickr25k, IAPR-TC12, and NUS-WIDE datasets with $L\eq 32$ and fixed $k_a\eq 2$. Fig. \ref{fig:anchor_to_anchor} shows the average of \textit{mAP} for Img $\to$ Txt and Txt $\to$ Img retrieval tasks when changing $k$. We can observe that the anchor-to-anchor mapping generally helps to achieve considerable improvements on the retrieval performance for all datasets. 
Note that the improvements come at a very small computational cost since $P\ll N$. 
In addition, at larger $k$ (e.g., $k\ge 4$ for MIR-Flickr25k and IAPR-TC12 and $k\ge 6$ for NUS-WIDE), the performance drops as the anchor graphs contain more spurious connections \cite{AnchorGraph}. 

Additionally, we also investigate the performance when $k_a$ varies. Fig. \ref{fig:ka} shows \textit{mAP} on MIR-Flickr25k, IAPR-TC12, and NUS-WIDE datasets for various $k_a$ values. $k_a=0$ means no anchor-to-anchor mapping is applied. The figure shows that generally the anchor-to-anchor mapping with small $k_a$ (e.g., $k_a\in\{1,2,3\}$), can help to achieve clear improvement gains. With larger $k_a$ (e.g., $k_a\ge 4$), the anchor-to-anchor mapping contains more spurious connections making the mapping less reliable.


\subsubsection{Deep and Deeper models}
To evaluate the effectiveness and necessary of the deep architecture DCSH, we compare DCSH with its two variants, which use shallower networks (in term of the number of trainable layers). Specifically, we use the widely used linear model, denoted as \textbf{DCSH-L}, and a simple 3 fully-connected layer DNN $[D^m\hspace{-0.05em}\to\hspace{-0.05em}1024\hspace{-0.05em}\to\hspace{-0.05em}512\to\hspace{-0.05em}L]$, \textbf{DCSH-F}, for $\func^m$.  We train the shallow models from extracted features, e.g., VGG FC7 for the image models and BOW for the text models, instead of images and document embeddings as DCSH. 

\begin{figure}[t]
\centering
\includegraphics[width=0.23\textwidth]{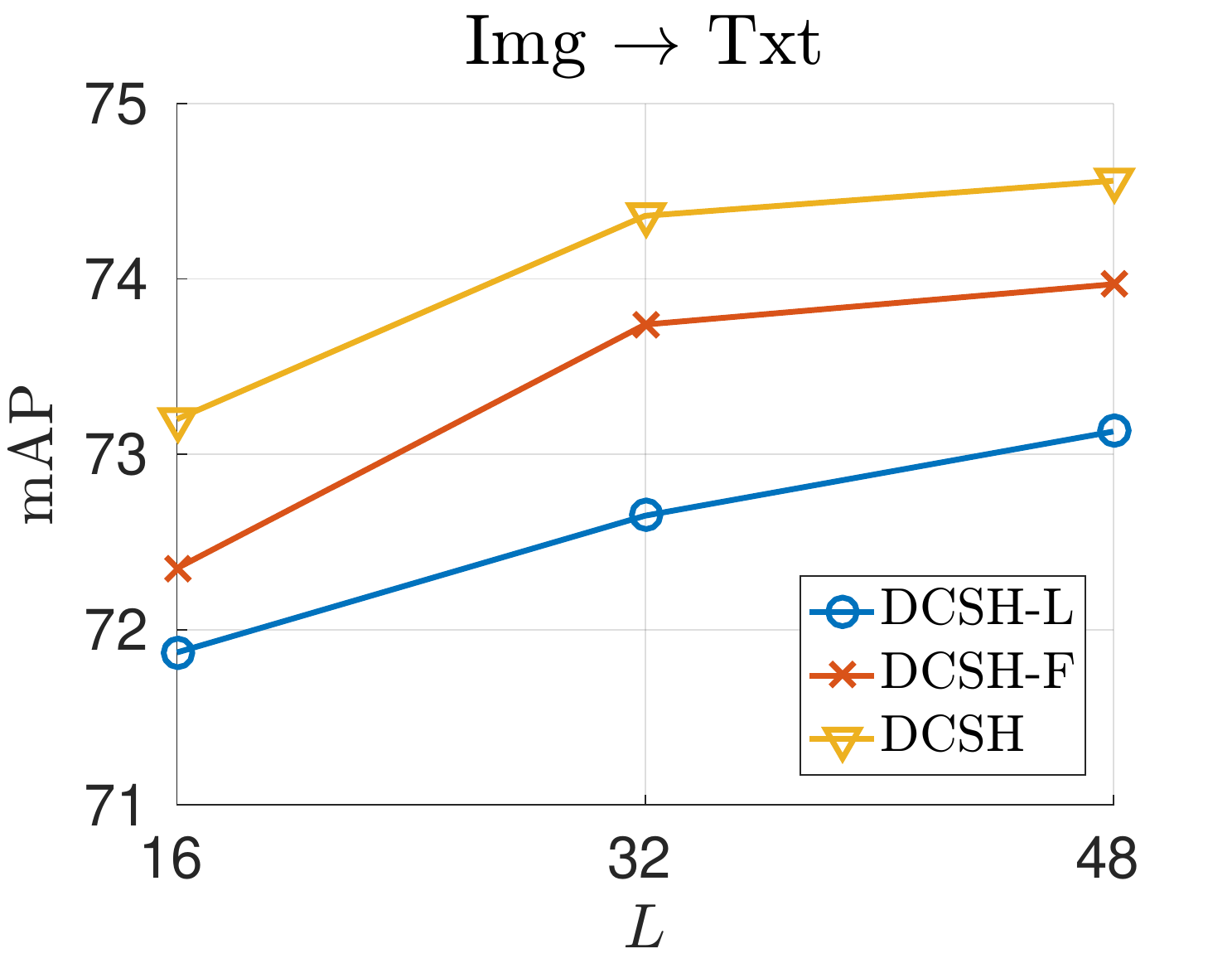}
~
\includegraphics[width=0.23\textwidth]{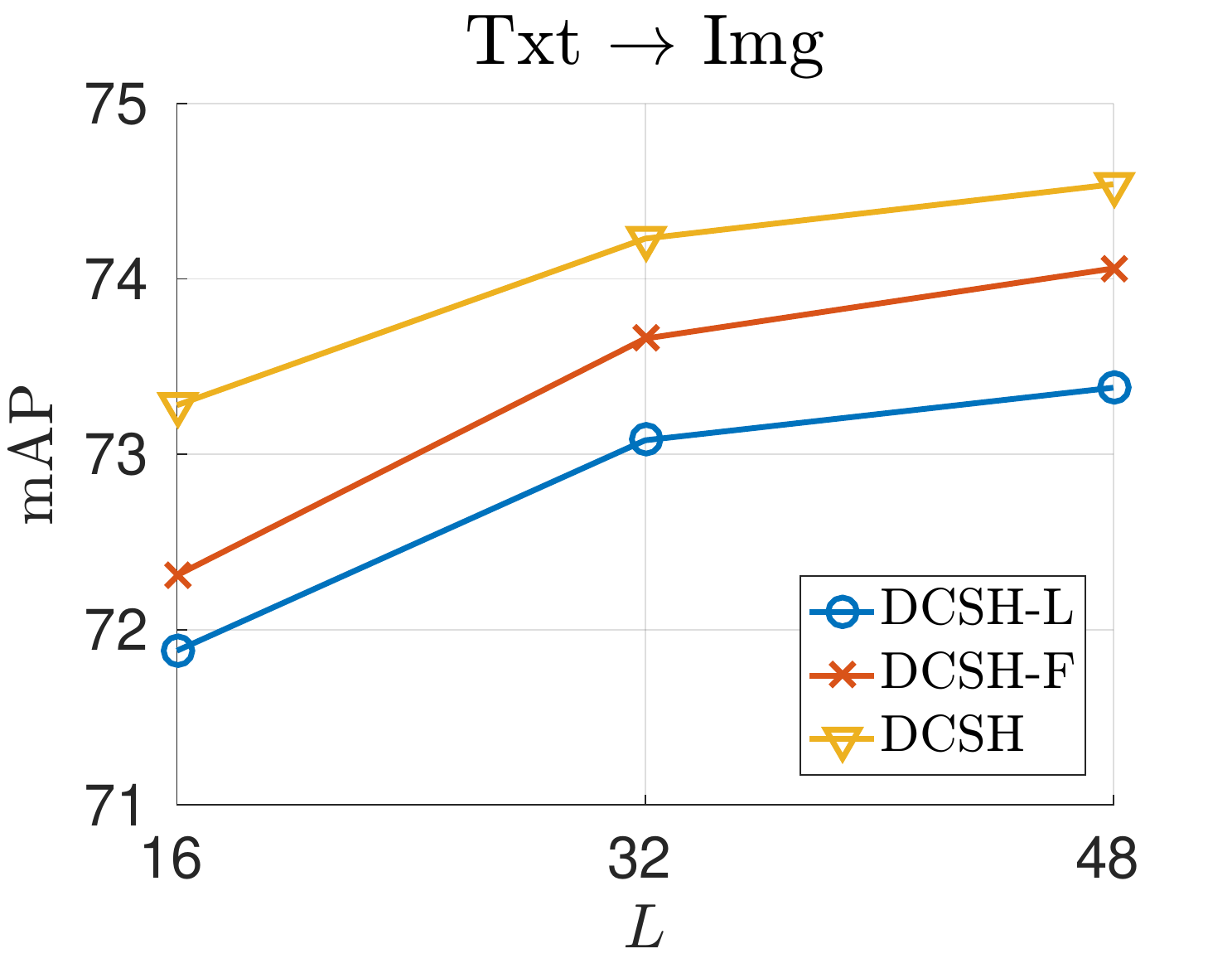}
\caption{The benefits of using deeper models and more informative data. The experiment is conducted on MIR-Flickr25K.
}
\label{fig:deep_deeper}
\end{figure}

Fig. \ref{fig:deep_deeper} shows the retrieval performance on MIR-Flickr25K dataset for various hash code lengths. We can observe clearly that by using deeper models: DCSH-L $\to$ DCSH-F $\to$ DCSH, we can achieve better performance. We note that using deeper DNN models, i.e., adding more layers, with extracted features is not helpful as the deeper DNN model is prone to overfitting. The experimental results show the necessary of having both deeper models and more informative inputs.
More specifically, using deeper models and more informative inputs can ensure that modality-specific hashing functions well capture the information contained in the cross-modality binary representation $\bB$, which significantly affects the retrieval performance. 

\subsection{Comparison with the states of the art}
\label{ssec:compare_state_of_art}

\begin{figure}[t]
\centering
\begin{subfigure}[b]{0.48\textwidth}
\includegraphics[width=0.49\textwidth]{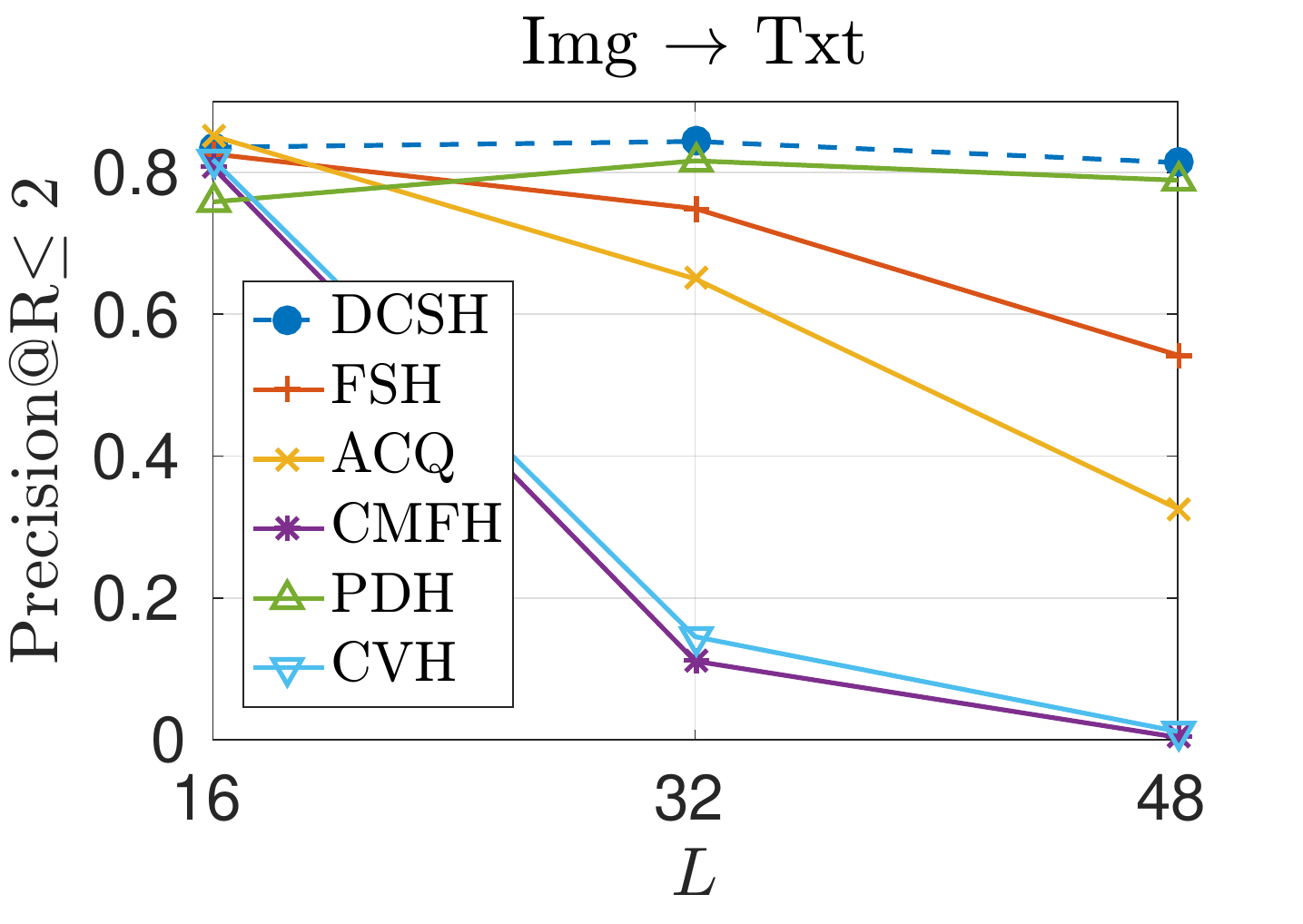}
\includegraphics[width=0.49\textwidth]{fig/mirflickr25k_preR2_img2txt.pdf}
\caption{MIR-Flickr25k}
\end{subfigure}
\begin{subfigure}[b]{0.48\textwidth}
\includegraphics[width=0.49\textwidth]{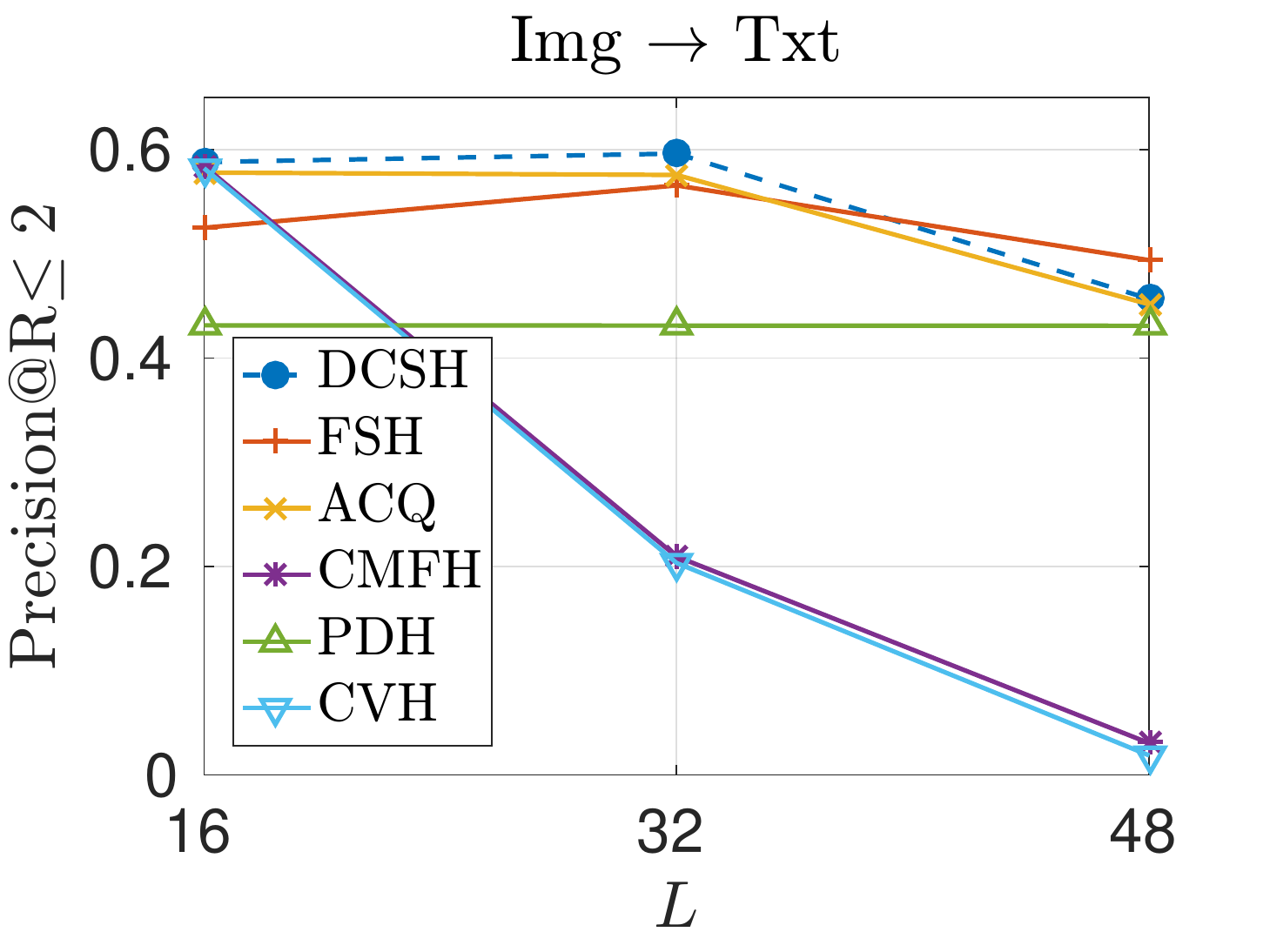}
\includegraphics[width=0.49\textwidth]{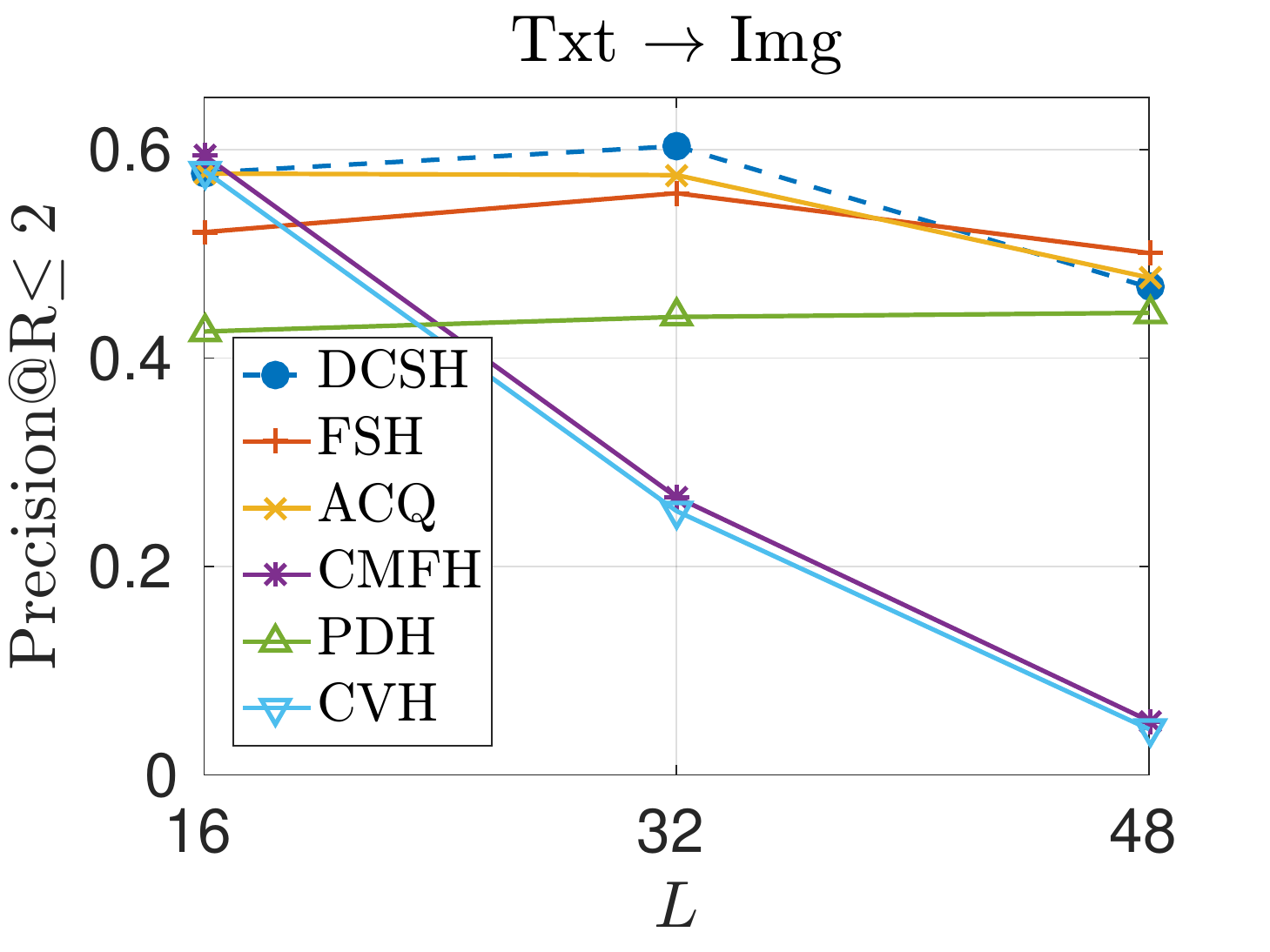}
\caption{IAPR-TC12}
\end{subfigure}
\begin{subfigure}[b]{0.48\textwidth}
\includegraphics[width=0.49\textwidth]{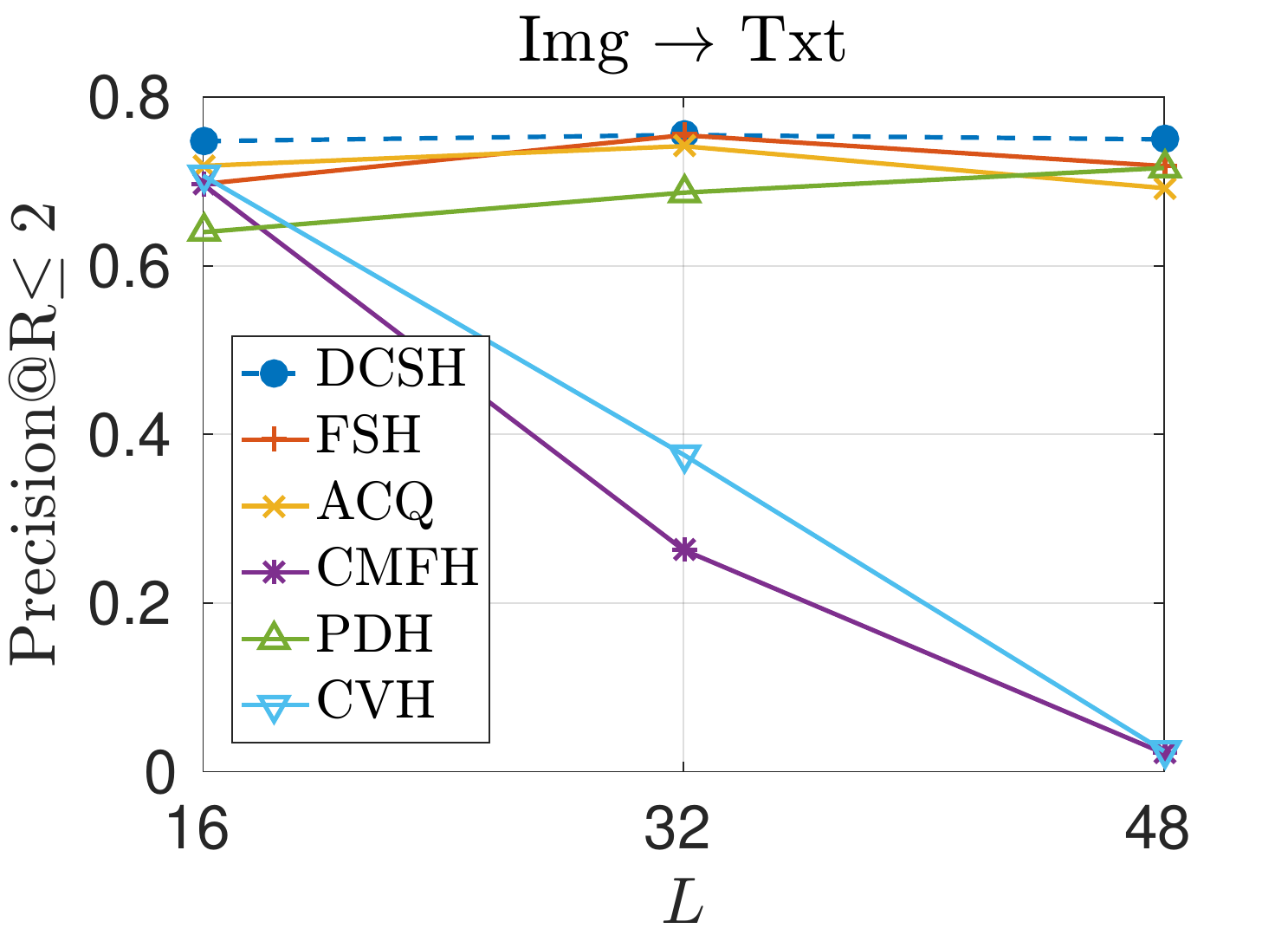}
\includegraphics[width=0.49\textwidth]{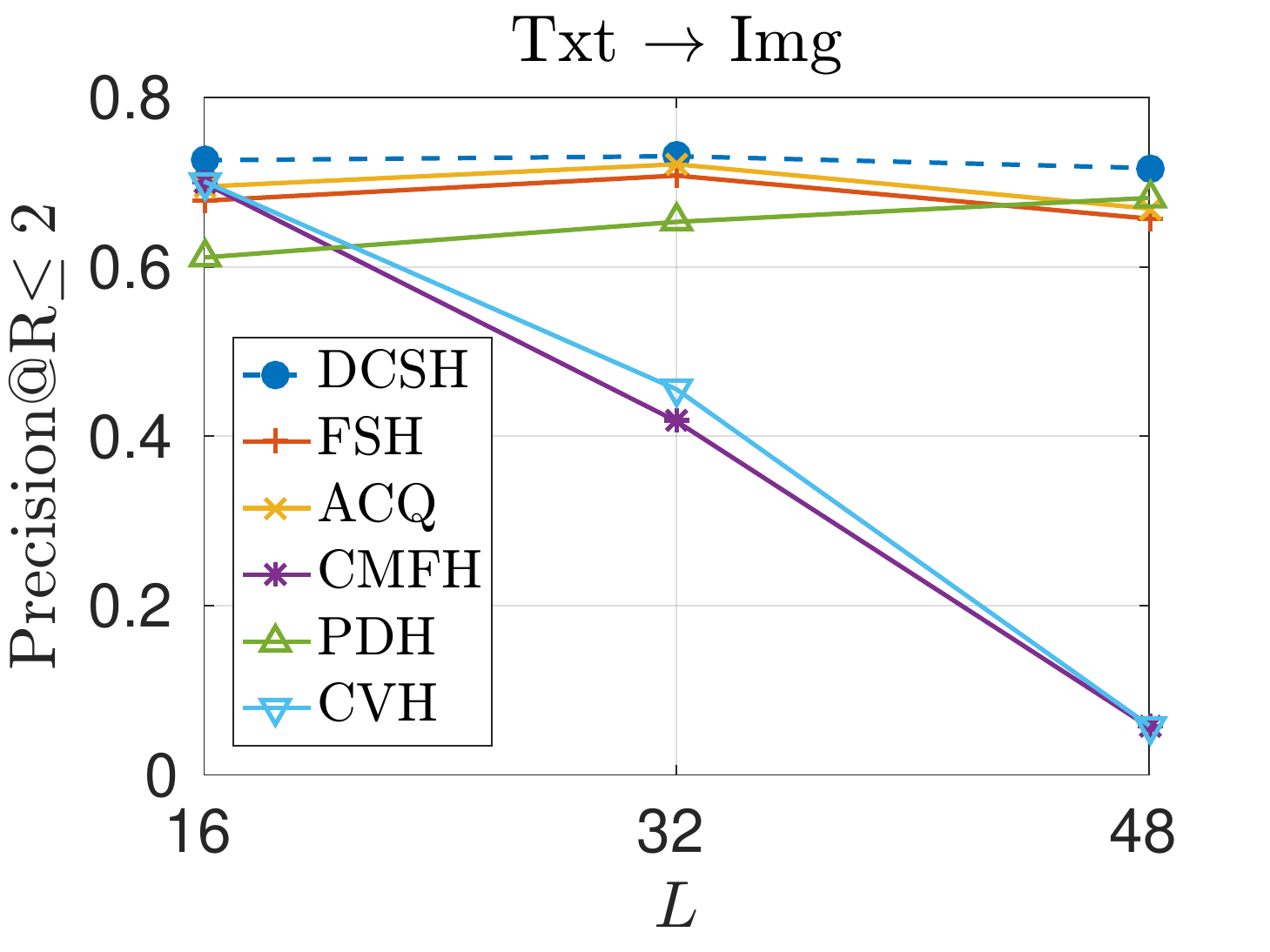}
\caption{NUS-WIDE}
\label{fig:preck}
\end{subfigure}

\caption{The \textit{Prec@R$\le$2} on three benchmark datasets at different code lengths, i.e., $L=16,32,48$.}
\label{fig:precR2}
\end{figure}

\begin{table*}[t]
\small
\centering
\def\arraystretch{1.1}
\setlength{\tabcolsep}{9pt}
\caption{Comparison results using \textit{mAP} on three benchmark datasets. 
}
\label{tb:img_txt}
\begin{tabular}{|c|l|c|c|c|c|c|c|c|c|c|}
\hline
\multirow{2}{*}{Task} & \multirow{2}{*}{Method} & \multicolumn{3}{c|}{MIR-Flickr25k} & \multicolumn{3}{c|}{IAPR-TC12} & \multicolumn{3}{c|}{NUS-WIDE} \\ \cline{3-11}
&  & 16 & 32 & 48 & 16 & 32 & 48 & 16 & 32 & 48  \\
\hline 
\multirow{7}{*}{\rotatebox{90}{Img$\to$Txt}} 
& CVH \cite{CVH} & 65.41 & 63.19 & 62.06 & 41.98 & 40.56 & 39.79 & 51.38 & 49.33 & 47.74 \\ \cline{2-11}
& PDH \cite{PDH} & 71.07 & 72.26 & 72.42 & 43.62 & 45.16 & 44.88 & 55.01 & 56.90 & 59.39 \\ \cline{2-11}
& CMFH \cite{CMFH} & 65.87 & 65.90 & 65.81 & 43.15 & 44.02 & 44.46 & 54.58 & 56.24 & 57.85\\ \cline{2-11}
& ACQ \cite{ACQ} & 71.42 & 71.57 & 71.76 & 45.54 & 46.21 & 46.85 & 57.89 & 59.03 & 58.75 \\ \cline{2-11}
& FSH \cite{FSH} & 70.07 & 72.35 & 72.25 & 44.71 & 45.78 & 46.88 & 57.56 & 60.37 & 59.45 \\ \cline{2-11}
& \textbf{DCSH-L} & 71.87 & 72.65 & 73.13 & 45.59 & 46.67 & 47.21 & 59.10 & 60.35 & 60.54 \\ \cline{2-11}
& \textbf{DCSH} & {\color{black}\textbf{73.20}} & {\color{black}\textbf{74.26}} &  {\color{black}\textbf{74.46}} & {\color{black}\textbf{46.35}} & {\color{black}\textbf{47.16}} & {\color{black}\textbf{47.79}} & {\color{black}\textbf{60.41}} & {\color{black}\textbf{61.92}} & {\color{black}\textbf{62.76}} \\ \hline
\multirow{7}{*}{\rotatebox{90}{Txt$\to$Img}} 
& CVH & 65.75 & 63.40 & 62.21 & 42.23 & 40.86 & 40.11 & 51.29 & 49.28 & 47.76 \\ \cline{2-11}
& PDH & 71.92 & 72.73 & 73.06 & 43.74 & 45.35 & 45.09 & 55.61 & 57.25 & 59.29 \\ \cline{2-11}
& CMFH & 66.06 & 66.12 & 65.94 & 43.85 & 44.65 & 44.97 & 54.78 & 56.81 & 58.15 \\ \cline{2-11}
& ACQ & 71.59 & 71.74 & 71.93 & 45.67 & 46.32 & 47.13 & 58.09 & 58.83 & 58.49 \\ \cline{2-11}
& FSH & 70.06 & 72.46 & 72.36 & 44.55 & 45.89 & 46.98 & 57.15 & 60.11 & 59.56 \\ \cline{2-11}
\cline{2-11}
& \textbf{DCSH-L} & 71.88 & 73.08 & 73.38 & 45.86 & 46.67 & 47.35 & 59.11 & 60.47 & 60.69 \\ \cline{2-11}
& \textbf{DCSH} & {\color{black}\textbf{73.58}} & {\color{black}\textbf{74.13}} &  {\color{black}\textbf{74.44}} & {\color{black}\textbf{46.46}} & {\color{black}\textbf{47.23}} & {\color{black}\textbf{47.93}} & {\color{black}\textbf{60.67}} & {\color{black}\textbf{62.12}} & {\color{black}\textbf{63.40}} \\ \hline
\end{tabular}
\end{table*}

\begin{table*}[t]
\small
\centering
\def\arraystretch{1.1}
\setlength{\tabcolsep}{9pt}
\caption{Comparison results using \textit{mAP} on three benchmark datasets.}
\label{tb:img_img_txt_txt}
\begin{tabular}{|c|l|c|c|c|c|c|c|c|c|c|}
\hline
\multirow{2}{*}{Task} & \multirow{2}{*}{Method} & \multicolumn{3}{c|}{MIR-Flickr25k} & \multicolumn{3}{c|}{IAPR-TC12} & \multicolumn{3}{c|}{NUS-WIDE} \\ \cline{3-11}
&  & 16 & 32 & 48 & 16 & 32 & 48 & 16 & 32 & 48  \\
\hline 
\multirow{7}{*}{\rotatebox{90}{Img$\to$Img}} 
& CVH & 66.92 & 64.28 & 62.95 & 42.14 & 40.66 & 39.84 & 53.24 & 51.06 & 49.11 \\ \cline{2-11}
& PDH & 74.81 & 76.01 & 76.12 & 43.96 & 45.60 & 45.86 & 58.64 & 60.29 & 62.95 \\ \cline{2-11}
& CMFH & 66.59 & 65.78 & 65.33 & 42.82 & 43.19 & 43.40 & 57.12 & 58.98 & 61.05 \\ \cline{2-11}
& ACQ & 74.63 & 74.68 & 74.86 & 45.99 & 46.67 & 47.22 & 62.24 & 63.21 & 64.13 \\ \cline{2-11}
& FSH & 71.81 & 76.06 & 75.91 & 45.16 & 46.17 & 47.34 & 62.96 & 63.49 & 64.06 \\ \cline{2-11}
& \textbf{DCSH-L} & 74.79 & 76.38 & 76.71 & 46.12 & 46.97 & 47.49 & 63.61 & 64.12 & 64.52 \\ \cline{2-11}
& \textbf{DCSH} & {\color{black}\textbf{75.97}} & {\color{black}\textbf{77.23}}  & {\color{black}\textbf{77.55}} & {\color{black}\textbf{46.85}} & {\color{black}\textbf{47.52}} & {\color{black}\textbf{48.19}} & {\color{black}\textbf{64.56}} & {\color{black}\textbf{65.31}} & {\color{black}\textbf{66.73}} \\ \hline
\multirow{7}{*}{\rotatebox{90}{Txt$\to$Txt}} 
& CVH & 64.39 & 62.41 & 61.34 & 42.24 & 40.86 & 40.12 & 49.74 & 47.91 & 46.71 \\ \cline{2-11}
& PDH & 68.68 & 69.75 & 69.86 & 43.42 & 44.96 & 44.63 & 52.56 & 54.28 & 56.20 \\ \cline{2-11}
& CMFH & 65.48 & 65.59 & 65.99 & 44.25 & 45.48 & 46.01 & 51.87 & 53.46 & 55.57 \\ \cline{2-11}
& ACQ & 68.84 & 69.04 & 69.26 & 45.24 & 45.91 & 46.72 & 54.71 & 55.01 & 55.51 \\ \cline{2-11}
& FSH & 67.38 & 69.61 & 69.30 & 44.18 & 45.54 & 46.55 & 54.58 & 54.74 & 55.88 \\ \cline{2-11}
& \textbf{DCSH-L} & 69.52 & 70.54 & 70.96 & 45.42 & 46.18 & 46.90 & 55.83 & 56.89 & 57.41 \\ \cline{2-11}
& \textbf{DCSH} & {\color{black}\textbf{71.54}} & {\color{black}\textbf{71.54}} & {\color{black}\textbf{71.87}} & {\color{black}\textbf{46.09}} & {\color{black}\textbf{46.56}} & {\color{black}\textbf{47.27}} & {\color{black}\textbf{57.09}} & {\color{black}\textbf{59.24}} & {\color{black}\textbf{59.80}} \\ \hline
\end{tabular}
\end{table*}

The quantitative results to compare our proposed method with other state-of-the-art unsupervised cross-modal hashing methods are shown in Tables \ref{tb:img_txt} and \ref{tb:img_img_txt_txt} (\textit{mAP}) and in Fig. \ref{fig:preck} (\textit{Prec@R$\le$2}).
%
Our proposed method DCSH consistently and significantly outperforms all compared methods in terms of \textit{mAP} across four retrieval tasks at all code lengths, and on three standard benchmark datasets, including the large scale NUS-WIDE dataset.

Additionally, as shown in Fig. \ref{fig:precR2}, DCSH also achieves more favorable performances in terms of \textit{Prec@R$\le$2}. Specifically, for short code lengths, e.g., $L=16$, DCSH achieves comparable performance with compared methods on MIR-Flickr25k and IAPR-TC12. While at larger code lengths, e.g., $L=32,48$, DCSH generally outperforms most of the compared methods, except being slight lower than FSH and comparable to ACQ at $L=48$ on IAPR-TC12 dataset.
Note that many previous hashing methods achieve worse retrieval performance with longer code lengths. This undesirable effect arises since the Hamming space will become increasingly sparse with longer code lengths and fewer data points will fall in the Hamming ball of radius 2 \cite{S3PLH}. It is worth noting that DCSH achieves a relatively mild decrease in accuracy using longer code lengths, validating that DCSH can concentrate hash codes of similar points together to be within Hamming radius 2, which is beneficial to Hamming space retrieval.

Furthermore, we want to emphasize that even with using linear models for modality-specific hashing models, i.e., DCSH-L, our proposed framework still consistently outperforms compared methods. Especially, the improvement margins are clearer, $\ge 0.3\%$ \textit{mAP}, at higher code lengths, e.g., $L=32$ and $48$.
This demonstrates the effectiveness of our proposed algorithm \ref{algo:cross_modal_bin} to learn the cross-modality binary representations.

\begin{table}[t]
\centering
\def\arraystretch{1.1}
\caption{Comparison with UGACH \cite{UGACH} using \textit{mAP} on MIR-Flickr25k and NUS-WIDE datasets. The results of UGACH are cited from \cite{UGACH}.}
\label{tb:compare_ugach}
\resizebox{1.0\columnwidth}{!}{
\begin{tabular}{|c|l|c|c|c|c|c|c|}
\hline
\multirow{2}{*}{Task} & \multirow{2}{*}{Method} & \multicolumn{3}{c|}{MIR-Flickr25k} & \multicolumn{3}{c|}{NUS-WIDE} \\ \cline{3-8}
&  & 16 & 32 & 64 & 16 & 32 & 64 \\ \hline
Img$\to$ & UGACH  & 68.5 & 69.3 & 70.4 & \textbf{61.3} & 62.3 & 62.8 \\ \cline{2-8}
Txt & \textbf{DCSH} & \textbf{72.48} & \textbf{74.62} & \textbf{75.29} & 60.81 & \textbf{62.58} & \textbf{63.25} \\ \hline
Txt$\to$ & UGACH  & 67.3 & 67.6 & 68.6 & 60.3 & 61.4 & \textbf{64.0} \\ \cline{2-8}
Img & \textbf{DCSH} & \textbf{71.93} & \textbf{73.61} & \textbf{74.23} & \textbf{60.92} & \textbf{62.76} & 63.94\\ \hline
\end{tabular}
}
\end{table}

\textbf{Comparison with Unsupervised Generative Adversarial Cross-Modal Hashing (UGACH) \cite{UGACH}:}
For a fair comparison, we conduct additional experiments on MIR-FLickr25k and NUS-WIDE datasets following the experiment settings from \cite{UGACH}. More specifically, the FC7 features of the pretrained 19-layer VGGNet are used for images, instead of the 16-layer VGGNet. 1,000-dimension BOW features are used for texts in both datasets. We take 1\% samples of the NUS-WIDE dataset and 5\% samples of the MIR-FLickr25k dataset as the query
sets, and the rest as the retrieval database. 
{For the very large dataset NUS-WIDE, we randomly select 20,000 pairs from the database to form the training set for our method.
The comparison results in term of \textit{mAP} are shown in Table \ref{tb:compare_ugach}. We can observe that, for MIR-Flickr25k dataset, our proposed DCSH consistently outperforms UGACH by a clear margin ($\ge 3.5\%$). For NUS-WIDE dataset, even though using less training samples, our proposed DCSH generally outperforms UGACH for the majority of cases.
}

{
\textbf{Comparison with Collective Reconstructive Embedding (CRE) \cite{CRE} and Fusion Similarity Hashing (FSH) \cite{FSH}:} Following the experiment setting of CRE and FSH, we conduct experiments with hand-crafted features on MIR-Flickr25k and NUS-WIDE datasets.
For MIR-Flickr25k dataset, images are represented with 100-dimensional BoW SIFT features and the texts are expressed as 500-dimensional tagging vectors. After removing the instances without any label or textual tag, 
we have 16,738 instances remained. 5\% of instances (i.e., 836) are sampled as the query set and the remaining are used as the database and training set. 
For NUS-WIDE dataset, each image is represented by 500-dimensional BoW SIFT features and each text is represented by a 1,000-dimension preprocessed BOW feature. We randomly select 2,000 pairs as the query set; the remaining are used as the database. We also sample 20,000 pairs from the database as the training set. For a fair comparison, we use the provided hand-crafted features as the input to learn the 3-layer DNN modal-specific hashing functions. 
We present the experiment results in Table \ref{tb:compare_new_methods}. The results in Table \ref{tb:compare_new_methods}, \ref{tb:img_txt}, and \ref{tb:img_img_txt_txt} also show that our method consistently outperforms FSH for both hand-crafted and CNN-based features. Besdies, our method outperforms CRE for both Img$\to$Txt and Txt$\to$Img tasks on NUS-WIDE dataset. However, on MIR-Flickr25k dataset, our method slightly underperforms CRE on Img$\to$Txt task, while clearly outperform CRE on Txt$\to$Img task.
}
\begin{table}[t]
\small
\centering
\def\arraystretch{1.1}
\caption{{Comparison with FSH \cite{FSH} and CRE \cite{CRE}  using \textit{mAP} on NUS-WIDE dataset. The results of FSH and CRE are cited from \cite{FSH} and \cite{CRE} respectively.}}
\label{tb:compare_new_methods}
\resizebox{1.0\columnwidth}{!}
{
\begin{tabular}{|c|l|c|c|c|c|c|c|}
\hline
\multirow{2}{*}{Task} & \multirow{2}{*}{Method} & \multicolumn{3}{c|}{MIR-Flickr25k} & \multicolumn{3}{c|}{NUS-WIDE} \\ \cline{3-8}
&  & 16 & 32 & 64 & 16 & 32 & 64 \\ \hline
& FSH \cite{FSH} & 59.68 & 61.89 & 61.95 & 50.59 & 50.63 & 51.71 \\ \cline{2-8}
Img$\to$ & CRE \cite{CRE} & \textbf{62.11} & \textbf{62.51} & \textbf{62.90} & 51.31 & 52.99 & 53.32 \\ \cline{2-8}
Txt & \textbf{DCSH-DNN}     & {62.08} & {62.34} & {62.57} & \textbf{53.05} & \textbf{53.71} & \textbf{54.21} \\ \hline
& FSH           & 59.24 & 61.28 & 60.91 & 47.90 & 48.10 & 49.65 \\ \cline{2-8}
Txt$\to$ & CRE & 61.49 & 61.82 & 62.17 & 49.27 & 50.86 & 51.49 \\ \cline{2-8}
Img & \textbf{DCSH-DNN} & \textbf{62.57} & \textbf{62.71} & \textbf{63.13} & \textbf{52.15} & \textbf{52.42} & \textbf{52.89} \\ \hline
\end{tabular}
}
\end{table}

{
\textbf{Comparison with Unsupervised Deep Cross Modal Hashing (UDCMH) \cite{UDCMH}:}
Similar to UDCMH, for images, 
we use the PyTorch pretrained AlexNet to extract the FC7 features to construct the similarity graph for the image modality.  
In addition, following UDCMH, we report the mAP of top-50 retrieved results (\textit{mAP@50}). The experiment results are shown in Table \ref{tb:compare_UDCMH}. 
We observe that our proposed method can outperform UDCMH by large margins for both MIR-Flickr25k and NUS-WIDE datasets. 
Additionally, the retrieval performance of UDCMH for the two tasks Img$\to$Txt and Txt$\to$Img for NUS-WIDE dataset is very different. This potentially indicates a misalignment between the hash code spaces of two modalities. While for our method, the retrieval performance of UDCMH for the two tasks Img$\to$Txt and Txt$\to$Img is comparable. This demonstrates the effectiveness of our contributions, including the building similarity anchor graph for multi-modal context and the optimization procedure. 
}
\begin{table}[t]
\small
\centering
\def\arraystretch{1.1}
\caption{{Comparison with UDCMH \cite{UDCMH} using \textit{mAP@50} on MIR-FLickr25k and NUS-WIDE datasets. The results of UDCMH are cited from \cite{UDCMH}.}}
\label{tb:compare_UDCMH}
\resizebox{1.0\columnwidth}{!}
{
\begin{tabular}{|c|l|c|c|c|c|c|c|}
\hline
\multirow{2}{*}{Task} & \multirow{2}{*}{Method} & \multicolumn{3}{c|}{MIR-Flickr25k} & \multicolumn{3}{c|}{NUS-WIDE} \\ \cline{3-8}
&  & 16 & 32 & 64 & 16 & 32 & 64 \\ \hline
Img$\to$ & UDCMH \cite{UDCMH} & 68.9 & 69.8 & 71.4 &  51.1 &  51.9 &  52.4  \\ \cline{2-8}
Txt & \textbf{DCSH-DNN} & \textbf{82.9} & \textbf{85.9} & \textbf{87.4} & \textbf{66.12} & \textbf{68.23} & \textbf{71.25} \\ \hline
Txt$\to$ & UDCMH    & 69.2 & 70.4 & 71.8 &  63.7 &  65.3 &  69.5 \\ \cline{2-8}
Img & \textbf{DCSH-DNN} & \textbf{81.7} & \textbf{83.1} & \textbf{85.9} & \textbf{66.41} & \textbf{68.99} & \textbf{72.54} \\ \hline
\end{tabular}
}
\end{table}

\red{
\textbf{Comparison between handcrafted features and deep-based features"}
In Table \ref{tb:handcrafted_deep}, we present the retrieval results of our proposed method when using handcrafted features  and deep-based features. Specifically, for the handcrafted features, we use BoW of SIFT for images and BoW of keywords for text. For deep-based features, we use FC7 features of VGG-16 network for images and Globe embeddings for text with Eq. \eqref{eq:document_embedding}. Unsurprisingly, the deep-based features can help to achieve significantly higher performance in compared with the handcrafted features \cite{8709820,8296975}. 
}
\begin{table}[h]
\small
\centering
\def\arraystretch{1.1}
\caption{\red{Comparison between the hand-crafted features and CNN features using our proposed method.}}
\label{tb:handcrafted_deep}
\resizebox{1.0\columnwidth}{!}
{
\red{
\begin{tabular}{|c|l|c|c|c|c|c|c|}
\hline
\multirow{2}{*}{Task} & \multirow{2}{*}{Features} & \multicolumn{3}{c|}{MIR-Flickr25k} & \multicolumn{3}{c|}{NUS-WIDE} \\ \cline{3-8}
&  & 16 & 32 & 48 & 16 & 32 & 48 \\ \hline
Img$\to$ & Hand-crafted & 61.58 & 62.08 & 62.35 & 53.05 & 53.71 & 54.02  \\ \cline{2-8}
Txt & Deep-based & 73.20 & 74.26 & 74.46 & 60.41 & 61.92 & 62.76 \\ \hline
Txt$\to$ & Hand-crafted & 61.67 & 62.24 & 62.45 & 52.15 & 52.42 & 52.73 \\ \cline{2-8}
Img & Deep-based & 73.58 & 74.13 & 74.44 & 60.67 & 62.12 & 63.40 \\ \hline
\end{tabular}
}
}
\end{table}
\begin{figure}[t]
\centering

\begin{subfigure}[b]{0.235\textwidth}
\includegraphics[width=\textwidth]{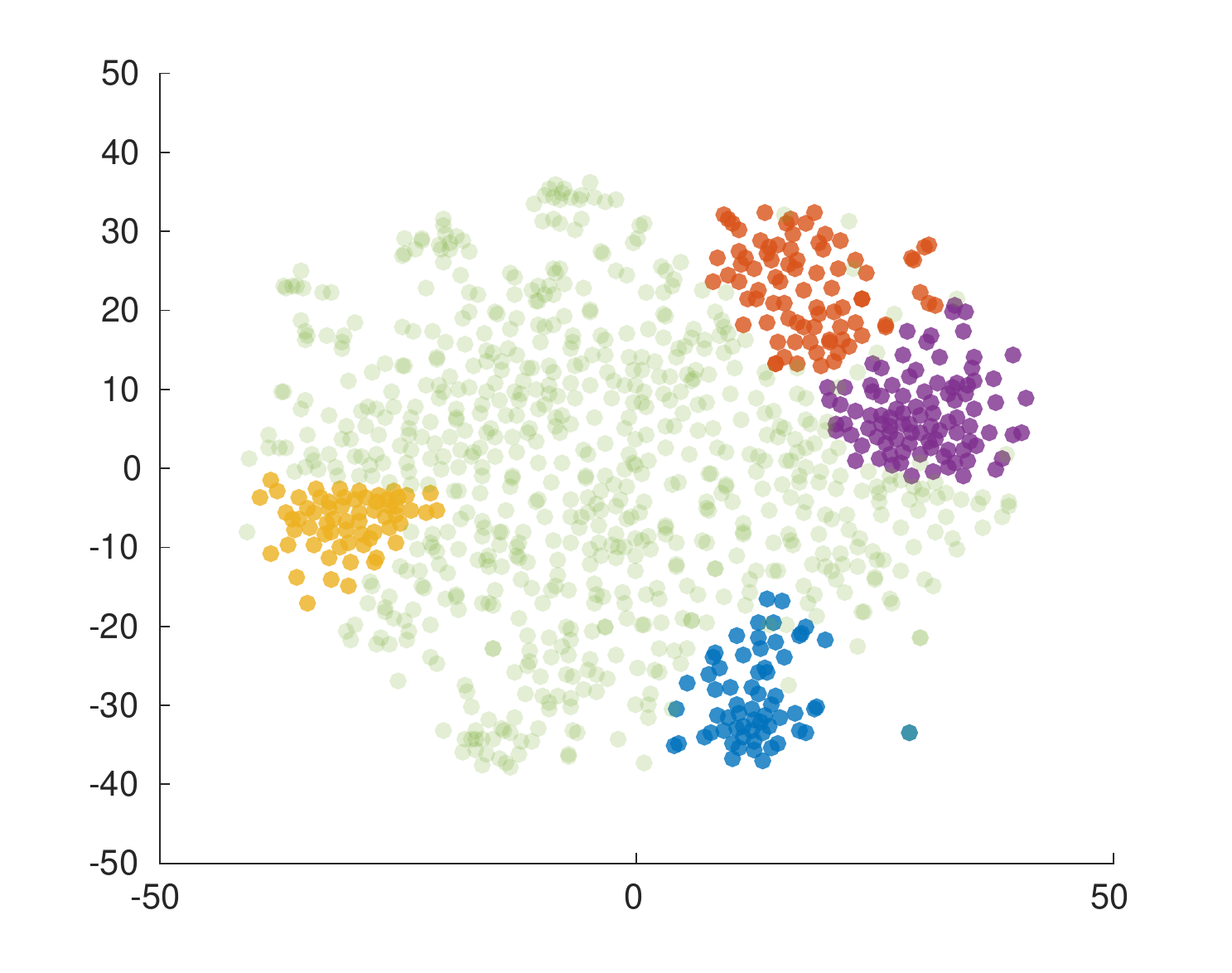}
\caption{Image modality data $\bX^1$}
\label{fig:mirflickr25k_tsne_img}
\end{subfigure}
~
\begin{subfigure}[b]{0.235\textwidth}
\includegraphics[width=\textwidth]{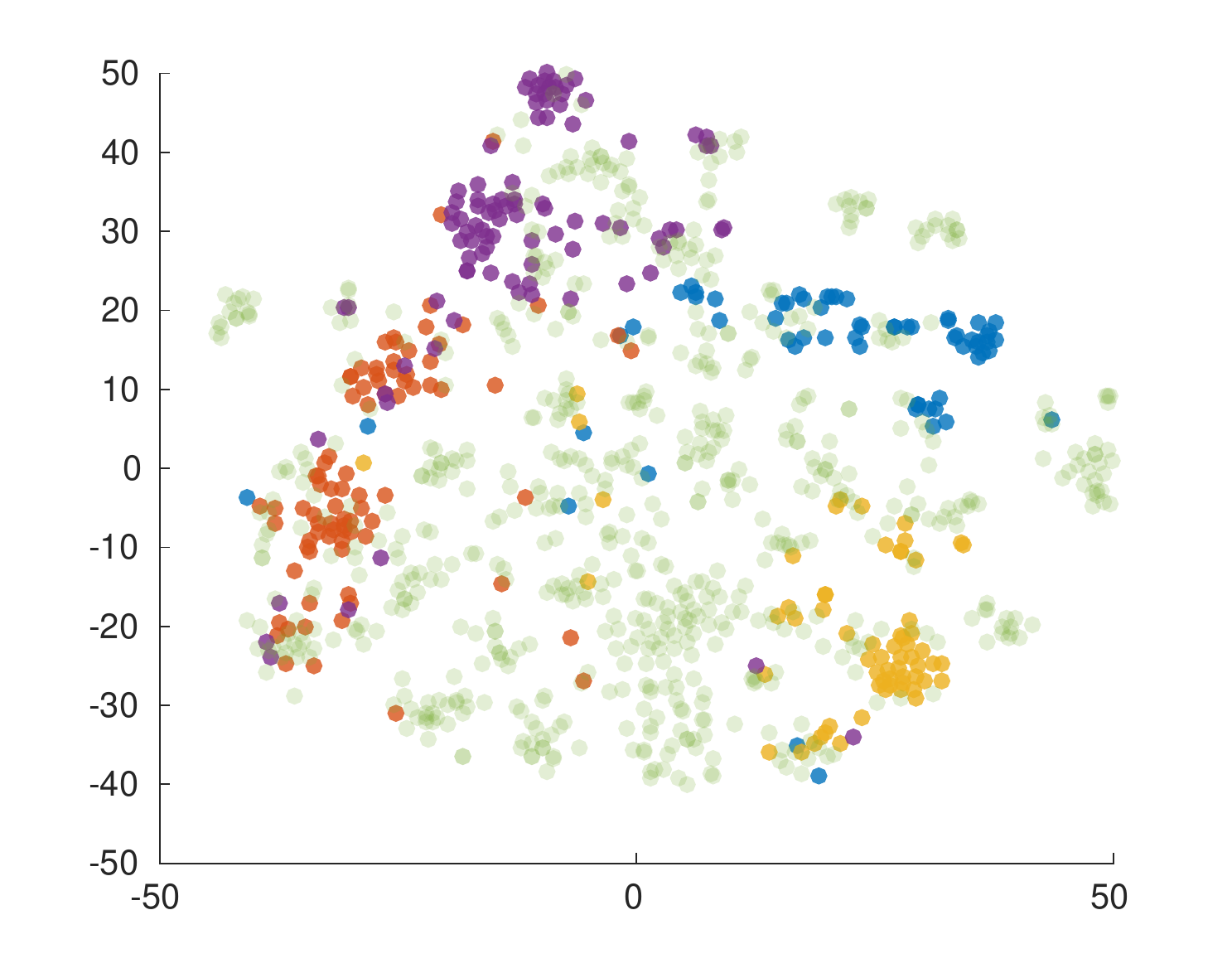}
\caption{Cross-modal binary code $\bB$}
\label{fig:mirflickr25k_tsne_img_code}
\end{subfigure}
\caption{The t-SNE visualization of MIR-Flickr25k input data points of image modality $\bX^1$ and their corresponding cross-modality 32-bit binary representation $\bB$. 
}
\label{fig:mirflickr25k_visualize_img}
\end{figure}

\begin{figure}[t]
\centering
\begin{subfigure}[b]{0.235\textwidth}
\includegraphics[width=\textwidth]{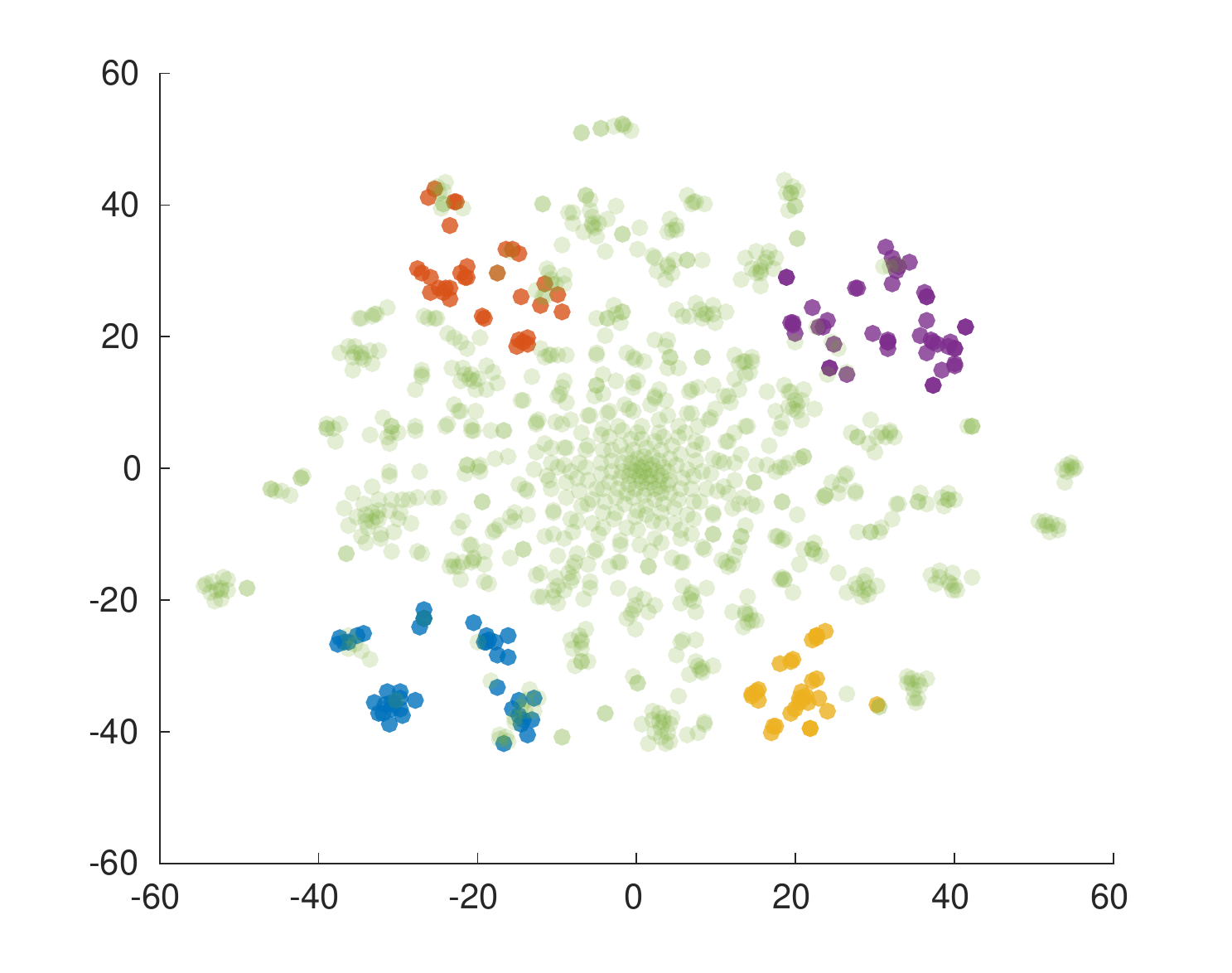}
\caption{Text modality data $\bX^2$}
\label{fig:mirflickr25k_tsne_txt}
\end{subfigure}
~
\begin{subfigure}[b]{0.235\textwidth}
\includegraphics[width=\textwidth]{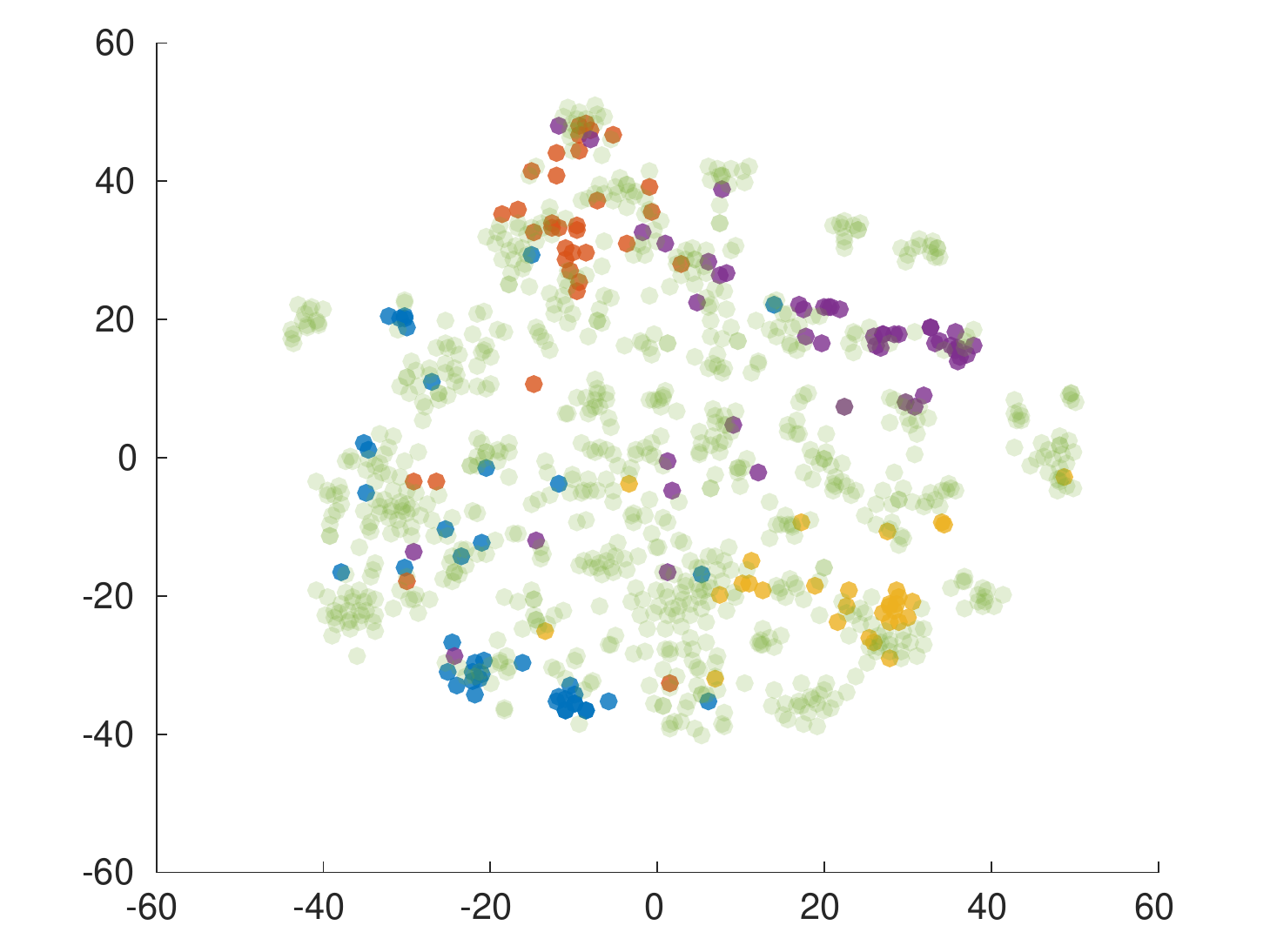}
\caption{Cross-modal binary code $\bB$}
\label{fig:mirflickr25k_tsne_txt_code}
\end{subfigure}

\caption{The t-SNE visualization of MIR-Flickr25k input data points of text  modality $\bX^2$ and their corresponding cross-modality 32-bit binary representation $\bB$. 
}
\label{fig:mirflickr25k_visualize_txt}
\end{figure}

\begin{figure}[h]
\centering

\begin{subfigure}[b]{0.235\textwidth}
\includegraphics[width=\textwidth]{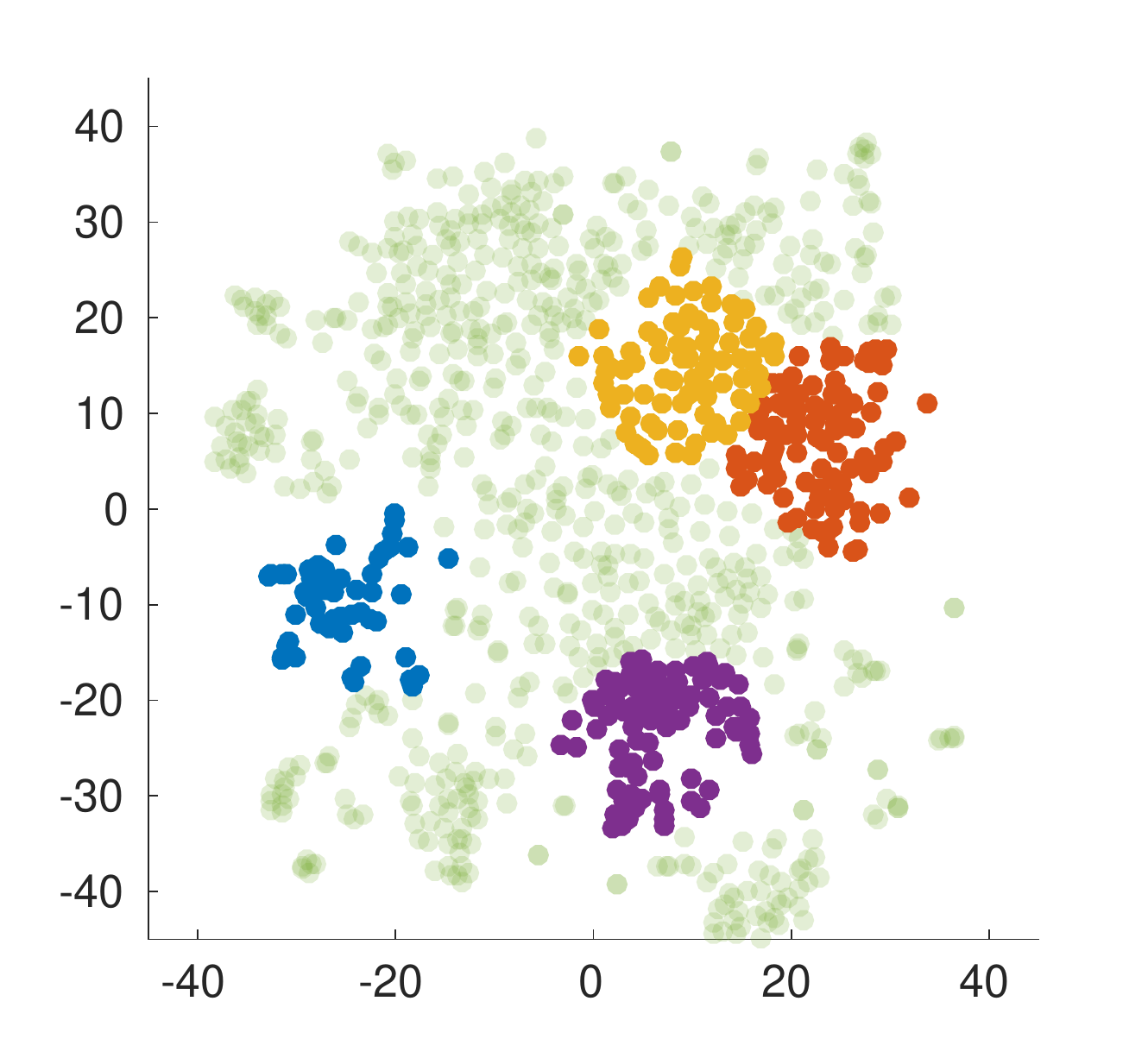}
\caption{Image modality data $\bX^1$}
\label{fig:nuswide_tsne_img}
\end{subfigure}
~
\begin{subfigure}[b]{0.235\textwidth}
\includegraphics[width=\textwidth]{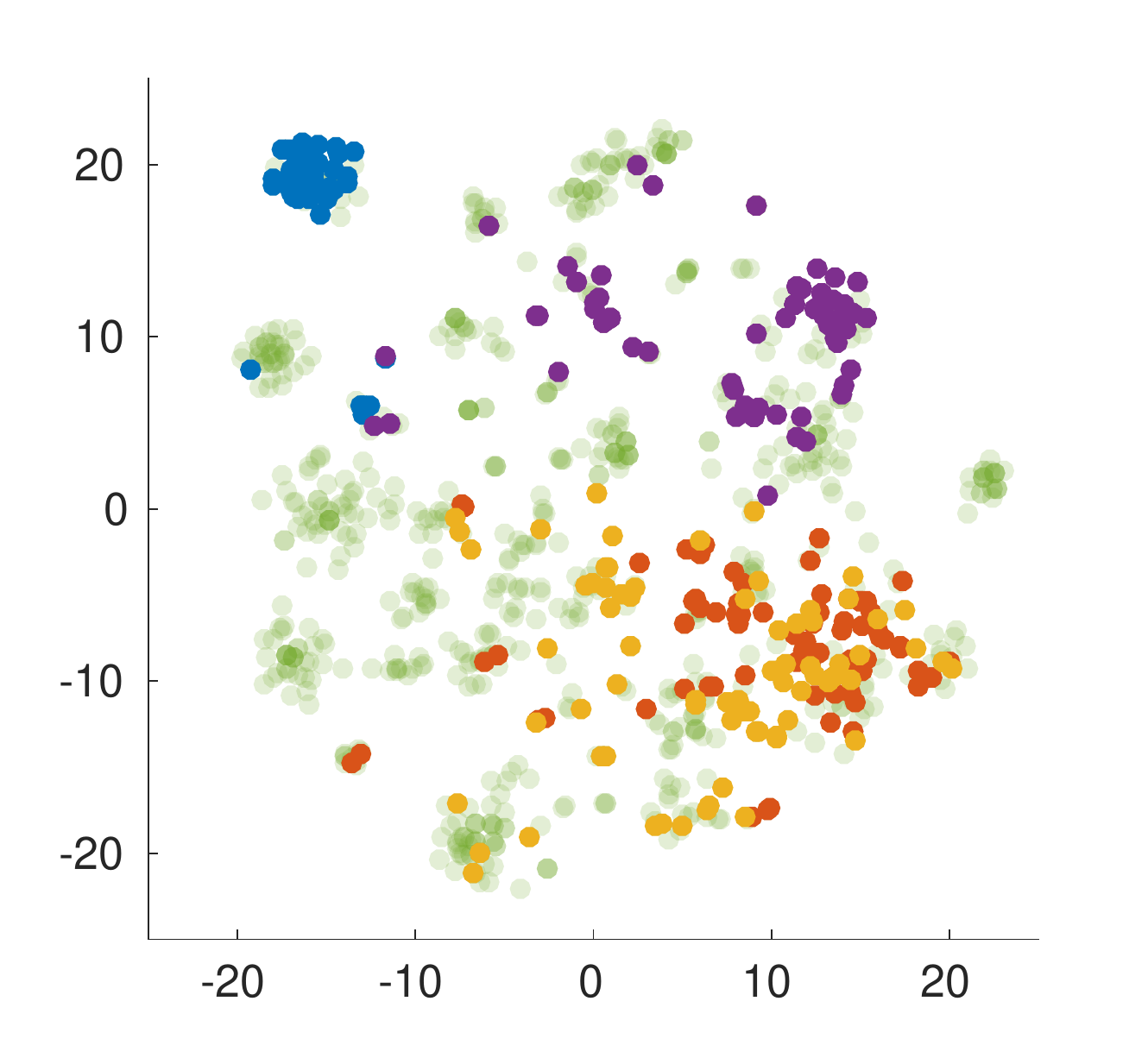}
\caption{Cross-modal binary code $\bB$}
\label{fig:nuswide_tsne_img_code}
\end{subfigure}
\caption{\red{The t-SNE visualization of NUS-WIDE input data points of image modality $\bX^1$ and their corresponding cross-modality 32-bit binary representation $\bB$.}
}
\label{fig:nuswide_visualize_img}
\end{figure}

\begin{figure}[h]
\centering

\begin{subfigure}[b]{0.235\textwidth}
\includegraphics[width=\textwidth]{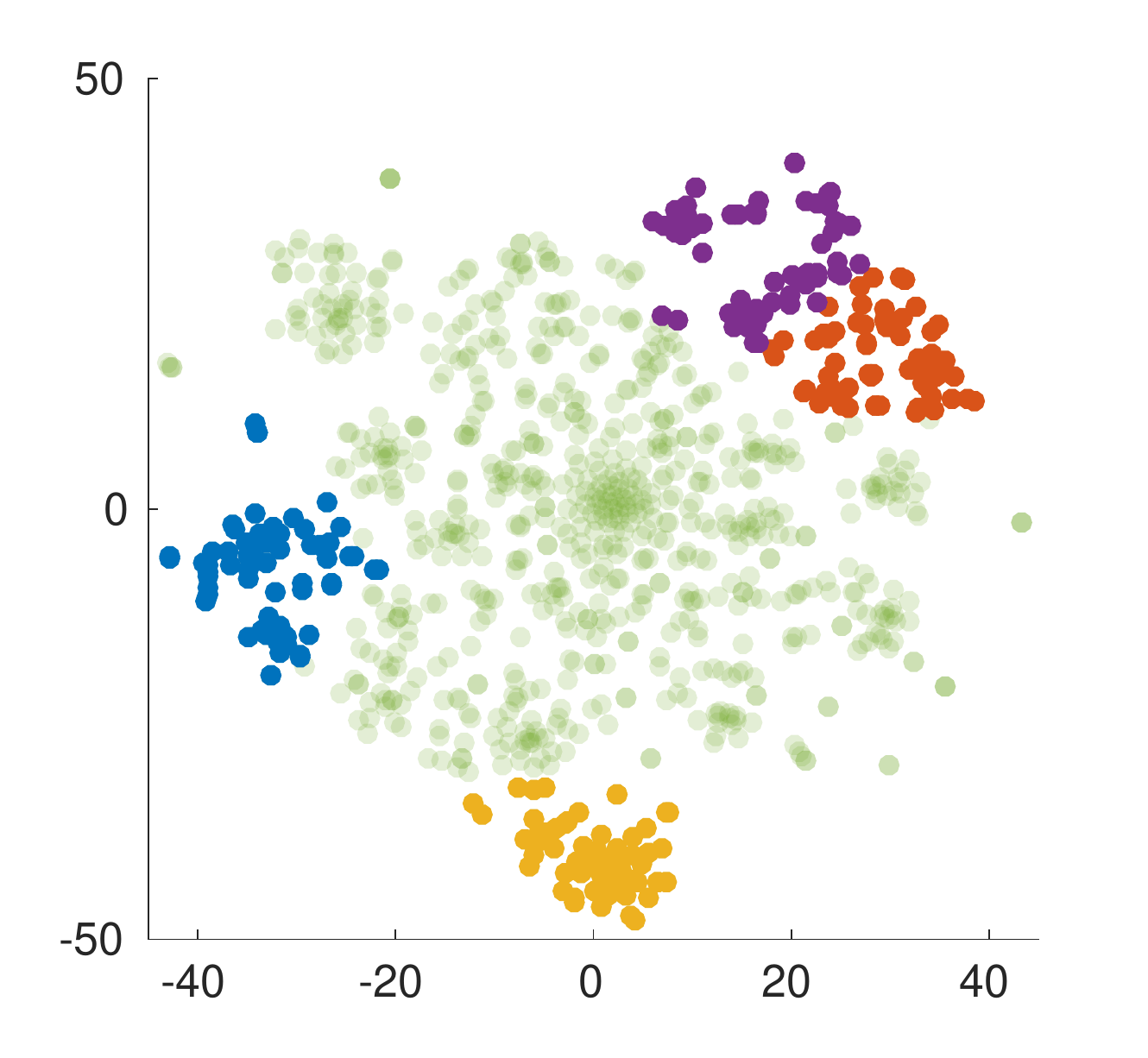}
\caption{Image modality data $\bX^1$}
\label{fig:nuswide_tsne_img}
\end{subfigure}
~
\begin{subfigure}[b]{0.235\textwidth}
\includegraphics[width=\textwidth]{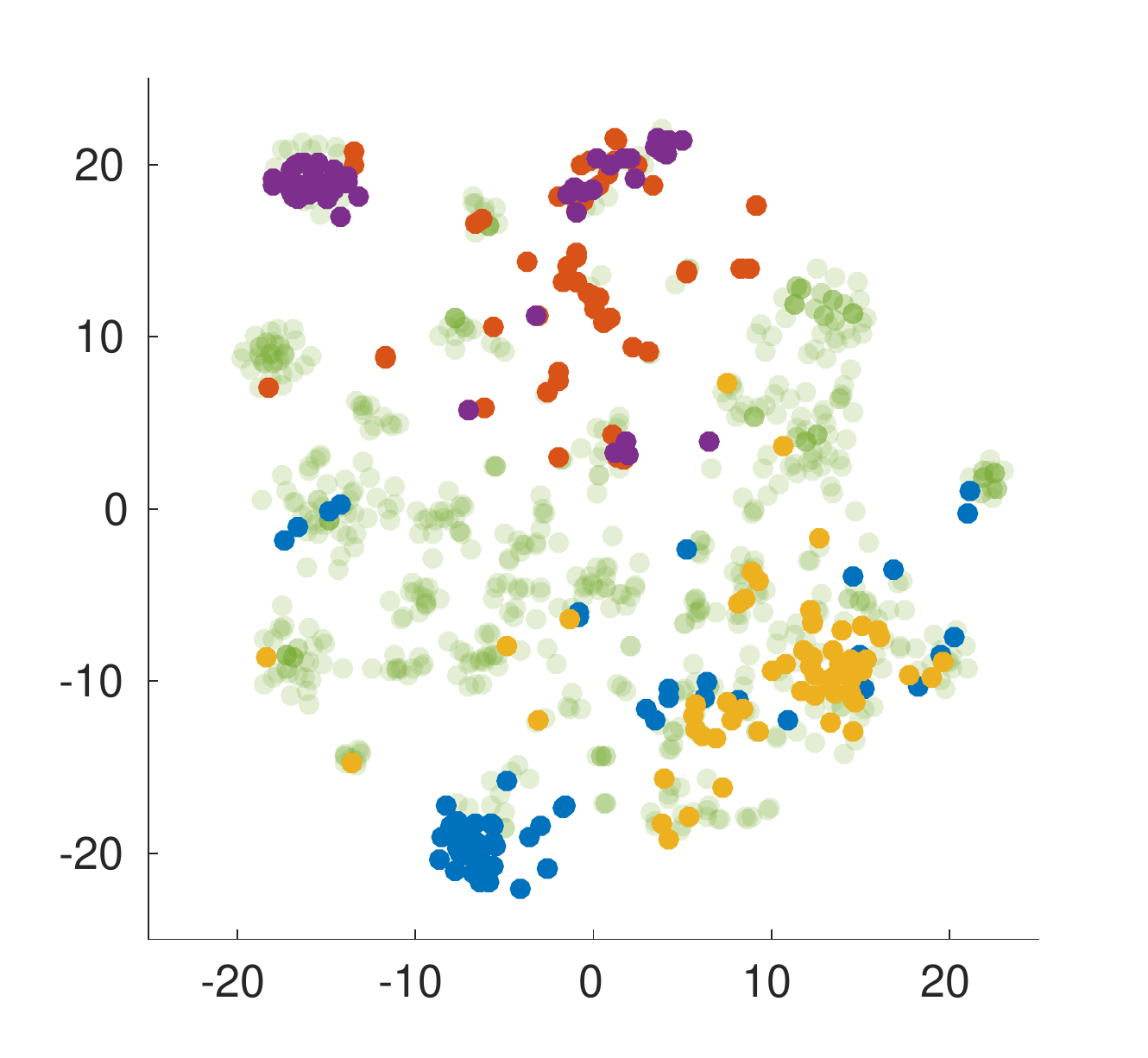}
\caption{Cross-modal binary code $\bB$}
\label{fig:nuswide_tsne_img_code}
\end{subfigure}
\caption{\red{The t-SNE visualization of NUS-WIDE input data points of image modality $\bX^1$ and their corresponding cross-modality 32-bit binary representation $\bB$.}
}
\label{fig:nuswide_visualize_txt}
\end{figure}

\subsection{Visualization}

\red{
Fig. \ref{fig:mirflickr25k_visualize_img} and \ref{fig:nuswide_visualize_img} show the t-SNE visualizations \cite{tsne} of the input data of image modality $\bX^1$ and their corresponding cross-modality binary representation $\bB$ for MIR-Flickr25k and NUS-WIDE datasets respectively. Similarly, Fig. \ref{fig:mirflickr25k_visualize_txt} and \ref{fig:nuswide_visualize_txt} show the t-SNE visualizations of the input data of text modality $\bX^2$ and their corresponding cross-modality binary representation $\bB$ for MIR-Flickr25k and NUS-WIDE datasets respectively. 
For clarity, only 1,000 data-pairs randomly selected from the training set are displayed.
Additionally, in order to clearly visualize the local structures of datasets which is necessary to show the effectiveness of our method, we apply $k$-means on 1,000 selected data-pairs to learn 15 clusters.
The data points, which belong to a same cluster and {share at least one common class label}, are displayed in a same color

Firstly, as seen in Fig. \ref{fig:mirflickr25k_visualize_img} and \ref{fig:mirflickr25k_visualize_txt}, our proposed method can learn cross-modality binary codes that well preserve the local structures of datasets. Specifically, data points, which belong to a same cluster in text or image modality (Fig. \ref{fig:mirflickr25k_tsne_img} and \ref{fig:mirflickr25k_tsne_txt}), are mapped to very similar binary codes (Fig. \ref{fig:mirflickr25k_tsne_img_code} and \ref{fig:mirflickr25k_tsne_txt_code}). Secondly, we also observe that if data points in the input space are far from each other, e.g., yellow, red, and blue data points in image and text modality spaces, their corresponding cross-modality binary codes are also far from other binary codes. Generally, from Fig. \ref{fig:nuswide_visualize_img} and \ref{fig:nuswide_visualize_txt}, we also have the similar observation for NUS-WIDE dataset.
In summary, the visualization clearly shows the effectiveness of our proposed DCSH method in fusing heterogeneous modalities to produce discriminative binary codes.
}

\subsection{Parameter Analysis}
\begin{figure}[t]
\centering
\includegraphics[width=0.25\textwidth]{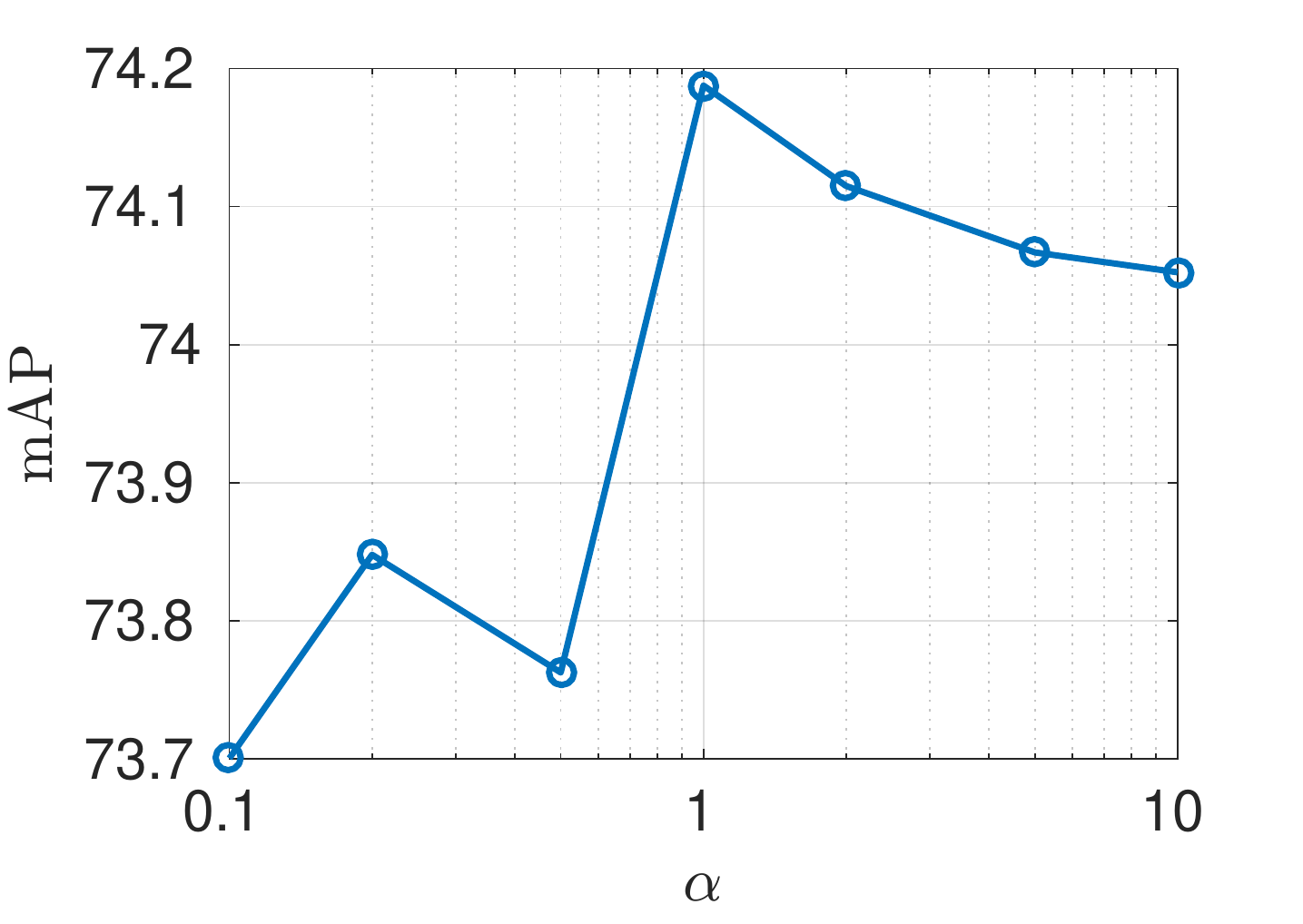}
\caption{$\alpha$ analysis on MIR-Flickr25k and $L=32$.}
\label{fig:alpha_analysis}
\end{figure}

In this section, we firstly analyze the most important parameter of the proposed method,
i.e., $\alpha$, which controls the contributions of spectral clustering loss and the maximizing correlation loss, i.e., intra and inter-modality similarity.
We conduct the analysis on MIR-Flickr25k dataset and $L\eq 32$; and then plot the \textit{mAP} curve as $\alpha$ changes on Fig. \ref{fig:alpha_analysis}.
From Fig. \ref{fig:alpha_analysis}, we can observe that at small $\alpha$, e.g., $\alpha\hspace{-0.25em}\le\hspace{-0.25em} 0.5$, the performances are slightly unstable as the optimization focuses more on learning the single-modality representations. At larger $\alpha$, e.g., $\alpha\hspace{-0.25em}\ge\hspace{-0.25em} 1$, we achieve much higher performances. Nevertheless, the performances slightly drop as $\alpha$ gets larger as the optimization pays too much attention on the maximizing correlation part, which results in lower-quality single-modality representations. Finally, we note that for $\alpha \in[0.1, 10]$, our proposed method still outperforms other compared methods.

\begin{figure}[t]
\centering

\begin{subfigure}[b]{0.48\textwidth}
\includegraphics[width=0.48\textwidth]{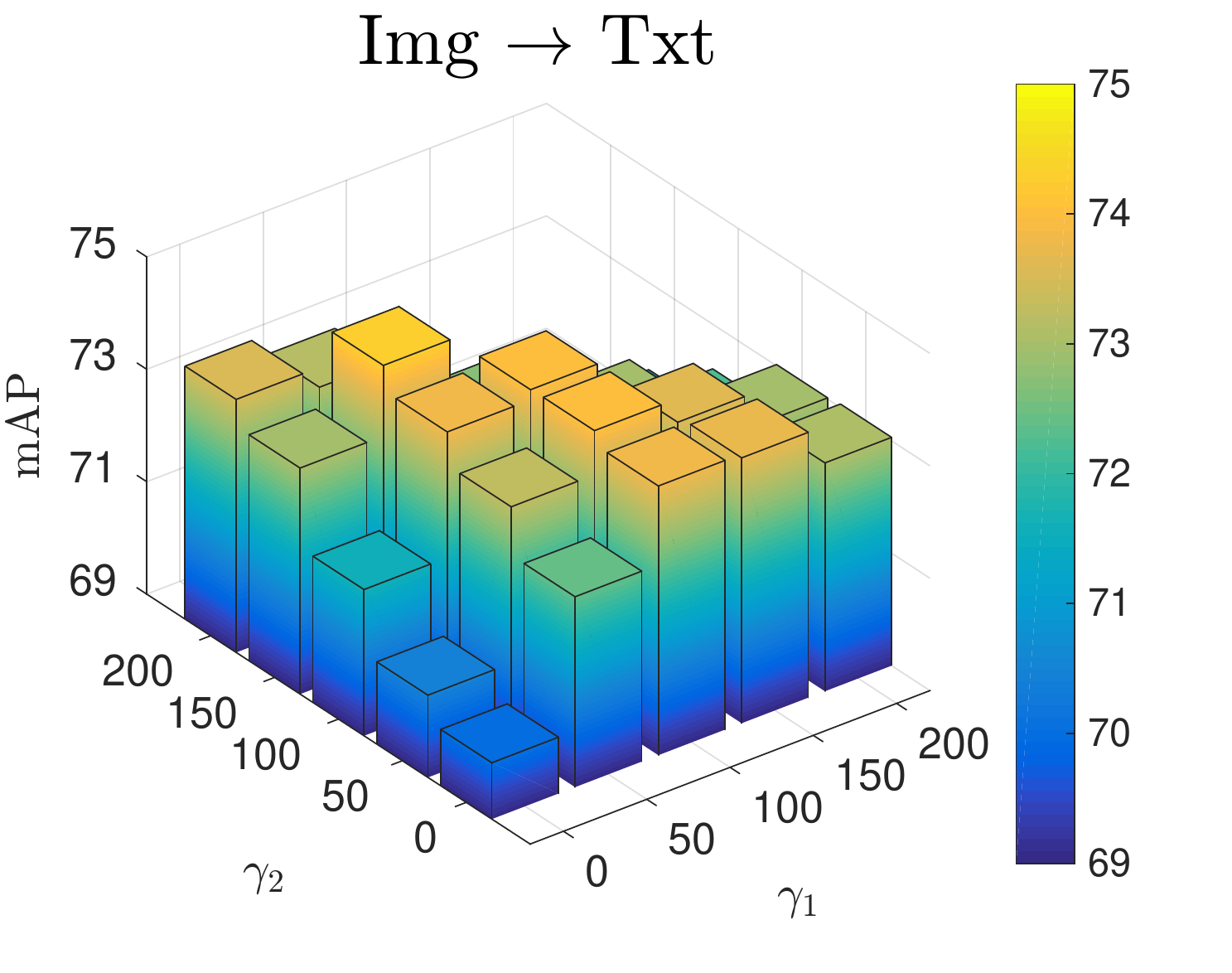}
\includegraphics[width=0.48\textwidth]{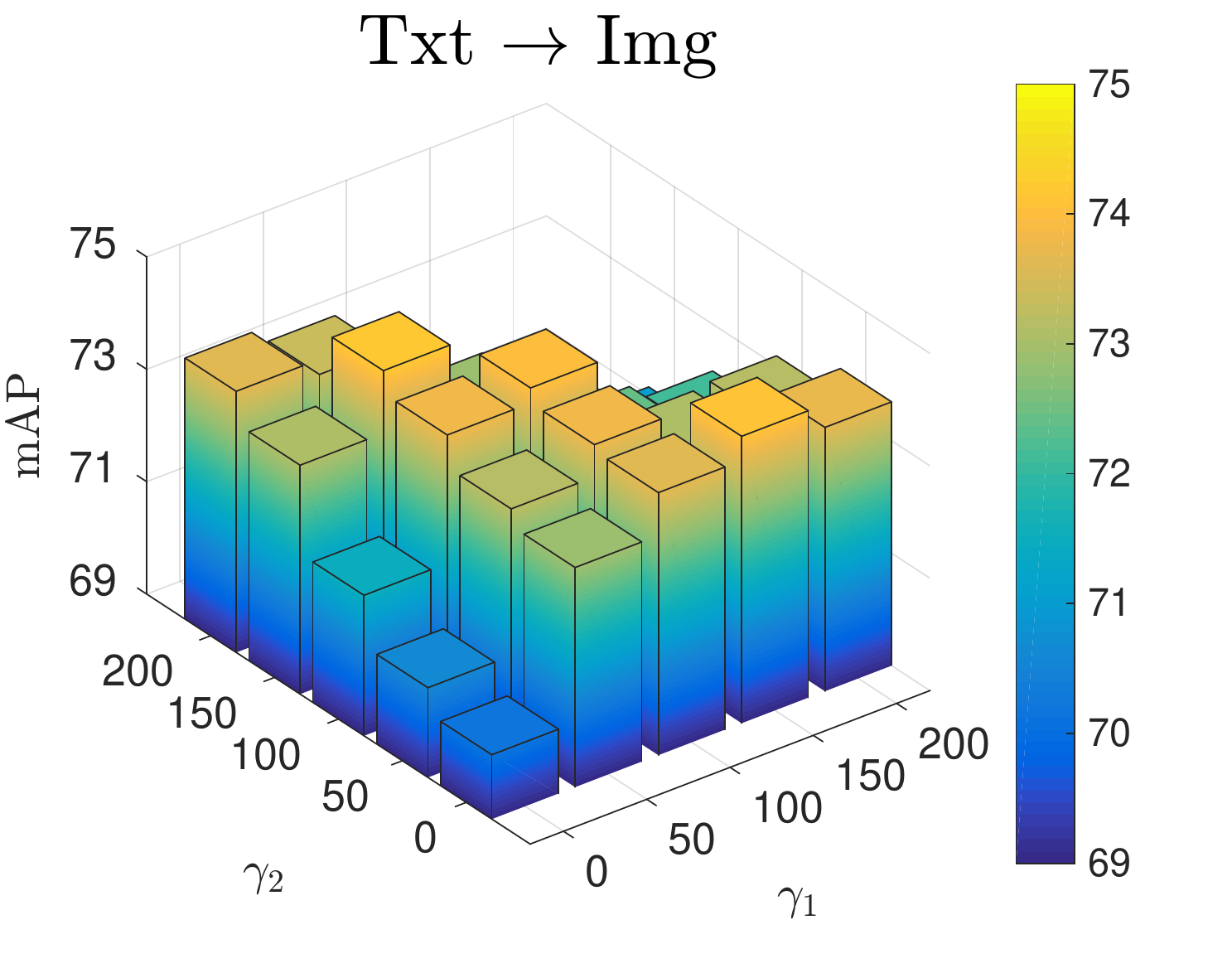}
\end{subfigure}

\caption{Cross-modal retrieval performance of DCSH with
different values of parameters $\gamma_1$ and $\gamma_2$ on MIR-Flickr25k dataset at $L=32$.}
\label{fig:gamma}
\end{figure}

We additionally conduct empirical analyses on the hyper-parameters of learning the modality-specific hashing models, i.e., $\{\gamma_1, \gamma_2\}$.
The experimental results are shown in Fig. \ref{fig:gamma}.
The figure shows that the proposed method DCSH generally achieves the best performance when both $\gamma_1$ and $\gamma_2$ are within the range of $[50, 100]$. The method has lower performance when $\{\gamma_1,\gamma_2\}$ are too small.
This fact emphasizes the importance of \textit{(i)} the \textit{independent} and \textit{(ii)} the \textit{high-correlation} properties of the hashing function outputs.
The performance is also low when $\{\gamma_1,\gamma_2\}$ are too large.
This effect is also understandable as the loss function pays too attention on learning the independent and high-correlation properties. 
While it loses the focus on minimizing the discrepancy between the outputs and the learned cross-modality binary representation $\bB$, which captures the data structures of  modalities.





\section{Conclusion}
\label{sec:concl}
In this paper, we proposed a novel unsupervised cross-modality hashing framework, named {\textit{Deep Cross-modality Spectral Hashing (DCSH)}}. The framework is based on the general two-step hashing method which helps to simplify the binary optimization and hashing function learning.
Specifically, in the first stage, by adopting the spectral embedding approach which is well-known in discovering the neighborhood structure of data, we proposed a novel spectral embedding-based algorithm, dubbed Maximized Correlation Cross-modal Spectral Hashing (MCCSH). In which, we carefully initialized and handle the challenging constraints to jointly learn the single-modality spectral representations and the cross-modality \textit{binary} spectral representation, which is well represented for all modalities. We additionally proposed the anchor-to-anchor mapping which is very beneficial in building better similarity graphs by considering the connection among anchors.
%
In the second stage, we proposed to use the deep and powerful CNN architecture to learn hashing functions from informative input data, e.g., images and document embeddings. The informative input data can be helpful in mitigate the discrete problem of textual features (words are considered different regardless their meaning).
Finally, extensive experiments on various benchmark datasets clearly demonstrated the superior performances of DCSH over state-of-the-art unsupervised cross-modality hashing methods.

\section*{Acknowledgement}
This work was supported by ST Electronics and the National Research Foundation(NRF), Prime Minister’s Office, Singapore under Corporate Laboratory @ University Scheme (Programme Title: STEE Infosec - SUTD Corporate Laboratory). This work was also supported by SUTD project PIE-SGP-AI-2018-01.


%
%

\begin{appendices}
\section{Maximized Correlation Cross-modal Binary Representation}
\label{sec:appendix_1}
\subsection{Fix $\{\bY^m\}_{m=1}^M$, update $\bB$} 
\label{sec:appendix_1_1}

In this section, we provide the details of applying MPEC-EPM
method on our binary optimization \eqref{eq:update_B}.

\begin{subequations}
\begin{equation}
\min_\bB\mathcal{L}(\bB), \quad \text{s.t. }\bB\in\{-1,+1\}^{N\times L},  \tag{\ref{eq:update_B}}
\end{equation}
where 
\begin{equation*}
\footnotesize
\begin{split}
    \mathcal{L}(\bB)=-\sum_{m=1}^M\tr\left(\bB\transpose\bY^m\right)
    +\frac{\lambda_1}{4}\left\|\bB\transpose\bB-N\bI_L\right\|_F^2+\frac{\lambda_2}{2}\left\|\bB\transpose\bOnes_{N\times 1}\right\|_F^2,
\end{split}
\end{equation*}
\end{subequations}

Firstly, we introduce an auxiliary variable $\bV \in \bbR^{N\times L}$ and reformulate \eqref{eq:update_B} equivalently as follows:
\begin{equation}
\begin{split}
    \min_{\bB,\bV}~~&\mathcal{L}(\bB),\\
    \text{s.t. }~&-\bOnes_{N\times L}\le \bB\le \bOnes_{N\times L},\\
                & \tr(\bB\transpose\bV)=NL,~~ \|\bV\|_F^2\le NL.
\end{split}
\end{equation}
We further introduce a penalty parameter $\rho$ for the bi-linear constraint $\tr(\bB\transpose\bV)=NL$, the problem becomes:
\begin{equation}
\label{eq:mpec}
\begin{split}
    \min\limits_{\bB,\bV}~&\mathcal{L}(\bB)+\rho(NL -\tr(\bB\transpose\bV)),\\
    \text{s.t. }&-\bOnes_{N\times L}\le \bB\le \bOnes_{N\times L},\|\bV\|_F^2\le NL,
\end{split}
\end{equation}
where $\bOnes_{N\times L}$ and $\bZeros_{N\times L}$ are the ${N\times L}$ matrices of all 1 and 0 respectively.
We alternatively update $\bB$ and $\bV$.

\smallskip
\textbf{Solve $\bB$}: Firstly, by fixing $\bV$, \eqref{eq:mpec} becomes:
\begin{subequations}
\label{eq:fix_v}
    \begin{align}
        \min\limits_{\bB}~&\mathcal{Q}(\bB) = \mathcal{L}(\bB)+\rho(NL -\tr(\bB\transpose\bV)), \tag{\ref{eq:fix_v}}\\
        \text{s.t. }&-\bOnes_{N\times L}\le \bB\le \bOnes_{N\times L}. \label{eq:fix_v_box_constraint}
    \end{align}
\end{subequations}

We propose to solve \eqref{eq:fix_v} using the projected gradient descent method. We initialize $\bV=\bZeros_{N\times L}$ for the sake of finding a reasonable starting point for $\bB$, which is a local optimum of the unconstrained optimization problem of \eqref{eq:update_B}.
With the box constraint \eqref{eq:fix_v_box_constraint}, we can easily project $\bB$ back to its feasible region using $Clamp$ operation, i.e., $\min(\max(\bx, -1), 1)$, after each gradient descent update.
The gradient of $\mathcal{Q}(\bB)$ is given as follows:
\begin{equation}
\label{eq:grad_B}
\footnotesize
\begin{split}
\hspace{-5pt}\nabla_\bB \mathcal{Q}(\bB)= -\hspace{-2pt}\sum_{m=1}^M \hspace{-2pt}\bY^m \hspace{-2pt}+\lambda_1\bB(\bB\transpose\bB-N\bI_L)
    +\lambda_2  \bOnes_{N\times N}\bB-\rho \bV.
\end{split}
\end{equation}

\textbf{Solve $\bV$}: On the other hand, when fixing $\bB$, $\bV$ is obtained by minimizing:
\begin{equation}
\min\limits_\bV -\tr(\bB\transpose\bV),\quad\text{s.t. }\|\bV\|_F^2\le NL.
\end{equation}
Equivalently, $\bV$ is obtained as the following closed-form solution:
\begin{equation}
\label{eq:v_solution}
\bV=\begin{cases}
\frac{\sqrt{NL}}{\|\bB\|_F}\bB, \quad~ \bB\neq \bZeros_{N\times L}\\
\bZeros_{N\times L}, \qquad \text{otherwise}.
\end{cases}
\end{equation}

Similar to \cite{MPEC}, we start the optimization process with a small value of $\rho$, e.g., $\rho=0.01$, and then it is increased by a small factor $\sigma$, e.g., $\sigma=1.05$, after each iteration of alternative updating $\bB$ and $\bV$ until convergence to binary values. 

\begin{algorithm}[t]
\caption{EPM for Binary Constraint}\label{algo:bin}
\SetKwData{Left}{left}\SetKwData{This}{this}\SetKwData{Up}{up}
\SetKwInOut{Input}{Input}\SetKwInOut{Output}{Output}
\Input{$\rho, \sigma, \eta, \lambda_1, \lambda_2,\{\bY^m\}_{m=1}^M$}
\Output{$\bB\in\{-1,+1\}^{N\times L}$}
\BlankLine

Initialize: $\rho=0.1$, $\sigma=1.05$, $\eta=0.01$, $\bV=\bZeros_{N\times L}$;\\
\Repeat{$\bB$ \textit{converges to binary}}{
Update $\bB$: solving the sub-problem \eqref{eq:mpec} using the projected gradient descent: \Repeat{\textit{converges}} {
$\bB \coloneqq Clamp(\bB - \eta \nabla_\bB \mathcal{Q}(\bB), -1, 1)$ where $\nabla_\bB \mathcal{Q}(\bB)$ is in \eqref{eq:grad_B} and a learning rate $\eta$; \\
}
Update $\bV$: using the closed-form solution \eqref{eq:v_solution};\\
Update the penalty: $\rho \coloneqq \rho \times \sigma$; \\
}
\Return{$\bB\in\{-1,+1\}^{N\times L}$}
\end{algorithm}

\subsection{Fix $\bB$, update $\{\bY^m\}_{m=1}^M$} 
\label{sec:appendix_1_2}
The \textit{unconstrained} Augmented Lagrangian function for solving $\bY^m$ is given as follows:
\begin{equation}
\mathcal{L}_{AL}(\bY^m,\Gamma,\mu)=\mathcal{J}(\bB, \bY^m)-\tr(\Gamma\transpose \Phi)+\frac{\mu}{2}\|\Phi\|_F^2, \tag{\ref{eq:LA_loss}}
\end{equation}

\textbf{Initialization of Augmented Lagrangian algorithm:}
The Algorithm \ref{algo:al} requires the initialization for $\bY_0^m$ and $\Gamma_0$. Firstly, we utilize the optimal solution of $\bY^m$ of previous update in Algorithm \ref{algo:cross_modal_bin} as the initialization for the current update $\bY_0^m$.
Secondly, given $\mu_0$ and $\bY_0^m$, we compute the corresponding $\Gamma_0$ by using the optimal condition for the unconstrained minimization \eqref{eq:LA_loss}, i.e.,
\begin{equation}
\nabla_{\bY^m}\mathcal{L}_{AL} = 2\bL^m\bY^m-\alpha\bB- 2\bY^m\Gamma
+2\mu \bY^m\Phi=0.
\end{equation}
We note that $\Phi_0=(\bY_0^m)\transpose \bY_0^m-\bI_L$  equals to zeros, we have
\begin{equation}
\begin{split}
\Gamma_0 &= \left((\bY_0^m)^\top\bY_0^m\right)^{-1}(\bY_0^m)^\top\left(\bL^m\bY_0^m-\frac{\alpha}{2}\bB\right)\\
&=(\bY_0^m)^\top\left(\bL^m\bY_0^m-\frac{\alpha}{2}\bB\right). 
\end{split}
\end{equation}

\begin{algorithm}[t]
\caption{Augmented Lagrangian Algorithm}\label{algo:al}
\SetKwInOut{Input}{Input}\SetKwInOut{Output}{Output}
\Input{$\bB, \bL^m, \alpha, \mu, \sigma, \epsilon$.}
\Output{$\bY^m$}
{Initialize:} $\mu_0=0.01, \sigma=2, \epsilon=10^{-3}$, $\bY^m_0, \Gamma_0$; \\
\For{$t=0\to T_{max}$}{
    Optimize \eqref{eq:LA_loss} w.r.t $\bY^m$: $\bY^m_{t+1}\coloneqq \arg\min_{\bY^m}\mathcal{L}_{AL}(\bY_t^m,\Gamma_t,\mu_t)$;\\
    \If{$|\mathcal{J}(\bB,\bY^m_{t+1}) - \mathcal{J}(\bB,\bY^m_t)|< \epsilon$}{break;\\}
    Update Lagrange multiplier: $\Gamma_{t+1}\coloneqq\Gamma_t-\mu\Phi_{t+1}$;\\
    Update penalty parameter: $\mu_{t+1} \coloneqq \mu_t \times \sigma$;
}
\Return{$\bY^m$}; 
\end{algorithm}

\end{appendices}

\ifCLASSOPTIONcaptionsoff
  \newpage
\fi

\vspace{-0.6em}
{
\bibliographystyle{IEEEtran}
\bibliography{hash}

\begin{thebibliography}{10}
\providecommand{\url}[1]{#1}
\csname url@samestyle\endcsname
\providecommand{\newblock}{\relax}
\providecommand{\bibinfo}[2]{#2}
\providecommand{\BIBentrySTDinterwordspacing}{\spaceskip=0pt\relax}
\providecommand{\BIBentryALTinterwordstretchfactor}{4}
\providecommand{\BIBentryALTinterwordspacing}{\spaceskip=\fontdimen2\font plus
\BIBentryALTinterwordstretchfactor\fontdimen3\font minus
  \fontdimen4\font\relax}
\providecommand{\BIBforeignlanguage}[2]{{%
\expandafter\ifx\csname l@#1\endcsname\relax
\typeout{** WARNING: IEEEtran.bst: No hyphenation pattern has been}%
\typeout{** loaded for the language `#1'. Using the pattern for}%
\typeout{** the default language instead.}%
\else
\language=\csname l@#1\endcsname
\fi
#2}}
\providecommand{\BIBdecl}{\relax}
\BIBdecl

\bibitem{CRH}
Y.~Zhen and D.-Y. Yeung, ``{Co-Regularized Hashing for Multimodal Data},'' in
  \emph{NIPS}, 2012, pp. 1376--1384.

\bibitem{HTH}
Y.~Wei, Y.~Song, Y.~Zhen, B.~Liu, and Q.~Yang, ``{Heterogeneous Translated
  Hashing: A Scalable Solution Towards Multi-Modal Similarity Search},''
  \emph{ACM Trans. Knowl. Discov. Data}, vol.~10, no.~4, 2016.

\bibitem{SMH}
D.~Zhangy and W.-J. Li, ``{Large-scale supervised multimodal hashing with
  semantic correlation maximization},'' in \emph{AAAI}, 2014, pp. 2177--2183.

\bibitem{QCH}
B.~Wu, Q.~Yang, W.-S. Zheng, Y.~Wang, and J.~Wang, ``{Quantized Correlation
  Hashing for Fast Cross-Modal Search},'' in \emph{IJCAI}, 2015, pp.
  3946--3952.

\bibitem{SePH}
Z.~Lin, G.~Ding, M.~Hu, and J.~Wang, ``{Semantics-preserving hashing for
  cross-view retrieval},'' in \emph{CVPR}, 2015, pp. 3864--3872.

\bibitem{DCH}
X.~Xu, F.~Shen, Y.~Yang, H.~T. Shen, and X.~Li, ``{Learning Discriminative
  Binary Codes for Large-scale Cross-modal Retrieval},'' \emph{IEEE TIP},
  vol.~26, no.~5, pp. 2494--2507, 2017.

\bibitem{DLFH}
Q.-Y. Jiang and W.-J. Li, ``{Discrete Latent Factor Model for Cross-Modal
  Hashing},'' \emph{IEEE TIP}, vol.~28, no.~7, pp. 3490--3501, 2019.

\bibitem{SMFH}
J.~Tang, K.~Wang, and L.~Shao, ``{Supervised Matrix Factorization Hashing for
  Cross-Modal Retrieval},'' \emph{IEEE TIP}, vol.~25, no.~7, pp. 3157--3166,
  2016.

\bibitem{8425016}
D.~Mandal, K.~N. Chaudhury, and S.~Biswas, ``{Generalized Semantic Preserving
  Hashing for Cross-Modal Retrieval},'' \emph{IEEE TIP}, vol.~28, no.~1, pp.
  102--112, Jan 2019.

\bibitem{8331146}
C.~Deng, Z.~Chen, X.~Liu, X.~Gao, and D.~Tao, ``{Triplet-Based Deep Hashing
  Network for Cross-Modal Retrieval},'' \emph{IEEE TIP}, vol.~27, no.~8, pp.
  3893--3903, Aug 2018.

\bibitem{THN}
Z.~Cao, M.~Long, J.~Wang, and Q.~Yang, ``{Transitive Hashing Network for
  Heterogeneous Multimedia Retrieval},'' in \emph{AAAI}, 2017, pp. 81--87.

\bibitem{DCMH}
Q.-Y. Jiang and W.-J. Li, ``{Deep Cross-Modal Hashing},'' in \emph{CVPR}, July
  2017, pp. 3232--3240.

\bibitem{DDCMH}
Z.-D. Chen, W.-J. Yu, C.-X. Li, L.~Nie, and X.-S. Xu, ``{Dual Deep Neural
  Networks Cross-Modal Hashing},'' in \emph{AAAI}, 2018, pp. 274--281.

\bibitem{CMDVH}
V.~E. Liong, J.~Lu, Y.-P. Tan, and J.~Zhou, ``{Cross-Modal Deep Variational
  Hashing},'' in \emph{ICCV}, 2017, pp. 4077--4085.

\bibitem{PRDH}
E.~Yang, C.~Deng, W.~Liu, X.~Liu, D.~Tao, and X.~Gao, ``{Pairwise Relationship
  Guided Deep Hashing for Cross-Modal Retrieval},'' in \emph{AAAI}, 2017, pp.
  1618--1625.

\bibitem{DVSH}
Y.~Cao, M.~Long, J.~Wang, Q.~Yang, and P.~S. Yu, ``{Deep Visual-Semantic
  Hashing for Cross-Modal Retrieval},'' in \emph{ACM SIGKDD}, 2016, pp.
  1445--1454.

\bibitem{CVH}
S.~Kumar and R.~Udupa, ``{Learning Hash Functions for Cross-view Similarity
  Search},'' in \emph{IJCAI}, 2011, pp. 1360--1365.

\bibitem{PDH}
M.~Rastegari, J.~Choi, S.~Fakhraei, D.~Hal, and L.~Davis, ``{Predictable
  Dual-View Hashing},'' in \emph{ICML}, 2013, p. 1328–1336.

\bibitem{IMH}
J.~Song, Y.~Yang, Y.~Yang, Z.~Huang, and H.~T. Shen, ``{Inter-media Hashing for
  Large-scale Retrieval from Heterogeneous Data Sources},'' in \emph{ACM
  SIGMOD}, 2013, pp. 785--796.

\bibitem{CMFH}
G.~Ding, Y.~Guo, J.~Zhou, and Y.~Gao, ``{Large-Scale Cross-Modality Search via
  Collective Matrix Factorization Hashing},'' \emph{IEEE TIP}, vol.~25, no.~11,
  pp. 5427--5440, 2016.

\bibitem{LSSH}
J.~Zhou, G.~Ding, and Y.~Guo, ``{Latent Semantic Sparse Hashing for Cross-modal
  Similarity Search},'' in \emph{ACM SIGIR}, 2014, pp. 415--424.

\bibitem{ACQ}
G.~Irie, H.~Arai, and Y.~Taniguchi, ``{Alternating Co-Quantization for
  Cross-Modal Hashing},'' in \emph{ICCV}, 2015, pp. 1886--1894.

\bibitem{FSH}
H.~Liu, R.~Ji, Y.~Wu, F.~Huang, and B.~Zhang, ``{Cross-Modality Binary Code
  Learning via Fusion Similarity Hashing},'' in \emph{CVPR}, 2017, pp.
  7380--7388.

\bibitem{CRE}
M.~Hu, Y.~Yang, F.~Shen, N.~Xie, R.~Hong, and H.~T. Shen, ``{Collective
  Reconstructive Embeddings for Cross-Modal Hashing},'' \emph{IEEE TIP},
  vol.~28, no.~6, pp. 2770--2784, June 2019.

\bibitem{DMHOR}
D.~Wang, P.~Cui, M.~Ou, and W.~Zhu, ``{Learning Compact Hash Codes for
  Multimodal Representations Using Orthogonal Deep Structure},'' \emph{IEEE
  TMM}, vol.~17, no.~9, pp. 1404--1416, 2015.

\bibitem{MSAE}
W.~Wang, B.~C. Ooi, X.~Yang, D.~Zhang, and Y.~Zhuang, ``{Effective Multi-modal
  Retrieval Based on Stacked Auto-encoders},'' \emph{VLDB Endowment}, vol.~7,
  no.~8, pp. 649--660, 2014.

\bibitem{CorrAE}
F.~Feng, X.~Wang, and R.~Li, ``{Cross-modal Retrieval with Correspondence
  Autoencoder},'' in \emph{ACM Multimedia}, 2014, pp. 7--16.

\bibitem{UGACH}
J.~G. Zhang, Y.~Peng, and M.~Yuan, ``{Unsupervised Generative Adversarial
  Cross-Modal Hashing},'' in \emph{AAAI}, 2018, pp. 539--546.

\bibitem{SCQ}
T.~Hoang, T.-T. Do, H.~Le, D.-K. Le-Tan, and N.-M. Cheung, ``Simultaneous
  compression and quantization: A joint approach for efficient unsupervised
  hashing,'' \emph{CVIU}, vol. 191, p. 102852, 2020.

\bibitem{SelfTaughtHF}
D.~Zhang, J.~Wang, D.~Cai, and J.~Lu, ``{Self-Taught Hashing for Fast
  Similarity Search},'' in \emph{SIGIR}, 2010, pp. 18--25.

\bibitem{2stephash}
G.~Lin, C.~Shen, D.~Suter, and A.~van~den Hengel, ``{A General Two-Step
  Approach to Learning-Based Hashing},'' 2013, pp. 2552--2559.

\bibitem{SpectralClusterTutorial}
U.~Von~Luxburg, ``A tutorial on spectral clustering,'' \emph{Statistics and
  computing}, vol.~17, no.~4, pp. 395--416, 2007.

\bibitem{ACMR}
B.~Wang, Y.~Yang, X.~Xu, A.~Hanjalic, and H.~T. Shen, ``{Adversarial
  Cross-Modal Retrieval},'' in \emph{ACM Multimedia}, 2017, p. 154–162.

\bibitem{SpH}
Y.~Weiss, A.~Torralba, and R.~Fergus, ``Spectral hashing,'' in \emph{NIPS},
  2009, pp. 1753--1760.

\bibitem{5539994}
J.~Wang, S.~Kumar, and S.-F. Chang, ``{Semi-supervised hashing for scalable
  image retrieval},'' in \emph{CVPR}, 2010, pp. 3424--3431.

\bibitem{AnchorGraph}
W.~Liu, J.~He, and S.-F. Chang, ``{Large Graph Construction for Scalable
  Semi-supervised Learning},'' in \emph{ICML}, 2010, pp. 679--686.

\bibitem{DGH}
W.~Liu, C.~Mu, S.~Kumar, and S.-F. Chang, ``{Discrete Graph Hashing},'' in
  \emph{NIPS}, 2014, pp. 3419--3427.

\bibitem{SADH}
F.~Shen, Y.~Xu, L.~Liu, Y.~Yang, Z.~Huang, and H.~T. Shen, ``{Unsupervised Deep
  Hashing with Similarity-Adaptive and Discrete Optimization},'' \emph{IEEE
  TPAMI}, vol.~40, no.~12, pp. 3034--3044, 2018.

\bibitem{glove}
J.~Pennington, R.~Socher, and C.~Manning, ``{Glove: Global vectors for word
  representation},'' in \emph{EMNLP}, 2014, pp. 1532--1543.

\bibitem{word2vec}
T.~Mikolov, K.~Chen, G.~Corrado, and J.~Dean, ``{Efficient estimation of word
  representations in vector space},'' in \emph{ICLR Workshop}, 2013, pp.
  1301--3781.

\bibitem{DBLP:journals/corr/WangLKC15}
J.~Wang, W.~Liu, S.~Kumar, and S.~Chang, ``{Learning to Hash for Indexing Big
  Data - A Survey},'' \emph{Proceedings of the {IEEE}}, vol. 104, no.~1, pp.
  34--57, 2016.

\bibitem{7915742}
J.~Wang, T.~Zhang, J.~Song, N.~Sebe, and H.~T. Shen, ``{A Survey on Learning to
  Hash},'' \emph{IEEE TPAMI}, vol.~40, no.~4, pp. 769--790, 2018.

\bibitem{Grauman_review}
K.~Grauman and R.~Fergus, ``{Learning Binary Hash Codes for Large-Scale Image
  Search},'' \emph{Machine Learning for Computer Vision}, pp. 49--87, 2013.

\bibitem{lsh_vldb09}
A.~Gionis, P.~Indyk, and R.~Motwani, ``{Similarity search in high dimensions
  via hashing},'' in \emph{VLDB}, 1999, p. 518–529.

\bibitem{KLSH_iccv09}
B.~Kulis and K.~Grauman, ``{Kernelized locality-sensitive hashing for scalable
  image search},'' in \emph{ICCV}, 2009, pp. 2130--2137.

\bibitem{KLSH_nips09}
M.~Raginsky and S.~Lazebnik, ``{Locality-Sensitive Binary Codes from
  Shift-Invariant Kernels},'' in \emph{NIPS}, 2009, pp. 1509--1517.

\bibitem{DBLP:journals/pami/KulisJG09}
B.~Kulis, P.~Jain, and K.~Grauman, ``{Fast Similarity Search for Learned
  Metrics},'' \emph{IEEE TPAMI}, vol.~31, no.~12, pp. 2143--2157, 2009.

\bibitem{ITQ}
Y.~Gong, S.~Lazebnik, A.~Gordo, and F.~Perronnin, ``{Iterative Quantization: A
  Procrustean Approach to Learning Binary Codes for Large-Scale Image
  Retrieval},'' \emph{IEEE TPAMI}, vol.~35, no.~12, pp. 2916--2929, 2013.

\bibitem{DBLP:conf/cvpr/HeWS13}
K.~He, F.~Wen, and J.~Sun, ``{K-Means Hashing: An Affinity-Preserving
  Quantization Method for Learning Binary Compact Codes},'' in \emph{CVPR},
  2013, pp. 2938--2945.

\bibitem{BA}
M.~A. Carreira-Perpinan and R.~Raziperchikolaei, ``Hashing with binary
  autoencoders,'' in \emph{CVPR}, 2015, pp. 557--566.

\bibitem{CVPR12:SphericalHashing}
J.-P. Heo, Y.~Lee, J.~He, S.-F. Chang, and S.-E. Yoon, ``{Spherical Hashing},''
  in \emph{CVPR}, 2012, pp. 2957--2964.

\bibitem{DeepHash_TIP17}
J.~Lu, V.~E. Liong, and J.~Zhou, ``{Deep Hashing for Scalable Image Search},''
  \emph{IEEE TIP}, vol.~26, no.~5, pp. 2352--2367, 2017.

\bibitem{deepbit2016}
K.~Lin, J.~Lu, C.-S. Chen, and J.~Zhou, ``{Learning Compact Binary Descriptors
  with Unsupervised Deep Neural Networks},'' in \emph{CVPR}, 2016, pp.
  1183--1192.

\bibitem{8801918}
T.-T. Do, T.~Hoang, D.-K. Le-Tan, A.-D. Doan, and N.-M. Cheung, ``Compact hash
  code learning with binary deep neural network,'' \emph{IEEE TMM}, vol.~22,
  no.~4, pp. 992--1004, 2020.

\bibitem{8709820}
T.-T. Do, D.-K. Le-Tan, T.~{Hoang}, H.~{Le}, T.~V. {Nguyen}, and N.-M.
  {Cheung}, ``Simultaneous feature aggregating and hashing for compact binary
  code learning,'' \emph{IEEE TIP}, vol.~28, no.~10, pp. 4954--4969, 2019.

\bibitem{Kulis_learningto}
B.~Kulis and T.~Darrell, ``{Learning to hash with binary reconstructive
  embeddings},'' in \emph{NIPS}, 2009, pp. 1042--1050.

\bibitem{CVPR12:Hashing}
W.~Liu, J.~Wang, R.~Ji, Y.-G. Jiang, and S.-F. Chang, ``{Supervised Hashing
  with Kernels},'' in \emph{CVPR}, 2012, pp. 2074--2081.

\bibitem{CVPR2014Lin}
G.~Lin, C.~Shen, Q.~Shi, A.~{van den Hengel}, and D.~Suter, ``{Fast Supervised
  Hashing with Decision Trees for High-Dimensional Data},'' in \emph{CVPR},
  2014, pp. 1971--1978.

\bibitem{Shen_2015_CVPR}
F.~Shen, C.~Shen, W.~Liu, and H.~Tao~Shen, ``{Supervised Discrete Hashing},''
  in \emph{CVPR}, 2015, pp. 37--45.

\bibitem{DBLP:conf/icml/NorouziF11}
M.~Norouzi and D.~J. Fleet, ``{Minimal Loss Hashing for Compact Binary
  Codes},'' in \emph{ICML}, 2011, pp. 353--360.

\bibitem{DBLP:journals/pami/WangKC12}
J.~Wang, S.~Kumar, and S.~Chang, ``{Semi-Supervised Hashing for Large-Scale
  Search},'' \emph{IEEE TPAMI}, vol.~34, no.~12, pp. 2393--2406, 2012.

\bibitem{DSRH}
F.~Zhao, Y.~Huang, L.~Wang, and T.~Tan, ``{Deep semantic ranking based hashing
  for multi-label image retrieval},'' in \emph{CVPR}, 2015, pp. 1556--1564.

\bibitem{DRSCH}
R.~Zhang, L.~Lin, R.~Zhang, W.~Zuo, and L.~Zhang, ``{Bit-Scalable Deep Hashing
  With Regularized Similarity Learning for Image Retrieval and Person
  Re-Identification},'' \emph{IEEE TIP}, vol.~24, no.~12, pp. 4766--4779, 2015.

\bibitem{simulfeature}
H.~Lai, Y.~Pan, Y.~Liu, and S.~Yan, ``{Simultaneous feature learning and hash
  coding with deep neural networks},'' in \emph{CVPR}, 2015, pp. 3270--3278.

\bibitem{8451192}
D.-K. Le-Tan, T.-T. Do, and N.-M. {Cheung}, ``Supervised hashing with
  end-to-end binary deep neural network,'' in \emph{IEEE ICIP}, 2018, pp.
  3019--3023.

\bibitem{DSH}
H.~Liu, R.~Wang, S.~Shan, and X.~Chen, ``{Deep Supervised Hashing for Fast
  Image Retrieval},'' in \emph{CVPR}, 2016, pp. 2064--2072.

\bibitem{DHN}
H.~Zhu, M.~Long, J.~Wang, and Y.~Cao, ``{Deep Hashing Network for Efficient
  Similarity Retrieval},'' in \emph{AAAI}, 2016, pp. 2415--2421.

\bibitem{DPSH}
W.~Li, S.~Wang, and W.~Kang, ``{Feature Learning based Deep Supervised Hashing
  with Pairwise Labels},'' in \emph{IJCAI}, 2016, pp. 1711--1717.

\bibitem{CauchyHashing}
Y.~Cao, M.~Long, B.~Liu, and J.~Wang, ``{Deep Cauchy Hashing for Hamming Space
  Retrieval},'' in \emph{CVPR}, 2018, pp. 1229--1237.

\bibitem{SSAH}
C.~Li, C.~Deng, N.~Li, W.~Liu, X.~Gao, and D.~Tao, ``{Self-Supervised
  Adversarial Hashing Networks for Cross-Modal Retrieval},'' in \emph{CVPR},
  2018, pp. 4242--4251.

\bibitem{CMHH}
Y.~Cao, B.~Liu, M.~Long, and J.~Wang, ``{Cross-Modal Hamming Hashing},'' in
  \emph{ECCV}, 2018, pp. 207--223.

\bibitem{NSS}
X.~Bai, S.~Bai, and X.~Wang, ``{Beyond diffusion process: Neighbor set
  similarity for fast re-ranking},'' \emph{Information Sciences}, vol. 325, pp.
  342 -- 354, 2015.

\bibitem{UH-BDNN}
T.-T. Do, A.-D. Doan, and N.-M. Cheung, ``Learning to hash with binary deep
  neural network,'' in \emph{ECCV}, 2016, pp. 219--234.

\bibitem{UDCMH}
G.~Wu, Z.~Lin, J.~Han, L.~Liu, G.~Ding, B.~Zhang, and J.~Shen, ``{Unsupervised
  Deep Hashing via Binary Latent Factor Models for Large-scale Cross-modal
  Retrieval},'' in \emph{IJCAI}, 2018, pp. 2854--2860.

\bibitem{DJSRH}
S.~Su, Z.~Zhong, and C.~Zhang, ``{Deep Joint-Semantics Reconstructing Hashing
  for Large-Scale Unsupervised Cross-Modal Retrieval},'' in \emph{ICCV}, 2019,
  pp. 3027--3035.

\bibitem{CYC-DGH}
L.~Wu, Y.~Wang, and L.~Shao, ``{Cycle-Consistent Deep Generative Hashing for
  Cross-Modal Retrieval},'' \emph{IEEE TIP}, vol.~28, no.~4, pp. 1602--1612,
  2019.

\bibitem{CycleGAN2017}
J.-Y. Zhu, T.~Park, P.~Isola, and A.~A. Efros, ``{Unpaired Image-to-Image
  Translation using Cycle-Consistent Adversarial Networks},'' in \emph{ICCV},
  2017, pp. 2223--2232.

\bibitem{DeepVQ}
D.-K. Le-Tan, H.~Le, T.~Hoang, T.-T. Do, and N.-M. Cheung, ``{DeepVQ: A Deep
  Network Architecture for Vector Quantization},'' in \emph{CVPR Workshop},
  2018, pp. 2579--2582.

\bibitem{mutual_kmeans}
P.~Kontschieder, M.~Donoser, and H.~Bischof, ``{Beyond Pairwise Shape
  Similarity Analysis},'' in \emph{ACCV}, 2010, pp. 655--666.

\bibitem{orthogonalProcrustes}
P.~H. Sch{\"o}nemann, ``A generalized solution of the orthogonal procrustes
  problem,'' \emph{Psychometrika}, vol.~10, pp. 1--10, 1966.

\bibitem{Nystrom_spectal_clustering}
C.~Fowlkes, S.~Belongie, F.~Chung, and J.~Malik, ``{Spectral grouping using the
  Nystrom method},'' \emph{IEEE Transactions on Pattern Analysis and Machine
  Intelligence}, vol.~26, no.~2, pp. 214--225, 2004.

\bibitem{power_method_spectal_clustering}
C.~Boutsidis, A.~Gittens, and P.~Kambadur, ``{Spectral Clustering via the Power
  Method - Provably},'' in \emph{ICML}, 2015, pp. 40--48.

\bibitem{minibatch_spectral_clustering}
Y.~Han and M.~Filippone, ``{Mini-batch spectral clustering},'' in \emph{IJCNN},
  2017, pp. 3888--3895.

\bibitem{MPEC}
G.~Yuan and B.~Ghanem, ``{An Exact Penalty Method for Binary Optimization Based
  on MPEC Formulation},'' in \emph{AAAI}, 2017, pp. 2867--2875.

\bibitem{nocedal2006numerical}
J.~Nocedal and S.~Wright, ``Numerical optimization,'' \emph{Springer series in
  operations research, New York, USA}, 2006.

\bibitem{alexnet}
A.~Krizhevsky, I.~Sutskever, and G.~E. Hinton, ``Imagenet classification with
  deep convolutional neural networks,'' in \emph{NIPS}, 2012, pp. 1097--1105.

\bibitem{VGG}
K.~Simonyan and A.~Zisserman, ``Very deep convolutional networks for
  large-scale image recognition,'' \emph{arXiv preprint arXiv:1409.1556}, 2014.

\bibitem{resnet}
K.~He, X.~Zhang, S.~Ren, and J.~Sun, ``Deep residual learning for image
  recognition,'' in \emph{CVPR}, June 2016, pp. 770--778.

\bibitem{MIRFlickr25k}
M.~J. Huiskes and M.~S. Lew, ``{The MIR Flickr Retrieval Evaluation},'' in
  \emph{ACM International Conference on Multimedia Information Retrieval},
  2008, pp. 39--43.

\bibitem{nuswide}
T.-S. Chua, J.~Tang, R.~Hong, H.~Li, Z.~Luo, and Y.~Zheng, ``{NUS-WIDE: A}
  real-world web image database from {N}ational {U}niversity of {S}ingapore,''
  in \emph{CIVR}, 2009, pp. 1--9.

\bibitem{graphHashing}
W.~Liu, J.~Wang, S.~Kumar, and S.-F. Chang, ``{Hashing with Graphs},'' in
  \emph{ICML}, 2011, pp. 1--8.

\bibitem{SSDH}
H.~Yang, K.~Lin, and C.~Chen, ``{Supervised Learning of Semantics-Preserving
  Hash via Deep Convolutional Neural Networks},'' \emph{IEEE TPAMI}, vol.~40,
  no.~2, pp. 437--451, 2018.

\bibitem{S3PLH}
J.~Wang, S.~Kumar, and S.-F. Chang, ``{Sequential Projection Learning for
  Hashing with Compact Codes},'' in \emph{ICML}, 2010, pp. 1127--1134.

\bibitem{8296975}
T.~Hoang, T.-T. Do, D.-K. Le-Tan, and N.-M. Cheung, ``Enhancing feature
  discrimination for unsupervised hashing,'' in \emph{IEEE ICIP}, 2017, pp.
  3710--3714.

\bibitem{tsne}
L.~van~der Maaten and G.~Hinton, ``{Visualizing data using t-SNE},'' \emph{The
  Journal of Machine Learning Research}, vol.~9, pp. 2579--2605, 2008.

\end{thebibliography}
}

\end{document}